\documentclass[letterpaper, 10 pt, conference]{ieeeconf}  

\IEEEoverridecommandlockouts                              

\usepackage{cite}
\usepackage{amsmath,amssymb,amsfonts}
\usepackage{algorithmic}
\usepackage{graphicx}
\usepackage{textcomp}
\usepackage{xcolor}
\usepackage{adjustbox, multirow, bm}
\usepackage{booktabs}
\usepackage{caption}
\usepackage{subcaption} 
\usepackage{placeins}
\usepackage{hyperref}
\usepackage[capitalise]{cleveref}
\newcommand{\bb}{\mathbf}
\title{\LARGE \bf
Bi-Manual Joint Camera Calibration and Scene Representation
}
\author{Haozhan Tang$^{1}$ \and  Tianyi Zhang$^{1}$ \and Matthew Johnson-Roberson$^{1, 2}$ \and Weiming Zhi$^{1}$
\thanks{Project page: \url{https://tomtang502.github.io/bijcr_web/}.}%
\thanks{$^{1}$ Robotics Institute, Carnegie Mellon University, Pittsburgh, USA.}%
\thanks{$^{2}$ College of Connected Computing, Vanderbilt University, USA.}%
\thanks{$^{*}$email: {\tt\small wzhi@andrew.cmu.edu}.}%
}

\begin{document}

\maketitle
\thispagestyle{empty}
\pagestyle{empty}

\begin{abstract}
Robot manipulation, especially bimanual manipulation, often requires setting up multiple cameras on multiple robot manipulators. Before robot manipulators can generate motion or even build representations of their environments, the cameras rigidly mounted to the robot need to be calibrated. Camera calibration is a cumbersome process involving collecting a set of images, with each capturing a pre-determined marker. In this work, we introduce the Bi-Manual Joint Calibration and Representation Framework (Bi-JCR). Bi-JCR enables multiple robot manipulators, each with cameras mounted, to circumvent taking images of calibration markers. By leveraging 3D foundation models for dense, marker-free multi-view correspondence, Bi-JCR jointly estimates: (i) the extrinsic transformation from each camera to its end-effector, (ii) the inter-arm relative poses between manipulators, and (iii) a unified, scale-consistent 3D representation of the shared workspace, all from the same captured RGB image sets. The representation, jointly constructed from images captured by cameras on both manipulators, lives in a common coordinate frame and supports collision checking and semantic segmentation to facilitate downstream bimanual coordination tasks. We empirically evaluate the robustness of Bi-JCR on a variety of tabletop environments, and demonstrate its applicability on a variety of downstream tasks.
\end{abstract}

\section{INTRODUCTION}
Robot manipulators with wrist-mounted cameras generally need to be meticulously calibrated offline to enable perceived objects to be transformed into the robot's coordinate frame. This is done via a procedure known as hand-eye calibration, where the manipulator is moved through a set of poses and take images of a known calibration marker, such as a checker board or AprilTag \cite{AprilTag}. Traditional hand-eye calibration methods focus on a single “eye-in-hand” camera and rely on external markers to compute the rigid transform between camera and end-effector. When extended to two independently moving arms, these approaches must be repeated separately for each arm, and then a secondary step is required to fuse the two coordinate frames. In this work, we tackle the problem of calibrating of dual manipulators with wrist-mounted cameras without using any calibration markers. Here, we assume that the poses of the cameras relative to the end-effectors, along with the relative poses of the manipulator bases are unknown and require estimation.
\begin{figure}[t]
    \centering
    \includegraphics[width=0.98\linewidth]{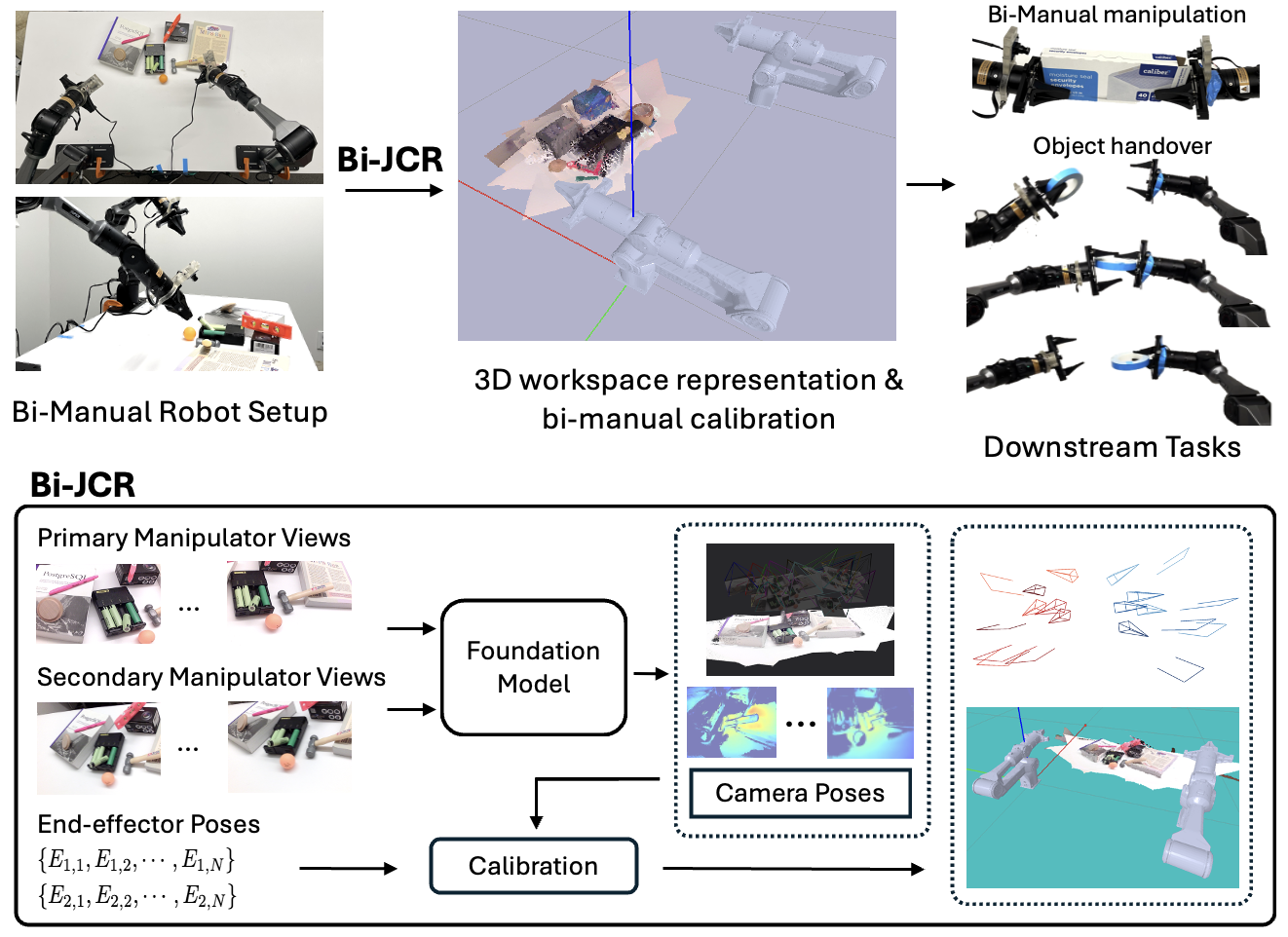}
    \caption{We tackle a bi-manual setup, where the extrinsics of both cameras and the relative poses of the robot bases to one another are unknown. Bi-JCR solves to recover all three transformations.}
    \label{fig:Setup}
    \vspace{-2em}
\end{figure}

Here, we propose a framework called Bi-manual Joint Representation and Calibration (Bi-JCR) that simultaneously builds a representation of the environment and calibrates both cameras for dual-manipulators with wrist-mounted cameras. Bi-JCR uses \textbf{the same set of images} for both calibrating the camera and constructing the environment representation. It completely avoids the need to calibrate offline with markers. Bi-JCR leverages modern \emph{3D foundation models} to efficiently estimate an unscaled representation along with unscaled camera poses, from a set of images captured by the dual manipulators. Then, by considering the forward kinematics of each arm, we formulate a joint scale recovery and dual calibration problem which can subsequently be solved via gradient descent on a manifold of transformation matrices. By optimizing across a single calibration problem defined using images from both arms, Bi-JCR simultaneously solves for each hand-eye transform, aligns the two robot base frames, recovers a missing scale factor, and directly yields the rigid transform between the two manipulators. This enables immediate fusion of visual data across both viewpoints for bimanual manipulation, without reliance on external markers or depth sensors. 

We empirically evaluate Bi-JCR and demonstrate its ability to accurately calibrate cameras on both manipulators, and produce a dense and size-accurate representation of the environment that can be transformed into the workspace coordinate frame. We leverage the representation into downstream manipulation and to execute successful grasps and bi-manual hand-overs. Concretely, our contributions include: 
\begin{itemize}
\item The Bi-manual Joint Calibration and Representation (Bi-JCR) method that leverages 3D foundation models to build an environment representation built from wrist-mounted cameras, while calibrating the cameras; 
\item The formulation of a novel optimization problem that recovers camera transformations on both manipulators, the relative pose between the manipulators, and a scale factor to obtain metric scale from representation; 
\item Rigorous evaluation on real-world data, and evaluation of performance ablating over many of the state-of-the-art 3D foundation models integrated into Bi-JCR.
\end{itemize}

\section{Related Work}

\begin{figure}[t]
\centering
\includegraphics[width=0.95\linewidth]{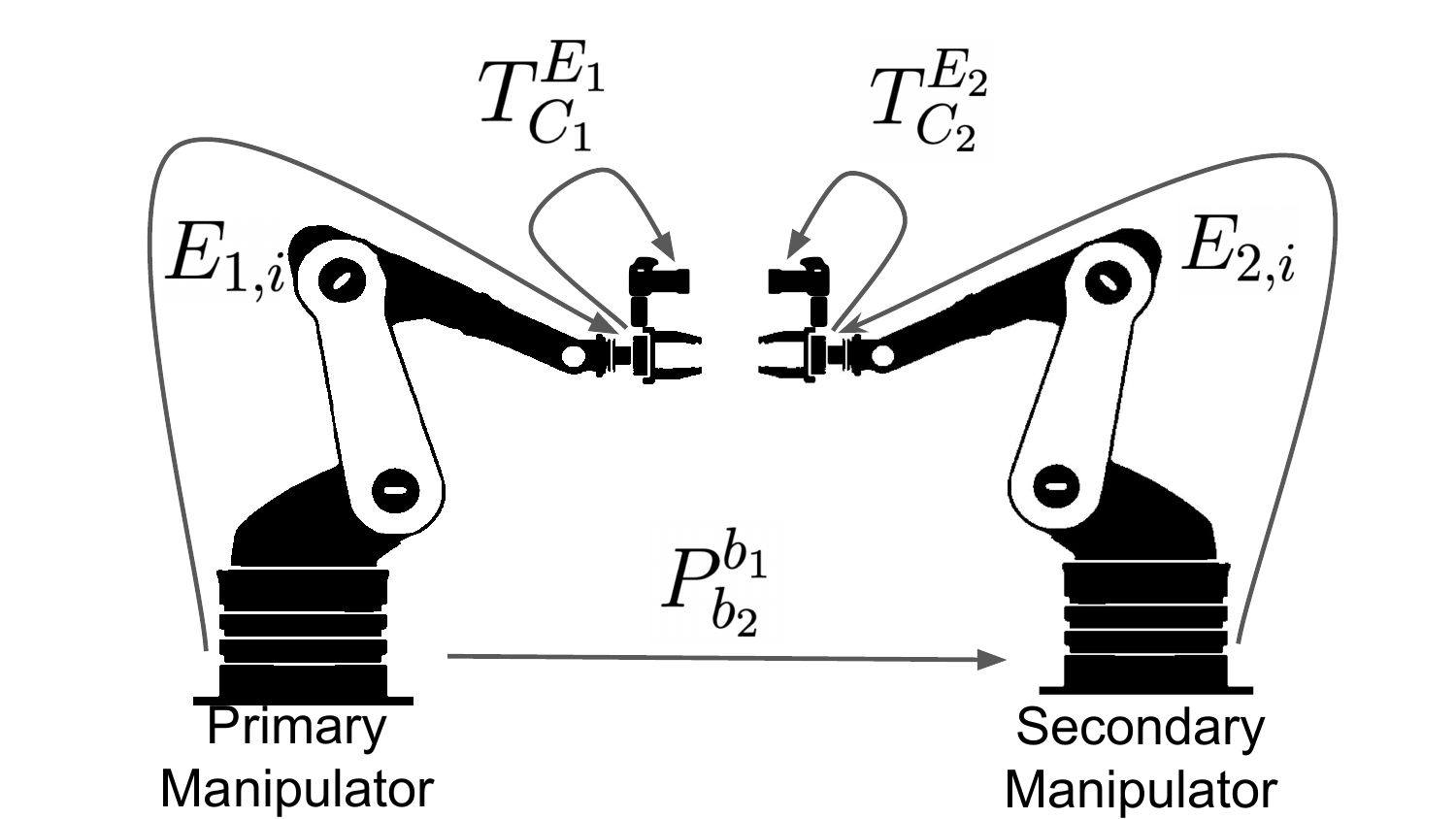}
\caption{The transformations between each arm, along with their cameras, are shown here. The transformations $T^{E_1}_{C_1}$, $T^{E_2}_{C_2}$, and $P^{b_{1}}_{b_{2}}$ are all unknown and will be recovered via Bi-JCR.}\label{fig: diagram}
\vspace{-1em}
\end{figure}
\textbf{Hand–Eye Calibration:} 
Decades-old closed-form solvers \cite{Tsai1988ANT, hand-eye, Park+Martin-1994} remain the de facto standard because they are fast and provably correct under marker-based conditions without noise.  Yet in modern labs, the very assumptions they rely on, static checkerboard, are routinely violated.  Recent learning-based variants regress the transform directly from images \cite{pmlr-v164-valassakis22a}, but demand gripper visibility or prior CAD models and often degrade sharply outside the synthetic domain in which they are trained. Other end-to-end policies learning-based methods bypass calibration entirely, mapping pixels to torques \cite{decision_trans,implicitBC}, but at the cost of losing an explicit transform that downstream planners and safety monitors still persist. Foundation models for calibration are also explored in \cite{zhi2024unifying, zhi20243d}, but have not been extended to the bi-manual setting.

\textbf{Bi-manual Calibration: } Extending single manipulator hand-eye calibration to the bi-manual setup is not trivial as it requires finding the pose of the secondary manipulator in the primary manipulator's frame. Previously, some methods relied on the secondary manipulator holding a checker board to perform bi-manual calibration \cite{wu2016simultaneous, fu2020dual}. A graph-based method that uses external markers to calibrate multiple manipulators simultaneously has also been explored in \cite{zhou2023simultaneously}.

\textbf{Scene Representation:}
Bimanual manipulation requires reasoning about \emph{shared} workspaces where two end-effectors and several movable objects compete for space.  Classical metric maps such as occupation grids \cite{OccupancyGridMaps} and signed distance fields \cite{SDF,kinect_fusion} give fast binary or distance queries for collision checking, yet they discretize space and struggle to capture fine contact geometry in small parts. Continuous alternatives such as Gaussian process maps \cite{SimonGPOM}, kernel regressors \cite{HM,sptemp}, and neural implicit surfaces \cite{Park_2019_CVPR}, offer subvoxel accuracy. Learning methods that \emph{directly} consume point clouds \cite{pointNet, pointnet_plusplus} or integrate them into trajectory optimization \cite{GeoFab_gloabL_opt}. In the computer vision community, proposed photorealistic NeRF models \cite{mildenhall2020nerf,mueller2022instant,zhang2024darkgs} produce photorealistic models, at the expense of accurate geometry.

\textbf{Foundation Models:}
Large-scale models trained in web corpora, LLMs in NLP \cite{touvron2023llama} and multimodal encoders in vision \cite{clip} have recently been explored for 3D perception \cite{DUSt3R_cvpr24, depthanything, mast3r_arxiv24, wang2025vggt}. In robotics, these large pre-trained deep learning models are referred to as ``foundation models'', gaining increasing applications when used as plug-and-play modules for downstream tasks \cite{Bommasani2021FoundationModels,Firoozi2023FoundationMI}. In our work, we leverage these 3D foundation models as components in our pipeline.

\begin{figure}[t]
\centering
\fbox{%
\includegraphics[width=0.363\linewidth]{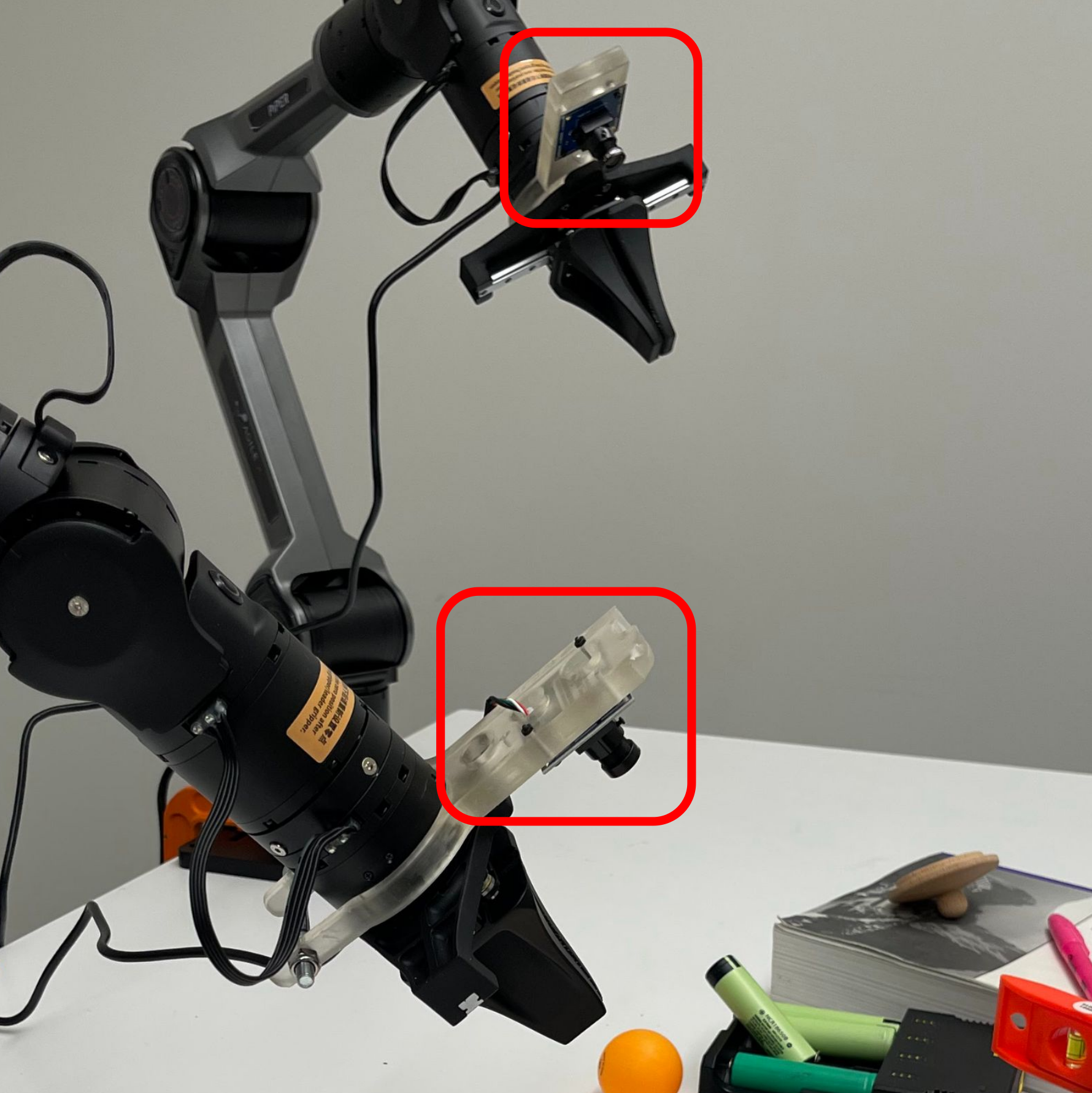}%
}
\fbox{%
\includegraphics[width=0.6\linewidth]{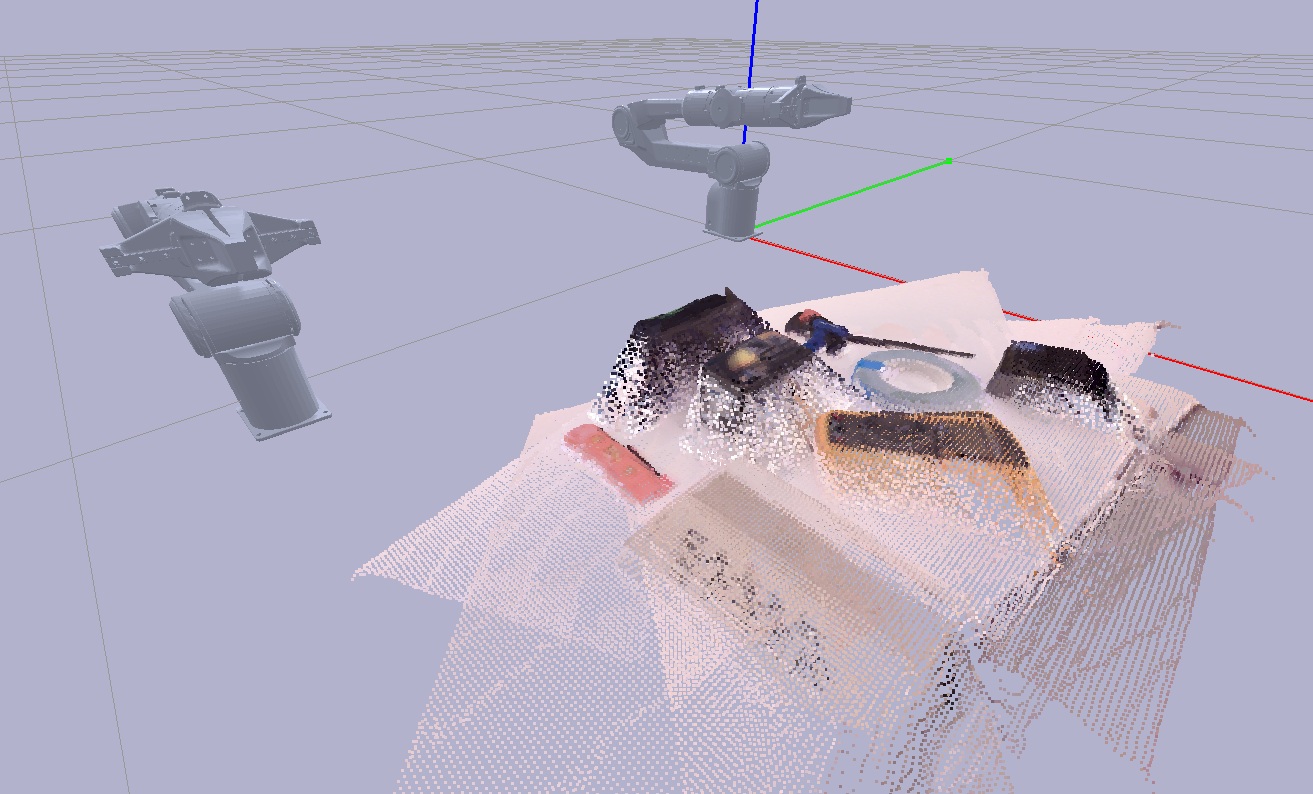}%
}%
\caption{Left: We have cameras (outlined in red) rigidly attached to dual manipulators; Right: With the camera calibrated, we can bring a 3D reconstruction directly into the frame of the robot, and even inject it into a simulator (visualised in PyBullet \cite{coumans2019}).} \label{fig: camera_sim}
\vspace{-1em}
\end{figure}

\section{Bi-manual Joint Representation and Calibration}
The proposed Bi-manual Joint Representation and Calibration (Bi-JCR) framework aims to solve the eye-to-hand calibration problem for both manipulators, and in the process, also recover the relative poses of the robot base. At the same time, we can recover a dense 3D representation of the tabletop scene, which can facilitate downstream manipulation. This process is done without relying on any camera pose information from external markers, such as checkerboards or AprilTags \cite{AprilTag}.

\subsection{Problem Setup:} We consider two manipulators, with one designated as the \emph{primary} manipulator and the other as the \emph{secondary}, each equipped with an end-effector mounted low-cost RGB camera. A set of objects is arranged on a tabletop within their shared workspace. 
The rigid transformations from each camera to its corresponding end–effector are unknown and must be estimated. To collect data, we command each manipulator through a sequence of $N$ distinct end–effector poses, capturing an RGB image at each pose. We denote the primary manipulator’s poses by $\{E_{1,1}, E_{1,2}, \cdots, E_{1,N}\}$ with $N$ corresponding RGB images $\{I_{1,1}, I_{1,2}, \cdots, I_{1,N}\}$. We denote the $N$ poses for the secondary manipulator as $\{E_{2,1}, E_{2,2}, \cdots, E_{2,N}\}$, with corresponding RGB images $\{I_{2,1}, I_{2,2}, \cdots, I_{2,N}\}$. 

Using only these end–effector poses and captured images, Bi‐JCR will recover all of the following:
The rigid transformation $T^{E_1}_{C_1}$ from Camera 1 to the primary end–effector; The rigid transformation $T^{E_2}_{C_2}$ from Camera 2 to the secondary end–effector; The scale factor $\lambda$ aligning the foundation model’s output frame with the real‐world metric frame; The pose of the secondary base frame $b_{2}$ relative to the primary base frame $b_{1}$, denoted $P^{b_{1}}_{b_{2}}$; The transformation $T^{b_1}_{w}$ from the foundation model’s output frame $w$ to the primary base frame $b_{1}$; A metric‐scale 3D reconstruction of the scene in the primary base frame $b_{1}$.

The transformations between the dual manipulators, along with their attached cameras, are illustrated in \cref{fig: diagram}. Here, we observe that only the forward kinematics of the manipulators, i.e. the transformation from the bases of the manipulators to their end-effector, is known. The relative position of the two manipulators are also initially unknown. Bi-JCR solves for all of the unknown transformations.

\begin{figure}[t]
\centering
\includegraphics[width=0.98\linewidth]{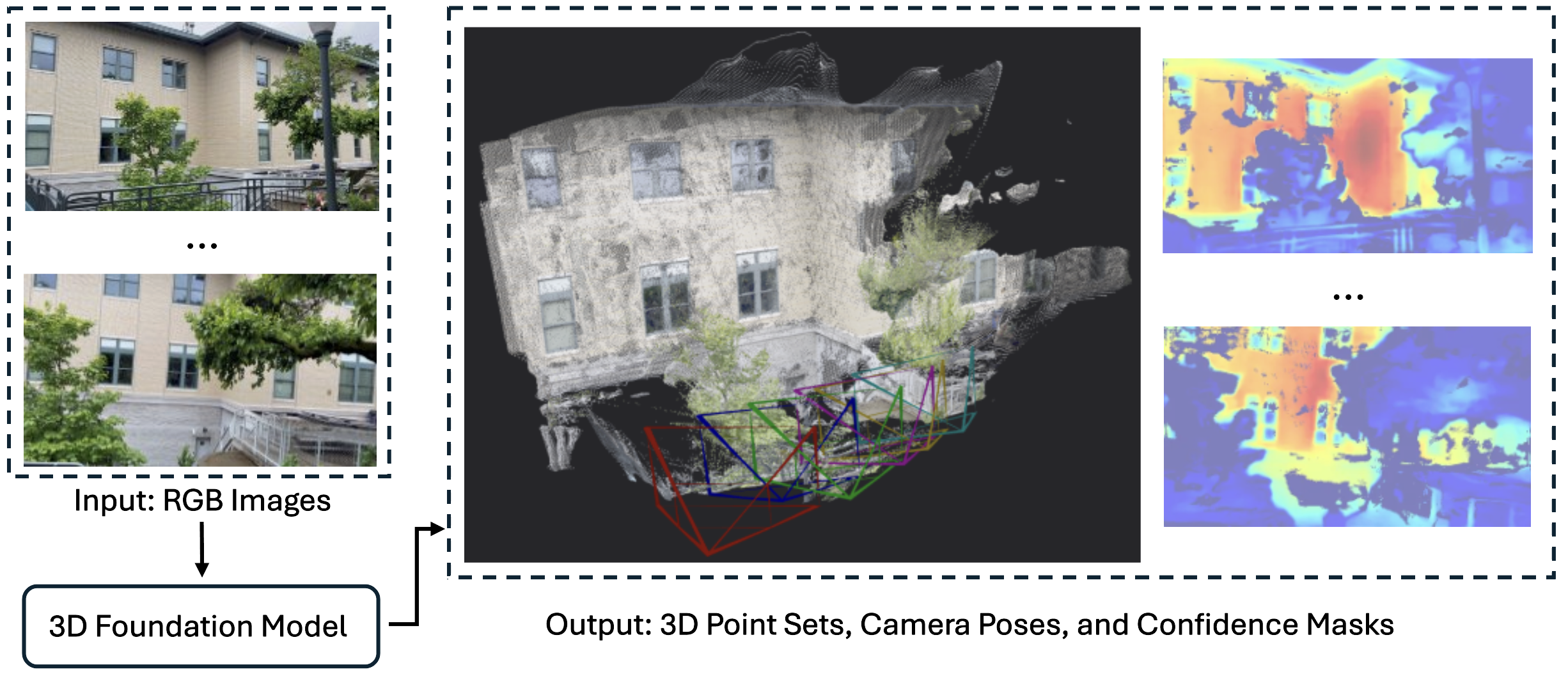}
\caption{3D foundation model taking a set of RGB images, and output 3D point sets, camera poses and confidence maps.}\label{fig: foundation_model}
\vspace{-1em}
\end{figure}

\subsection{3D Foundation Models in the Pipeline} 
The two image sets can be fed into a 3D foundation model to obtain a reconstruction of the scene in an arbitrary coordinate frame and scale. We recover both of the relative camera poses 
\begin{equation}
\{P_{1,1}, \dots, P_{1,N},\,P_{2,1}, \dots, P_{2,N}\}
\end{equation}
along with corresponding point sets containing the reconstruction,
\begin{equation}
\{\hat{X}_{1,1}, \dots, \hat{X}_{1,N},\,\hat{X}_{2,1}, \dots, \hat{X}_{2,N}\},
\end{equation}
where each point in a set corresponds to a pixel in the associated input image, and confidence values for each pixel can also be recovered. Pre-trained 3D foundation models are often used to extract structure from images of indoor scenes and building structures, and their application for table-top scenes has been under-explored. Example outputs from the model, DUSt3R \cite{DUSt3R_cvpr24}, is given in \cref{fig: foundation_model}. Because the foundation model recovers geometry only up to an unknown scale, both the estimated camera poses and the aggregated point cloud are not expressed in real‐world metric units. To resolve this scale ambiguity, we introduce a \emph{scale factor} $\lambda$ so that the pose of camera $i$ on manipulator $m$ in the real‐world (or base) frame $w$ becomes
\begin{align}
P^{w}_{m,i}(\lambda)
&=
\begin{bmatrix}
R_{m,i} & \lambda\,t_{m,i}\\
0 & 1
\end{bmatrix}
\;\in\;\mathrm{SE}(3),
\label{eq:scaled_c2w}
\end{align}
where we have $i = 1,\dots,N$ and the index, $m\in\{1,2\}$. In this expression, $R_{m,i}\in\mathrm{SO}(3)$ denotes the rotation and $t_{m,i}\in\mathbb{R}^{3}$ is the translation estimated by the foundation model. Next, defining the transform from the scaled foundation frame $w$ to each manipulator’s base frame $b_m$ as $T^{b_m}_{w}$, the camera poses in the real‐world base frames are
\begin{align}
P^{b_m}_{m,i}(\lambda)
&=
T^{b_m}_{w}\,P^{w}_{m,i}(\lambda),
\quad
i = 1,\dots,N.
\label{eq:scaled_c2b}
\end{align}


\subsection{Solving for Initial Calibration Solution}
In Bi-JCR, we seek to simultaneously solve for $\lambda$, $T^{b_1}_w$, and $T^{b_2}_w$ in the process of solving bi-manual hand-eye calibration for the two camera frame to base frame transformations $T^{E_1}_{C_1}$ and $T^{E_2}_{C_2}$. During the sequence of manipulator motions, the transformation between scaled camera poses and end effector poses for manipulator $m \in \{1, 2\}$ can be formulated as the classical hand-eye calibration equations \cite{Tsai1988ANT}:
\begin{align}
{E_{m, i}}^{-1} E_{m, i+1} T^{E_m}_{C_m} = T^{E_m}_{C_m}{P^{b_{m}}_{m, i}}^{-1}(\lambda)P^{b_{m}}_{m, i+1}(\lambda)\label{eq:axxb}
\end{align}
Now, we denote the transformation between poses by,
\begin{align}
T^{E_{m, i+1}}_{E_{m, i}} &= {E_{m, i}}^{-1} E_{m, i+1}, \\
T^{P^{w}_{m, i+1}}_{P^{w}_{m, i}}(\lambda) &= {P^{w}_{m, i}}^{-1}(\lambda)P^{w}_{m, i+1}(\lambda). \label{eq:pose_transforms}
\end{align}
Then, the hand-eye equations can be formulated as
\begin{align}
T^{E_{m, i+1}}_{E_{m, i}} T^{E_m}_{C_m} = T^{E_m}_{C_m} T^{P^{w}_{m, i+1}}_{P^{w}_{m, i}}(\lambda).
\label{eq:axxb_w}
\end{align}
Here we observe that \cref{eq:axxb_w} admits the $AX=XB$ form of the classical hand-eye calibration problem, with the right-hand side dependent on the scale factor $\lambda$.  

The first phase of Bi-JCR aims to obtain an initial solution for the scale and desired transformation, which we denote as $\lambda'$, ${T^{E_1}_{C_1}}'$ and ${T^{E_2}_{C_2}}'$. Since the rotation component here are in $\mathbf{SO}(3)$, we can first solve for the rotation components of the transformations, which are invariant to the scale factor. This can be achieved via linear algebra on the manifold of rotation matrices by following \cite{Park+Martin-1994}. We convert rotation components of $T^{E_{m, i+1}}_{E_{m, i}}$ and $T^{P^{w}_{m, i+1}}_{P^{w}_{m, i}}$ into the log map of $\mathbf{SO}(3)$ to its lie algebra ($\mathfrak{so}(3)$) where for some $R\in\mathbf{SO}(3)$. This gives,
\begin{align}
\omega=&\arccos(\frac{\mathrm{Tr}(R)-1}{2}), \\
\mathrm{LogMap}(R):=&\frac{\omega}{2\sin(\omega)}\begin{bmatrix}
R_{3,2}-R_{2,3}\\ R_{1,3}-R_{3,1} \\ R_{2,1}-R_{1,2}
\end{bmatrix} \in \mathfrak{so}(3).
\end{align}
where, $\mathrm{Tr}(\cdot)$ indicates the trace operator and the subscripts indicate the elements' indices in $R$. Then we can find best fit rotational components via:
\begin{align}
{R_{C_m}^{E_m}}^{\prime} &=(M_m^{\top}M_m)^{-\frac{1}{2}}M_m^{\top}, \label{eq:solve_R}\\
\text{where } M_m &=\sum_{i=1}^{N-1} \mathrm{LogMap}(R_{E_{m, i}}^{E_{m, i+1}})\otimes \mathrm{LogMap}(R^{P^{w}_{m, i+1}}_{P^{w}_{m, i}}), \nonumber
\end{align}
The $\otimes$ is the outer product, and the matrix inverse square root can be computed efficiently via singular value decomposition. 

Next, we solve the translation components along with the scale factor jointly, by minimizing the residuals of the similar scale recovery problem formulated in \cite{zhi2024unifying} for each arm, assuming that the scale factor is consistent for the results of each arm:
\begin{align}
\text{SRP:} \quad \arg\min_{{t_{C_m}^{E_m}}',\lambda'}&\sum^{N-1}_{i=1}\lvert\lvert {Q_{i}}{t_{C_m}^{E_m}}' - \bb{d}_{i}(\lambda')\lvert\lvert_{2}^{2}, \label{eq:srp} \\
\text{where }  {Q_{i}} &= I-R_{E_{m, i}}^{E_{m, i+1}},\\ 
\text{and } \bb{d}_{i}(\lambda')&=t_{E_{m, i}}^{E_{m, i+1}}-{R_{C_m}^{E_m}}'(\lambda) t^{P^{w}_{m, i+1}}_{P^{w}_{m, i}}.
\end{align}
\Cref{eq:srp} can be solved via least-squares, and we obtain our solutions $\lambda'$, ${T^{E_1}_{C_1}}'$ and ${T^{E_2}_{C_2}}'$, which can be further refined via gradient-based optimisation.

\subsection{Refine Calibration through Gradient-based Optimization}
We further refine the solutions via gradient descent to improve estimation. Here, we first rearrange \Cref{eq:axxb_w} into,
\begin{align}
T^{E_{m, i+1}}_{E_{m, i}} T^{E_m}_{C_m} - T^{E_m}_{C_m} T^{P^{w}_{m, i+1}}_{P^{w}_{m, i}} = 0, \label{eq:axxb_0}
\\ \text{ for } i\in \{1,\cdots, N-1\}, m\in \{1, 2\}.\nonumber
\end{align}
then we can solve the calibration problem by minimizing the difference between transformation matrices $T^{E_{m, i+1}}_{E_{m, i}} T^{E_m}_{C_m}$ and $T^{E_m}_{C_m} T^{P^{w}_{m, i+1}}_{P^{w}_{m, i}}$. Specifically, we define a cost function as,
\begin{align}
\ell&(\lambda, T^{E_1}_{C_1}, T^{E_2}_{C_2})= \\
&\sum_{m\in \{1, 2\}}\bigl(\frac{1}{N-1}\sum_{i = 1}^{N-1} \alpha 
 D_R(T^{E_m}_{C_m})\!+\!(1\!-\!\alpha)D_t(\lambda, T^{E_m}_{C_m})\bigr), \nonumber\\
&\text{where } D_R(T^{E_m}_{C_m}) \!=\!\arccos \bigl(\operatorname{tr}({RE_{m, i}^{m, i+1}}^{\top}\!RC_{m, i}^{m, i+1})\!-1\!\bigr), \nonumber\\
& \text{and }D_t(\lambda, T^{E_m}_{C_m})\!=\!\bigl\lVert te_{m, i}^{m, i+1}\!-\!tc_{m, i}^{m, i+1} \bigr\rVert_{2},\nonumber
\end{align}
where $\operatorname{tr}(\cdot)$ is the trace operator. We can then minimize the cost function via gradient descent \cite{bishop2006pattern} while constraining the rotation to be on the $\mathbf{SO}(3)$ manifold. Here, we use the results from the previous section as the initial solution. Furthermore, to ensure that the rotational components in $\mathbf{SO}(3)$ during the backpropagation process, we follow \cite{bregier2021deepregression} and first pull each rotation into the Lie algebra with the logarithm map, then perform the gradient update on the resulting axis–angle vector in $\mathbb{R}^3$, and push it back onto the manifold via the exponential map. With the solutions that minimize the cost function, we can then obtain the world-to-base transformations $T^{b_1}_w$, and $T^{b_2}_w$ via
\begin{align}
T^{b_m}_w = \operatorname*{AVG_{SE3}}\limits_{i\in \{1, \cdots, N-1\}} \bigl(T^{E_{m, i+1}}_{E_{m, i}} {T^{E_m}_{C_m}T^{P^{w}_{m, i+1}}_{P^{w}_{m, i}}}^{-1} \bigr)
\label{eq:w2b_avg}
\end{align}
where $\operatorname*{AVG_{SE3}}$ is the average over a set of transformation matrices on $\mathbf{SE}(3)$, by considering the average of rotation and translation separately.

\subsection{Obtaining Metric-Scale 3D Representation}
Here, we seek to build a real-world metric scale 3D representation of the environment under the primary manipulator's frame.
Following \Cref{eq:scaled_c2w}, we first scale camera poses by $\lambda$ to get $\{P^{w}_{1,1}, \cdots, P^{w}_{1,N}\}$ and $\{P^{w}_{2,1}, \cdots, P^{w}_{2,N}\}$, we can then get the calibrated metric scale camera poses by
\begin{align}
P^{b_1}_{m,i} = T^{b_1}_w P^w_{m,i}, \text{ for } m\in\{1, 2\}, i\in\{1, \cdots, N\}.
\label{eq:w2b_avg}
\end{align}

Next, we also want to use the point sets from each arm, associated with each input image, $\{\hat{X}_{1,1}, \cdots, \hat{X}_{1,N}\}$ and $\{\hat{X}_{2,1}, \cdots, \hat{X}_{2,N}\}$ and their associated confidence maps $\{C_{1,1}, \cdots, C_{1,N}\}$ and $\{C_{2,1}, \cdots, C_{2,N}\}$ from the output of the foundation model to recover a rich and high-quality representation of the environment. We first use a confidence threshold to filter out points below this threshold in each $\hat{X}_{m, i}$. Then, we transform the points from the filtered point sets, $\{\bb{x}_{i}\}_{i=1}^{N_{pc}}$, to get a point cloud in real-world metric scale and primary manipulator's base frame, $\{\bb{x}^{b_1}_{i}\}_{i=1}^{N_{pc}}$ through
 \begin{align}
\{ \bb{x}^{b_1}_{i} = T^{b_1}_w (\lambda \bb{x}_{i}), \text{ for } i \in \{1, \cdots, N_{pc}\}\}.
\label{eq:points}
\end{align}

Furthermore, the pose of the secondary manipulator's base in the primary manipulator's base frame can be computed as
\begin{align}
P^{b_1}_{b_2} = T^{b_1}_w {T^{b_2}_w}^{-1}.
\label{eq:base2base}
\end{align}
The pose of the secondary manipulator's base in the primary manipulator's base frame enables us to compute end-effector poses for downstream tasks of both manipulators in a single unified frame. This facilitates downstream processes, such as object segmentation along with grasping generation, to operate.



\begin{table*}[t]
\begin{adjustbox}{width=\textwidth,center}
\begin{tabular}{@{}ll|rrr|rrr|rrr@{}}
\toprule[1pt]\midrule[0.3pt]
\multicolumn{1}{l}{} & \multicolumn{1}{l}{} & \multicolumn{1}{r}{} & \multicolumn{1}{l}{Scene A} & \multicolumn{1}{l}{} & \multicolumn{1}{r}{} & \multicolumn{1}{l}{Scene B} & \multicolumn{1}{l}{} & \multicolumn{1}{r}{} & \multicolumn{1}{l}{Scene C} & \multicolumn{1}{l}{} \\
\midrule
      & Images Per Manipulator   & 4 images  & 7 images & 9 images & 4 images & 7 images & 9 images & 4 images & 7 images & 9 images \\ \midrule 
 & Converged & $\bm{\checkmark}$ & $\bm{\checkmark}$ & $\bm{\checkmark}$ 
                  & $\bm{\checkmark}$ & $\bm{\checkmark}$ & $\bm{\checkmark}$ 
                  & $\bm{\checkmark}$ & $\bm{\checkmark}$ & $\bm{\checkmark}$ \\
Bi-JCR (Ours) & Residual $\delta_R$ & 0.0769 & 0.0724 & 0.0668 & 0.0740 & 0.0617 & 0.0569 & 0.0743 & 0.0634 & 0.0612 \\ 
          & Residual $\delta_\bb{t}$      & 0.0461 & 0.0391 & 0.0378 & 0.0587 & 0.0351 & 0.0340 & 0.0424 & 0.0341 & 0.0373 \\
          & No. of Poses (Left)             & 4     & 7     & 9     & 4     & 7     & 9     & 4     & 7     & 9 \\ 
      & No. of Poses (Right)             & 4     & 7     & 9     & 4     & 7     & 9    & 4     & 7     & 9 \\ \midrule

      & Converged & \textbf{\texttimes} & \textbf{\texttimes} & \textbf{\texttimes} 
                  & $\bm{\checkmark}$   & $\bm{\checkmark}$   & $\bm{\checkmark}$ 
                  & \textbf{\texttimes} & $\bm{\checkmark}$   & $\bm{\checkmark}$ \\
COLMAP \cite{schoenberger2016sfm} & Residual $\delta_R$ & NA & NA & NA & 0.0865 & 0.0583 & 0.0584 & NA & 0.0591 & 0.0586 \\
      + Calibration               & Residual $\delta_\bb{t}$      & NA & NA & NA & 0.0317 & 0.0269 & 0.0274 & NA & 0.0277 & 0.0271 \\
       & No. of Poses (Left)             & 0     & 4     & 2     & 4     & 7     & 9     & 0     & 7     & 9 \\ 
      & No. of Poses (Right)             & 4     & 0     & 0     & 4     & 7     & 9    & 4     & 7     & 9 \\ \midrule

Ray Diffusion \cite{zhang2024raydiffusion} & Residual $\delta_R$ & 0.4845 & 0.4288 & 0.7951 & 0.4948 & 0.3839 & 0.3634 & 0.4504 & 0.2610 & 0.2223 \\ 
      + Calibration               & Residual $\delta_\bb{t}$      & 0.2118 & 0.1416 & 0.1309 & 0.2067 & 0.2051 & 0.1637 & 0.2125 & 0.1858 & 0.1760 \\ 
      & No. of Poses (Left)            & 4     & 7     & 9     & 4     & 7     & 9     & 4     & 7     & 9 \\ 
      & No. of Poses (Right)            & 4     & 7     & 9     & 4     & 7     & 9    & 4     & 7     & 9 \\
\midrule[0.3pt]\bottomrule[1pt]
\end{tabular}
\end{adjustbox}
\caption{Quantitative evaluation on Bi-JCR's calibration residual error ($\delta_R$ and $\delta_t$) against baseline methods. Lower residual indicates more accurate calibrations.}
\label{table:res_cali}
\vspace{-2em}
\end{table*}

\section{Empirical Evaluation}
In this section, we rigorously evaluate our proposed Bi-Manual Joint Calibration and Representation (Bi-JCR) method. Our bi-manual setup consists of two AgileX Piper 6 degree-of-freedom manipulators, each with a low-cost USB webcam mounted on the gripper. We seek to answer the following questions:
\begin{enumerate}
    \item Can Bi-JCR produce correct hand-eye calibration for both arms, even when the number of images provided is low?
    \item Can Bi-JCR recover the scale accurately such that our representation's sizes match the physical world?
    \item Can high-quality 3D environment representations, in the correct coordinate frame, be built? 
    \item How do different 3D foundation models change the calibration accuracy?
    \item Does refinement via additional gradient descent improve calibration accuracy?
    \item Does Bi-JCR facilitate downstream bi-manual manipulation tasks?
\end{enumerate}

\subsection{Eyes-to-Hands Calibration with Bi-JCR}
\textbf{Baselines: COLMAP-based pose estimation and Ray Diffusion.} To assess the calibration quality of Bi-JCR, we compare against two alternatives. First, in the absence of high-contrast markers such as checkerboards or AprilTags \cite{AprilTag}, we use SfM via COLMAP \cite{schoenberger2016sfm} and apply the Park–Martin algorithm \cite{Park+Martin-1994} to compute eye-to-hand transformations for each manipulator. Second, we evaluate Ray Diffusion \cite{zhang2024raydiffusion}, a sparse-view diffusion model trained on large datasets \cite{reizenstein21co3d} that directly regresses camera poses.

\textbf{Task and Metrics:} We take images in three different environments: a scene on a darker light condition with 9 items (scene A), two others of which are under brighter light conditions with different sets of objects of 9 and 10 items respectively (scene B, scene C). Bi-JCR and the two baseline methods are evaluated with an increasing number of input images (4, 7 and 9 images per manipulator), then check whether the calibration has converged correctly by considering residual losses via \Cref{eq:axxb_0} with ground truth, obtained via Apriltags \cite{AprilTag}, on the right-hand side. Lower residual values indicate a higher degree of consistency.

\textbf{Results:} We tabulate our results in \Cref{table:res_cali}. We observe that COLMAP often results in diverged calibration, as images where correspondence cannot be found are discarded. Whether the calibration has converged and the number of images used, from each manipulator, are also shown in \Cref{table:res_cali}. 
Although Ray Diffusion registered all images, it registered them in an inconsistent way under this tabletop setup of cluttered, partially visible objects, causing the calibration optimizer to accumulate large errors. Our Bi-JCR method consistently produce smaller residual under both lower and higher number of views for both manipulators, showing remarkable image efficiency. We also observe a residual loss reduce trend as the number of images gradually increase, in comparison to the residual loss fluctuation in the other two baseline methods, which shows Bi-JCR's reliable precision gain with increasing number of views.


\begin{figure}[t]
  \centering
  \begin{subfigure}[b]{0.165\linewidth}
    \centering
    \includegraphics[width=\linewidth]{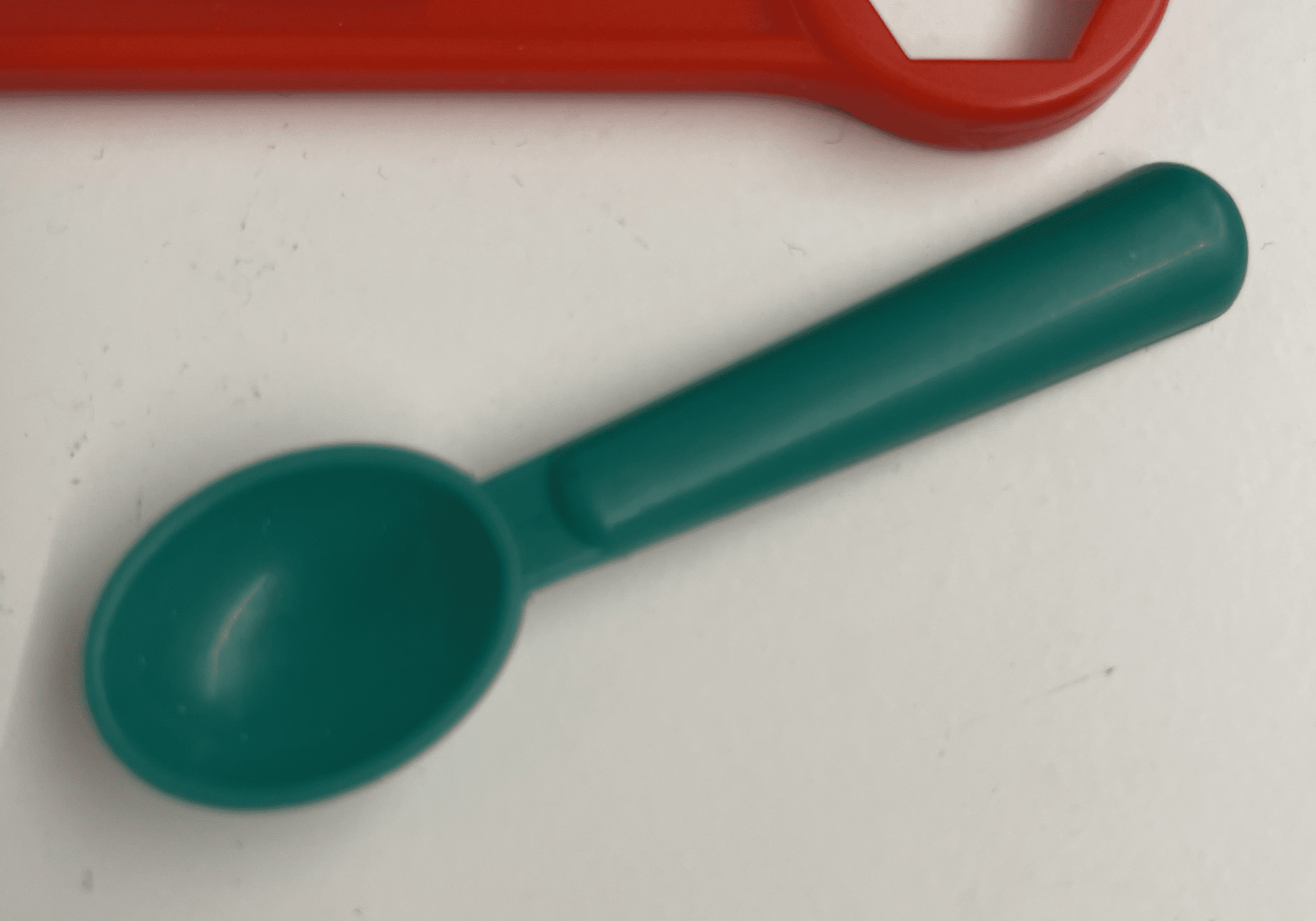}
    \includegraphics[width=\linewidth]{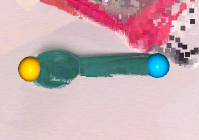}
    \caption{Spoon}
  \end{subfigure}%
  \begin{subfigure}[b]{0.165\linewidth}
    \centering
    \includegraphics[width=\linewidth]{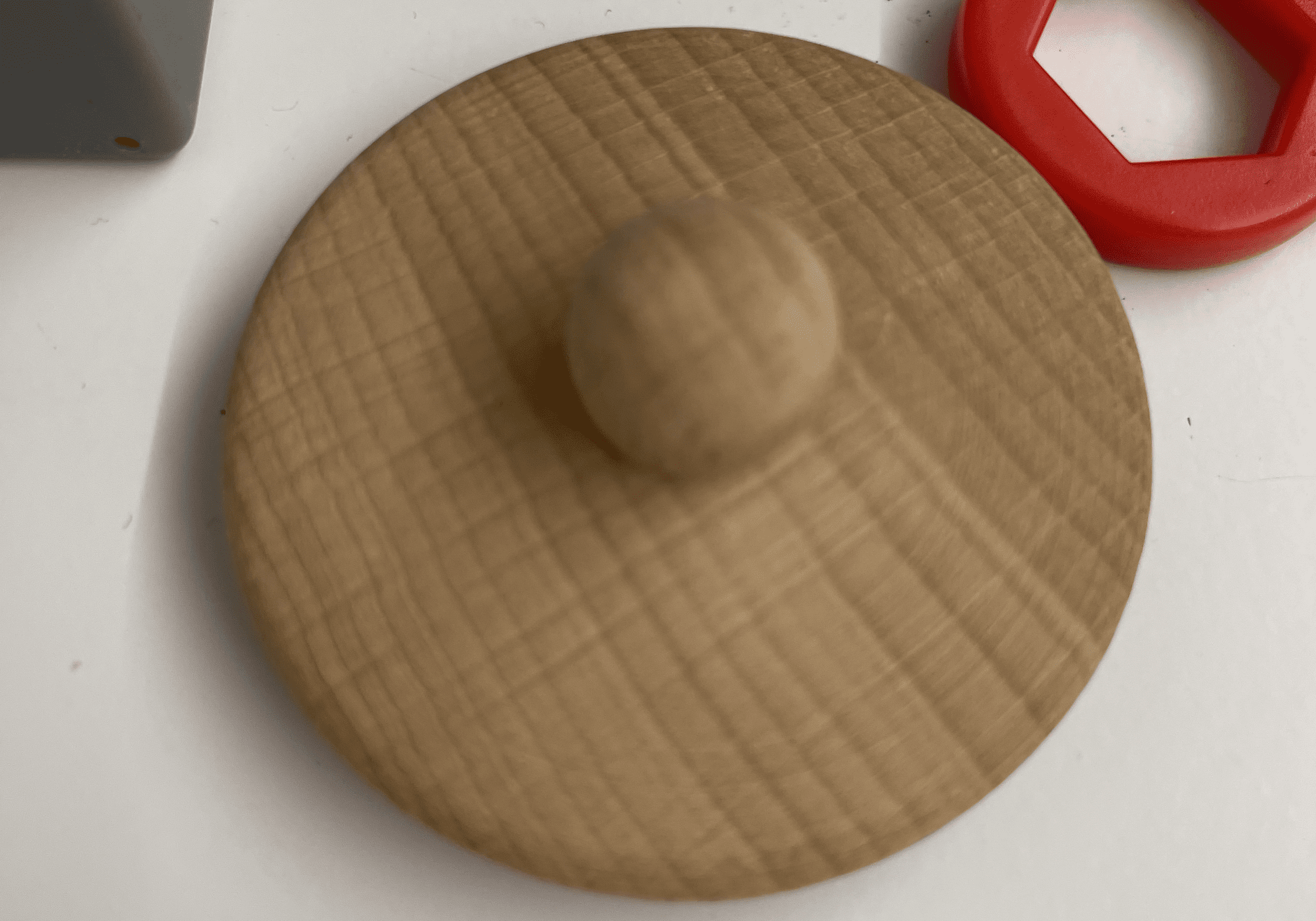}
    \includegraphics[width=\linewidth]{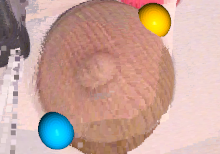}
    \caption{Tea Lid}
  \end{subfigure}%
  \begin{subfigure}[b]{0.165\linewidth}
    \centering
    \includegraphics[width=\linewidth]{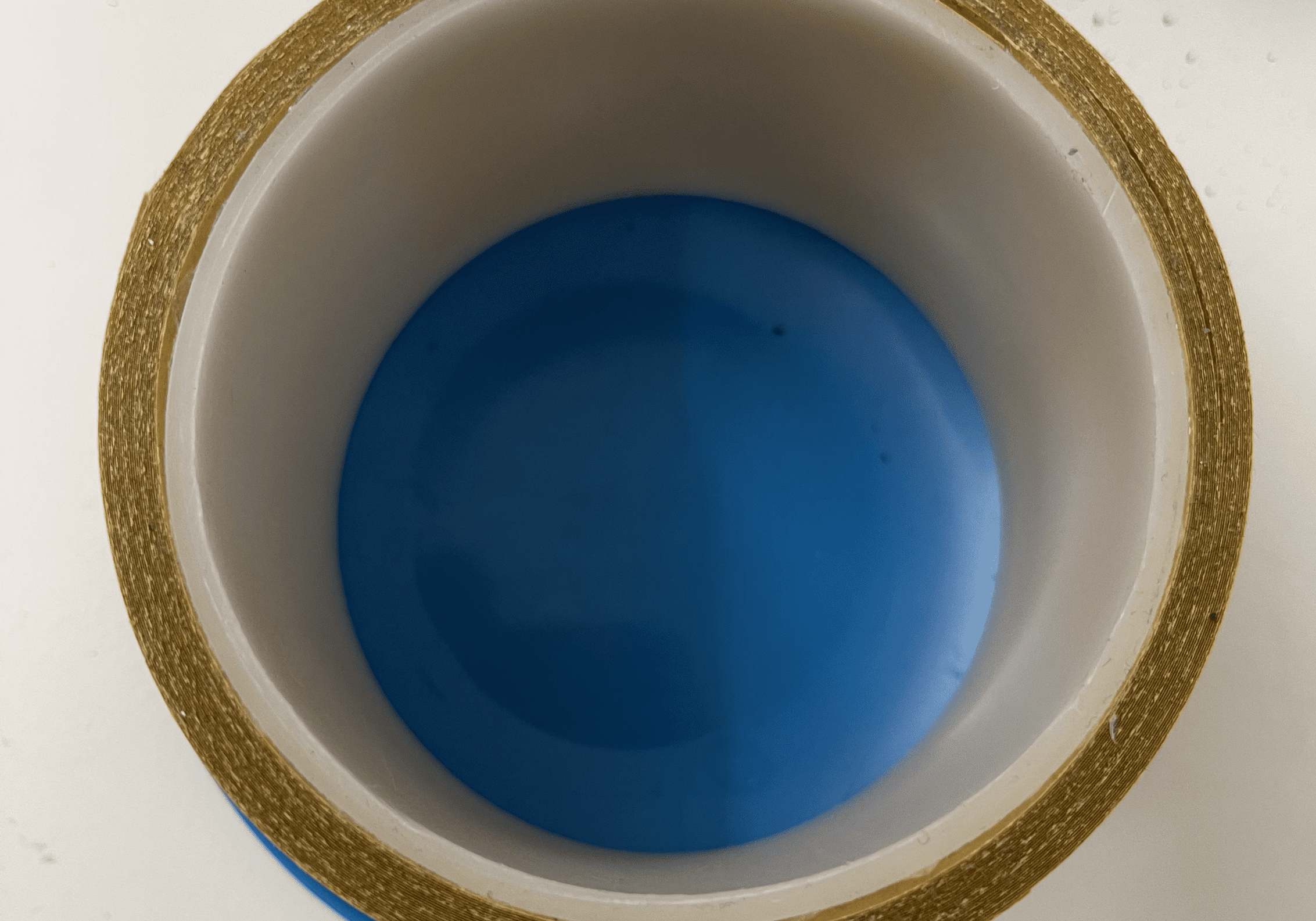}
    \includegraphics[width=\linewidth]{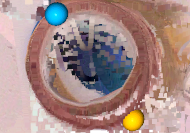}
    \caption{Tape}
  \end{subfigure}%
  \begin{subfigure}[b]{0.165\linewidth}
    \centering
    \includegraphics[width=\linewidth]{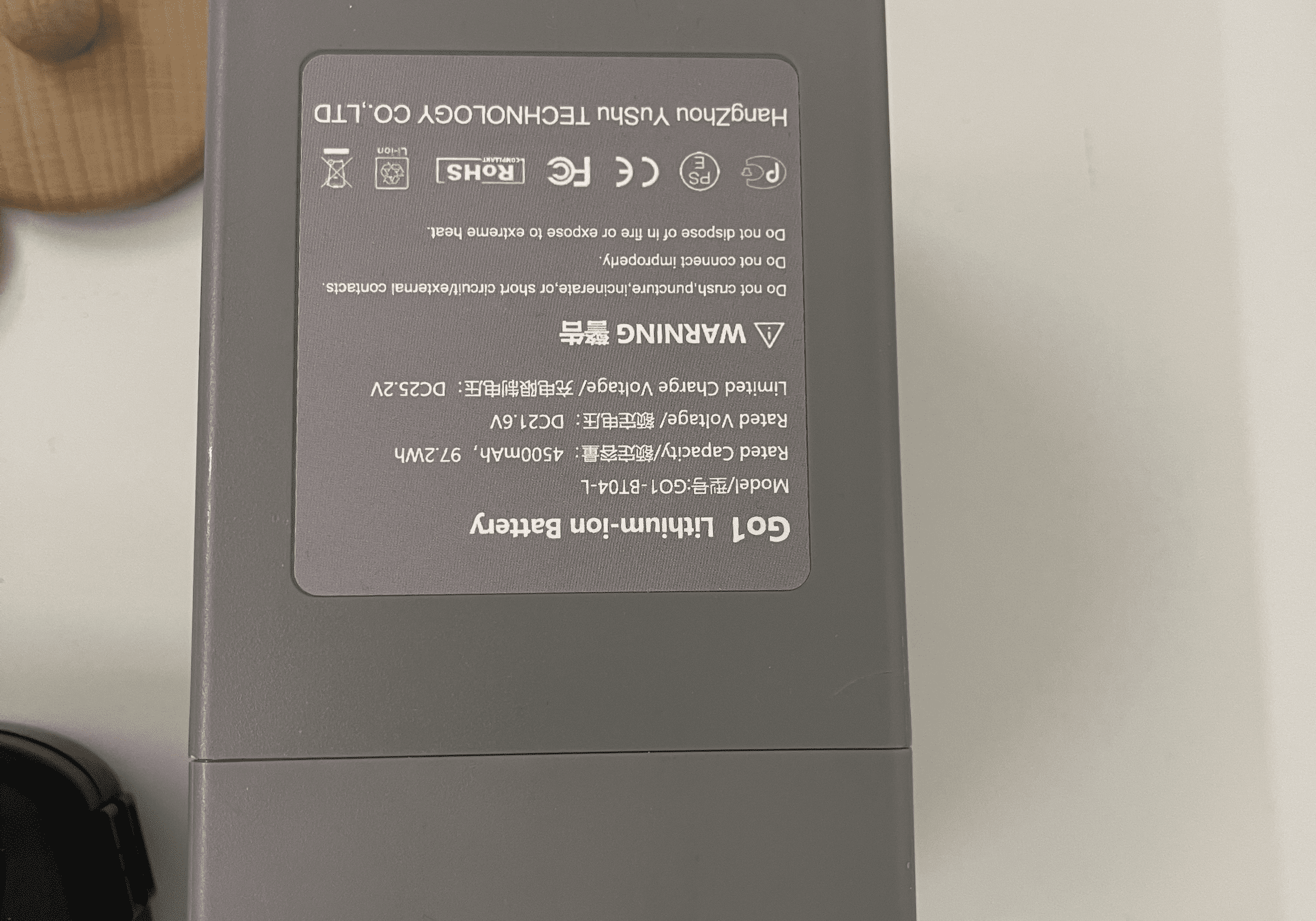}
    \includegraphics[width=\linewidth]{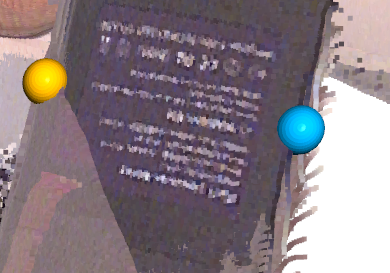}
    \caption{Battery}
  \end{subfigure}%
  \begin{subfigure}[b]{0.165\linewidth}
    \centering
    \includegraphics[width=\linewidth]{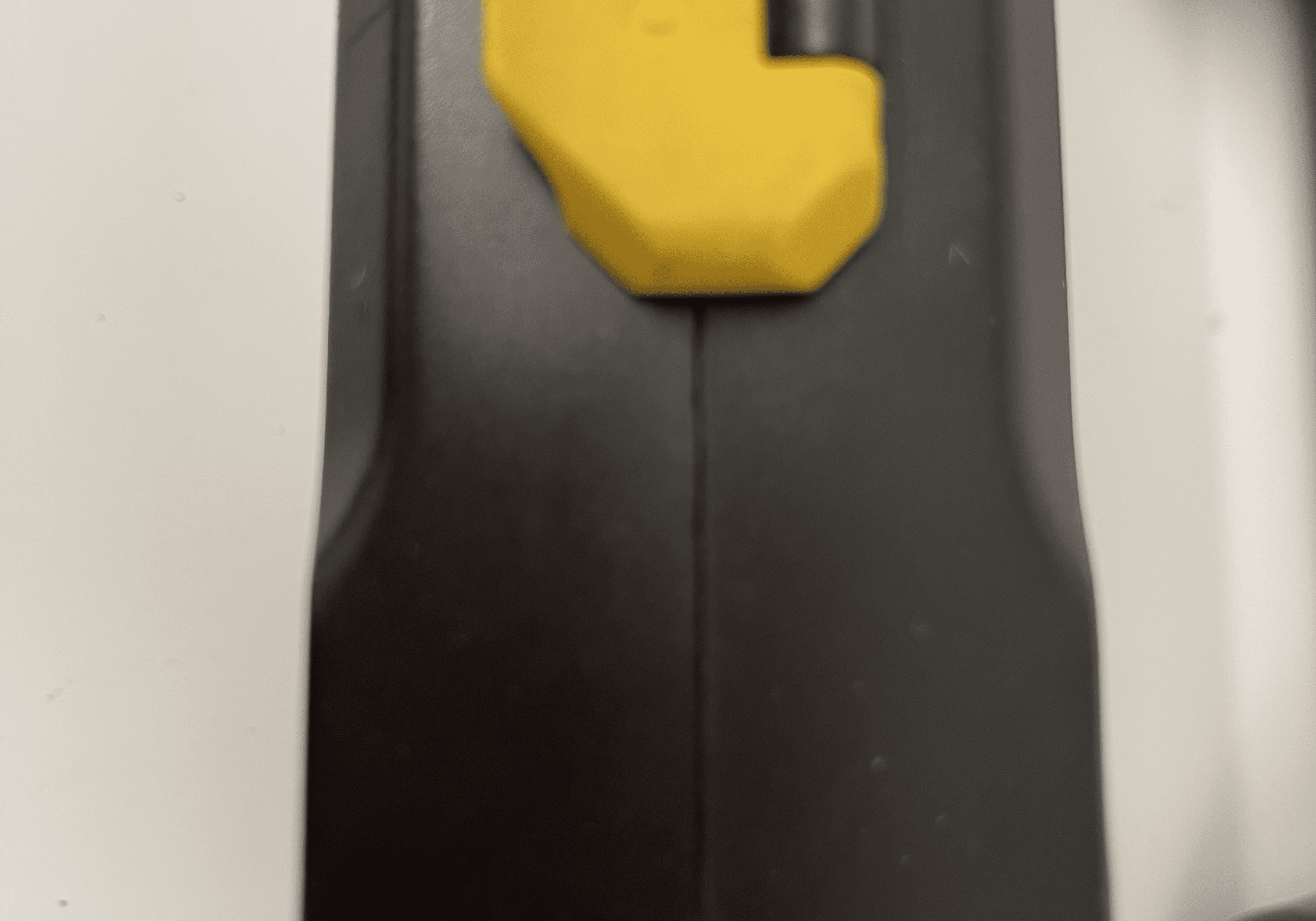}
    \includegraphics[width=\linewidth]{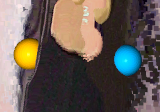}
    \caption{Toolbox}
  \end{subfigure}%
  \begin{subfigure}[b]{0.165\linewidth}
    \centering
    \includegraphics[width=\linewidth]{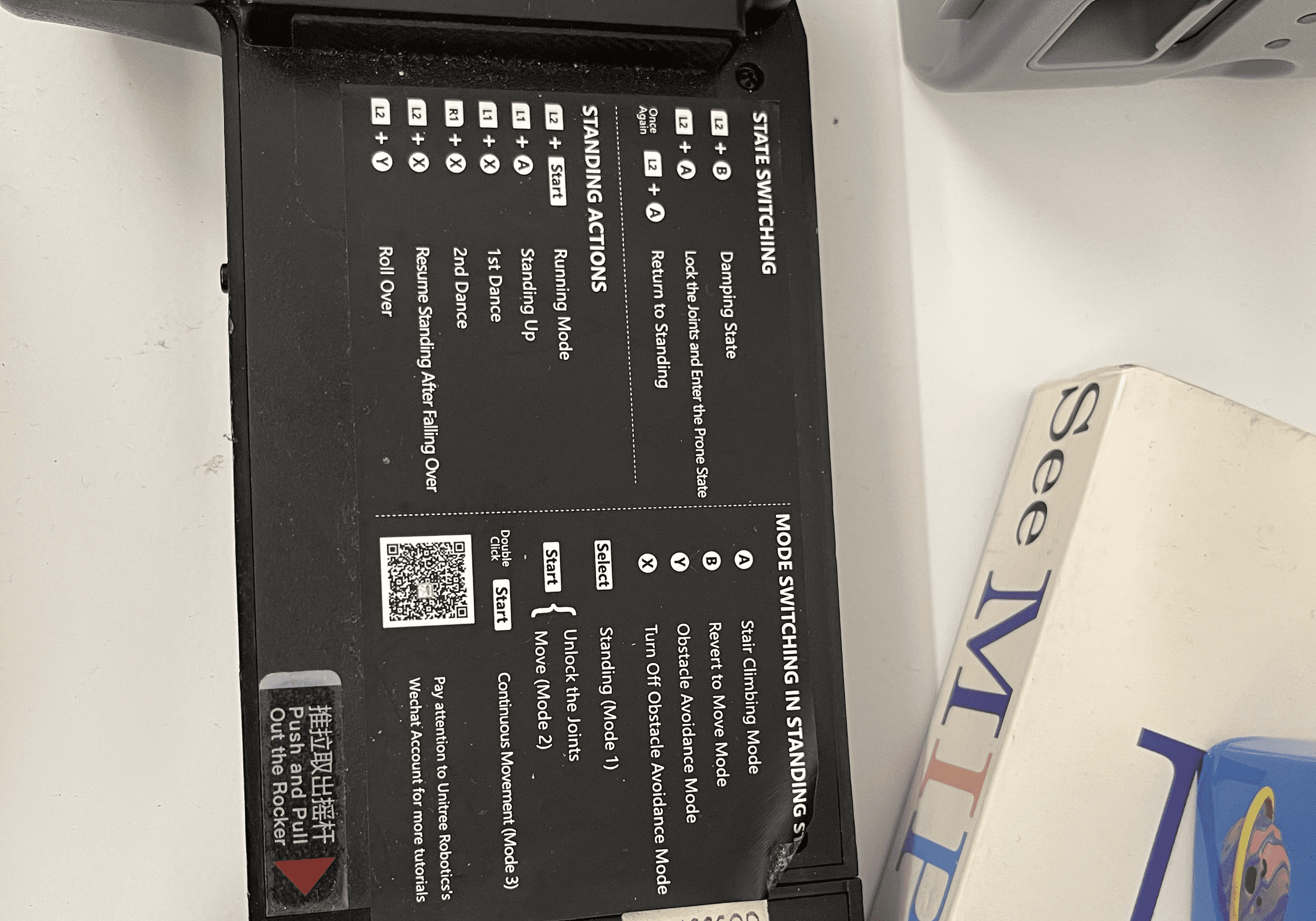}
    \includegraphics[width=\linewidth]{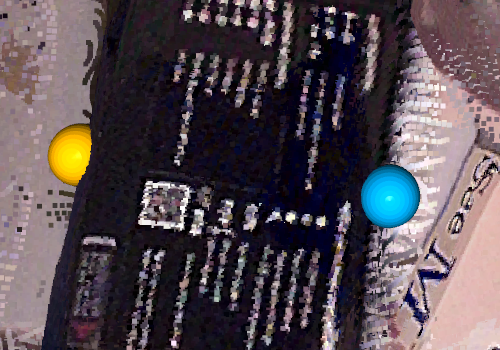}
    \caption{Joystick}
  \end{subfigure}
  \caption{Visualization of the objects we used for scale validation in \Cref{table:scale_validation}. For each object, the top is their real-world appearances, and the bottom is the reconstructions. The blue and yellow dots specify the length measured for scale validation.}
\label{fig:object_measure}
\vspace{-1em}
\end{figure}

\begin{table}[t]
\centering
\begin{adjustbox}{width=\linewidth,center}
\begin{tabular}{l|rrrrrr}
\toprule[1pt]\midrule[0.3pt]
Object      & Spoon   & Tea Lid   & Tape     & Battery & Toolbox & Joystick\\ \hline
5 Images & 4.90\%  & 4.3\%     & 2.73\%   & 1.59\%  & 10.90\% & 7.73\%  \\
8 Images & 1.61\%  & 0.26\%    & 1.34\%   & 1.10\%  & 2.53\%  & 2.98\%  \\ 
\midrule[0.3pt]\bottomrule[1pt]
\end{tabular}
\end{adjustbox}
\caption{Percentage of error of object dimensions to compare the size of real-world objects against reconstructed objects. Within 8 images per manipulator, the percentage errors of reconstructed sizes are at most 2.98\%, indicating accurate scale recovery.}
\label{table:scale_validation}
\vspace{-1.5em}
\end{table}

\begin{figure}[t]
  \centering
    \fbox{\includegraphics[width=.32\linewidth]{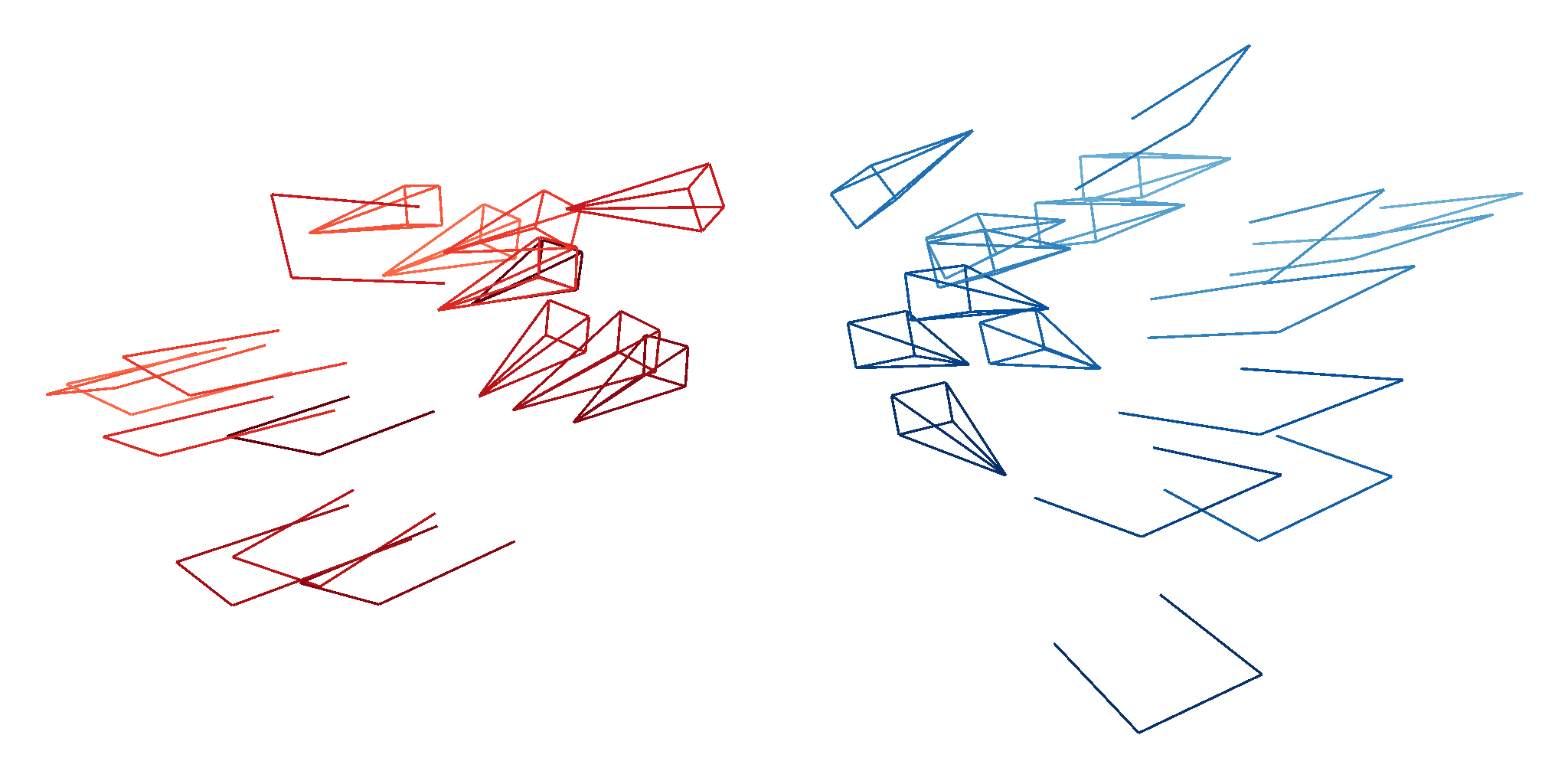}}%
    \fbox{\includegraphics[width=.32\linewidth]{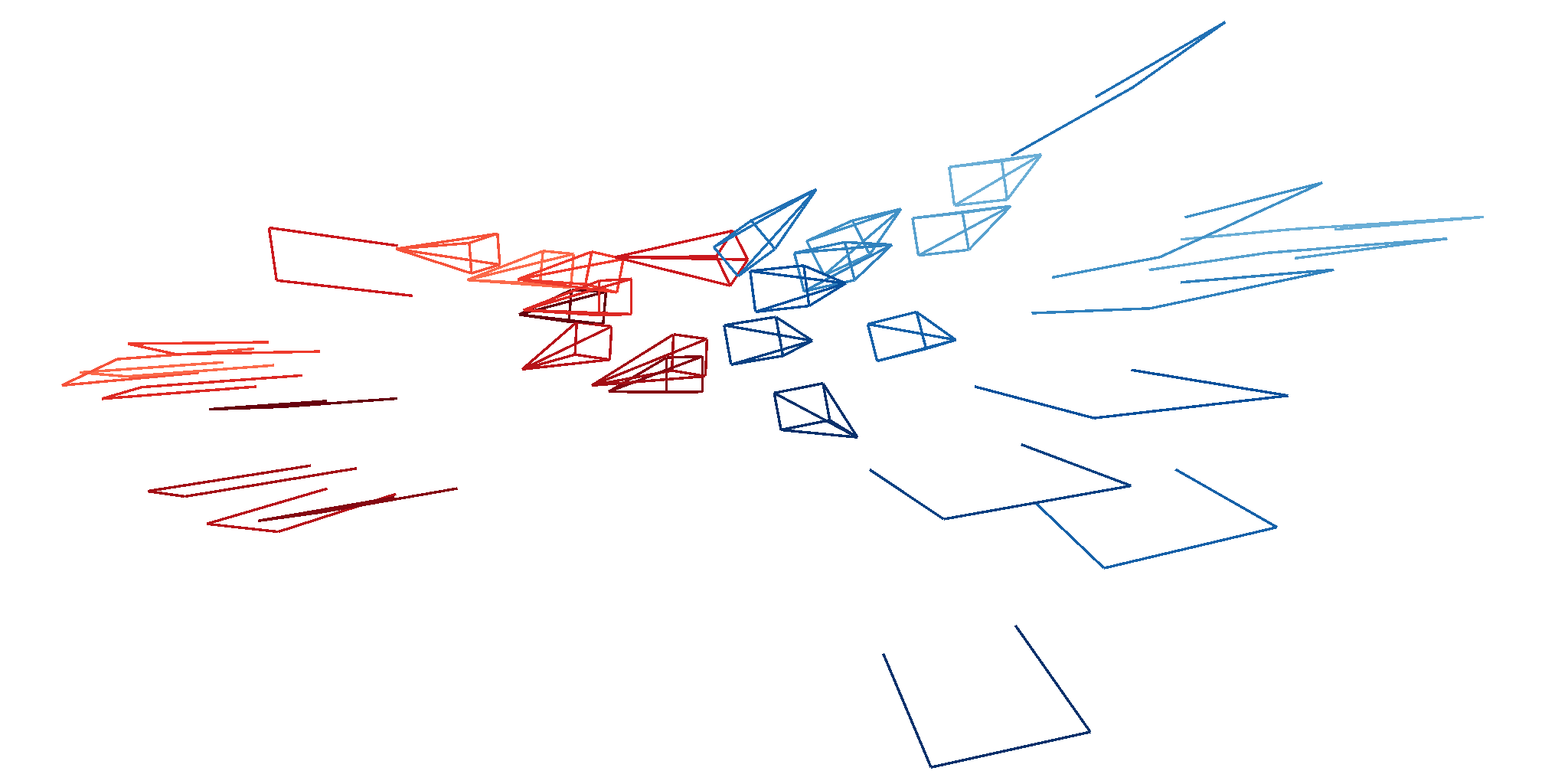}}%
    \fbox{\includegraphics[width=.32\linewidth]{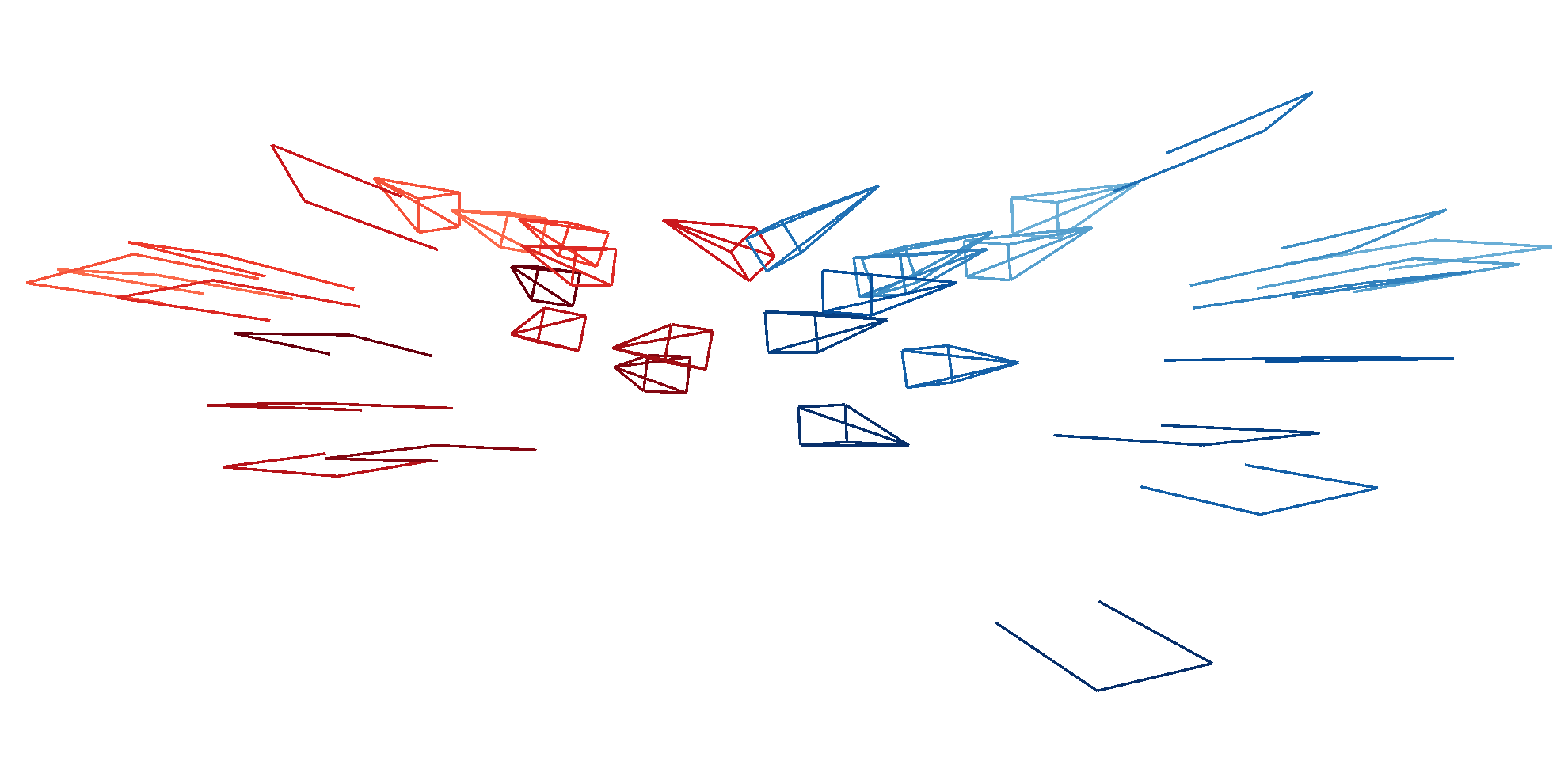}}
\caption{Qualitative evaluation of the camera calibration in the three scenes. We observe that the cameras indicated as cones are aligned with the end-effector poses, indicating accurate calibration. The end-effector and camera poses of the left manipulator are colored in shades of red and those of the right in blue.}
  \label{fig:calibration_viz}
  \vspace{-1.em}
\end{figure}

\begin{figure*}[t]
  \centering
  \begin{subfigure}[b]{0.105\linewidth}
    \centering
    \fbox{\includegraphics[width=\linewidth]{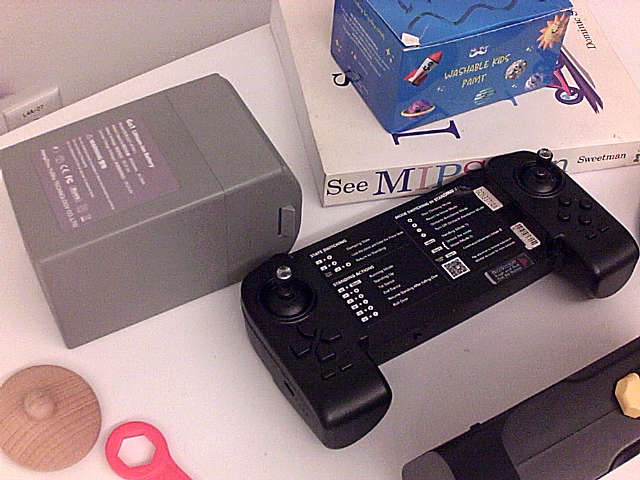}}
    
    \fbox{\includegraphics[width=\linewidth]{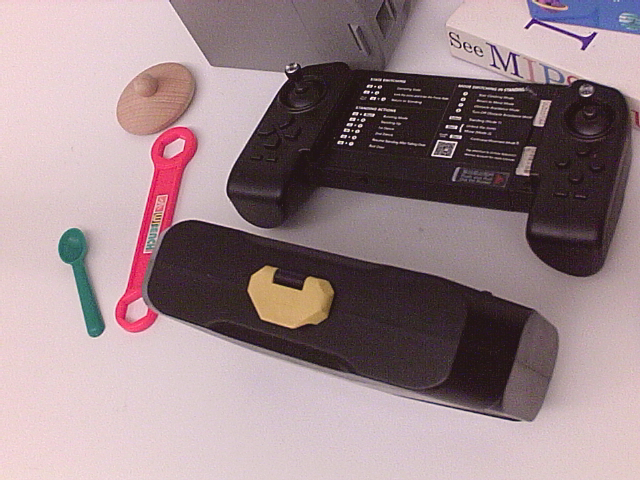}}
  \end{subfigure}
  \begin{subfigure}[b]{0.215\linewidth}
    \centering
    \fbox{\includegraphics[width=\linewidth]{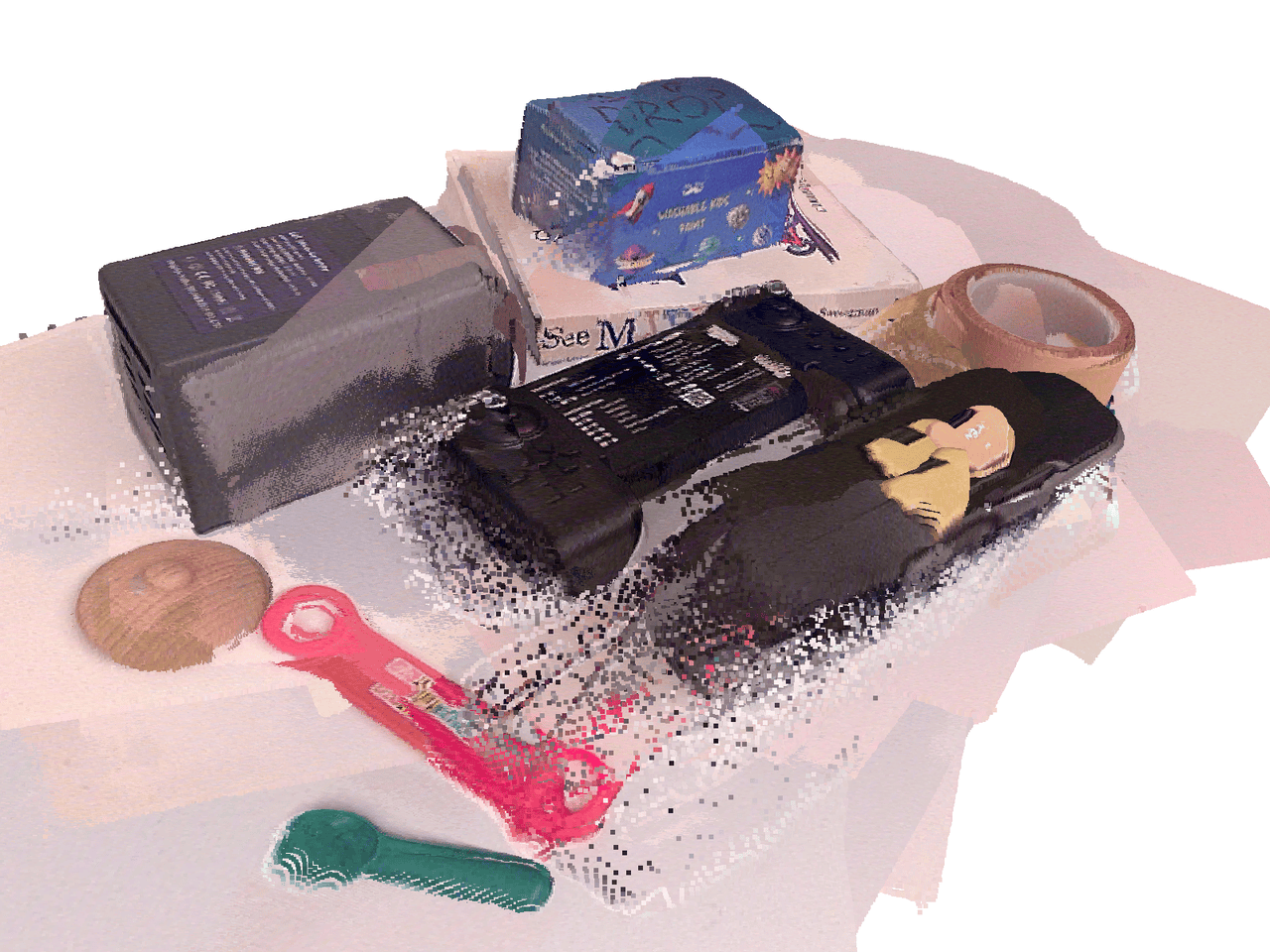}}
  \end{subfigure}
   \hfill
  \begin{subfigure}[b]{0.105\linewidth}
    \centering
    \fbox{\includegraphics[width=\linewidth]{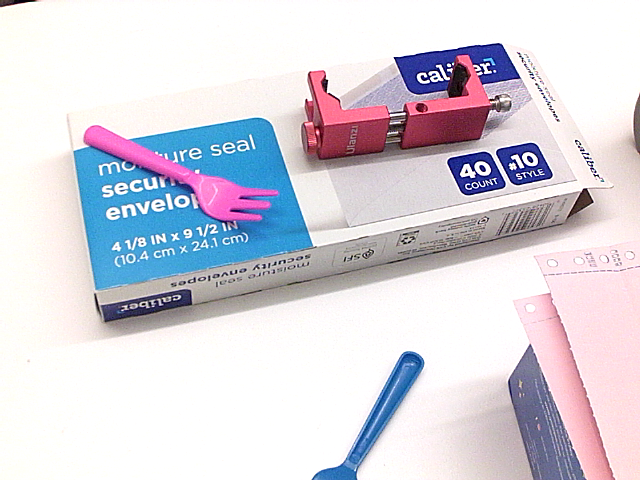}}
    
    \fbox{\includegraphics[width=\linewidth]{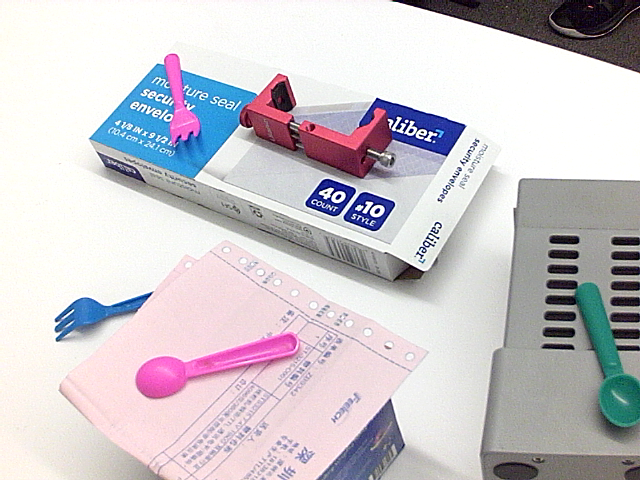}}
  \end{subfigure}
  \begin{subfigure}[b]{0.215\linewidth}
    \centering
    \fbox{\includegraphics[width=\linewidth]{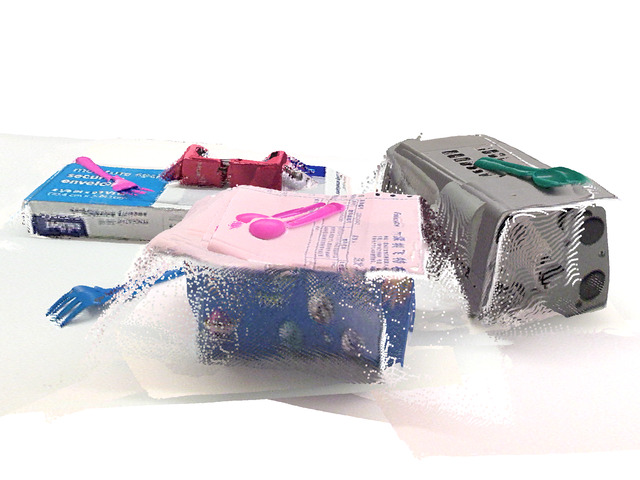}}
  \end{subfigure}
  \hfill
    \begin{subfigure}[b]{0.105\linewidth}
    \centering
    \fbox{\includegraphics[width=\linewidth]{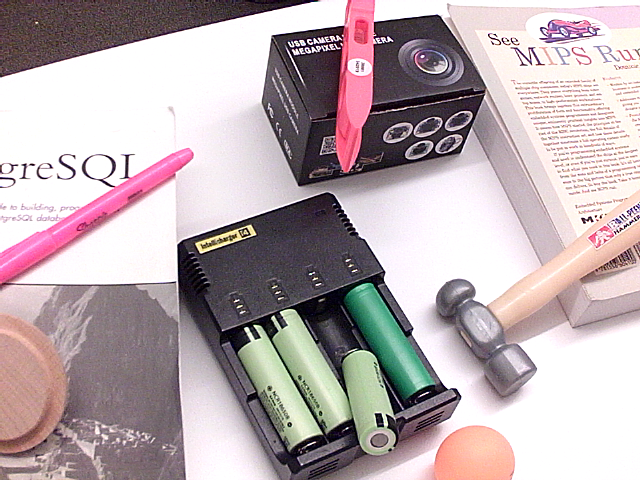}}
    
    \fbox{\includegraphics[width=\linewidth]{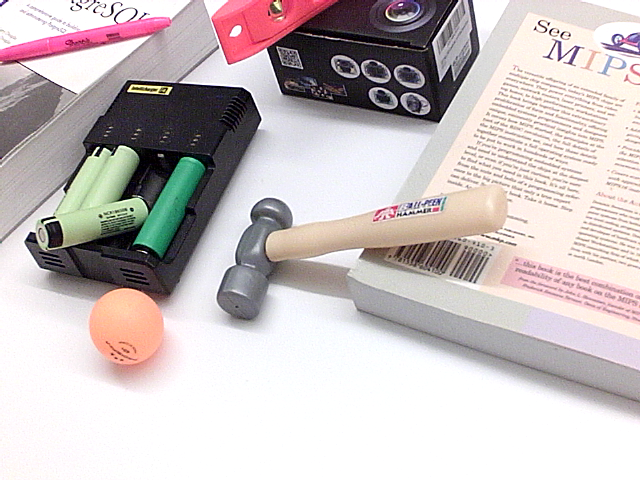}}
  \end{subfigure}
  \begin{subfigure}[b]{0.215\linewidth}
    \centering
    \fbox{\includegraphics[width=\linewidth]{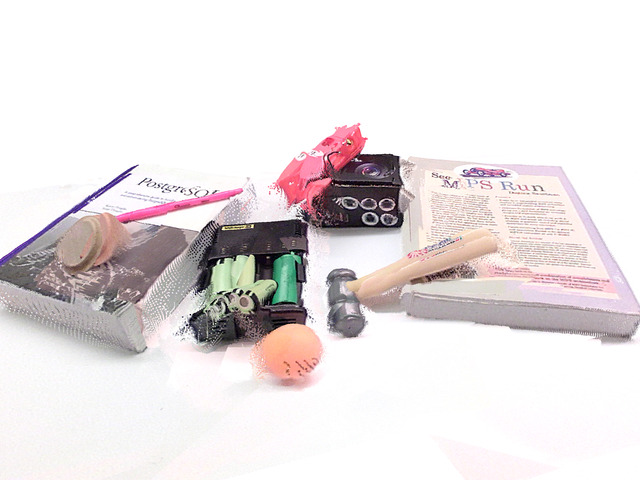}}
  \end{subfigure}
  \caption{Qualitative evaluations of the recovered 3D reconstructions, including scene A, scene B, scene C from left to right. The reconstruction of the scene is dense and geometrically accurate.}
  \label{fig:reconstruction_quality}
  \vspace{-1.em}
\end{figure*}

Here, we also visualize the aligned camera and end-effector poses after calibration via Bi-JCR in \Cref{fig:calibration_viz}. The end-effectors and outlined as U-shapes and cameras are represented by cones. Both end-effector poses and camera poses are transformed to the primary manipulator's base frame using the base to base transformation estimated by Bi-JCR. Primary manipulator end-effector and eye-to-hand transformed camera are colored in red, and the secondary manipulator's are colored in blue. As shown in \Cref{fig:calibration_viz}, Bi-JCR successfully recovers eye-to-hand transformations that consistently align camera and end-effector poses for both primary and secondary manipulator across all scenes.

\subsection{Accurate Metric Scale Recovery with Bi-JCR}
Bi-JCR also reconstructs a 3D dense point cloud on a metric scale of the real-world environment. Here, we evaluate the accuracy of scale recovery by comparing the difference between the side length of the real-world object and the side length of the reconstructed objects with 5 and 8 images collected from each manipulator, as shown in \Cref{fig:object_measure}. The error, computed by
\begin{align}
\operatorname{err}_{obj} = \frac{|s_{\text{reconstructed}}-s_{\text{real world}}|}{s_{\text{real world}}},
\label{eq:w2b_avg}
\end{align}
is reported in \Cref{table:scale_validation}. With only 8 images per manipulator, Bi-JCR is able to reduce the error to a median of 1.48\% and at most 2.98\%, which marks precise scale recovery giving real-world metric scale.

\begin{table*}[t]
\begin{adjustbox}{width=0.95\textwidth,center}
\begin{tabular}{ll|rrr|rrr|rrr}
\toprule[1pt]\midrule[0.3pt]
\multicolumn{2}{l|}{}                                                     & \multicolumn{3}{l|}{Darker Light Condition (9 items)} & \multicolumn{3}{l|}{Brighter Light Condition (8 items)} & \multicolumn{3}{l}{Brighter Light Condition (7 items)} \\ \hline
\multicolumn{2}{l|}{Images per Manipulator}                            & 4                & 7               & 9              & 4                & 7               & 9               & 4                & 7               & 9              \\ \hline
\multirow{2}{*}{DUSt3R\cite{DUSt3R_cvpr24} + Bi-JCR}  & Residual $\delta_R$ & \textbf{0.0769}  & \textbf{0.0724}  & \textbf{0.0668} & \textbf{0.0740}  & \textbf{0.0617}  & \textbf{0.0569}  & \textbf{0.0743}  & \textbf{0.0634}  & \textbf{0.0612} \\
                                                                      & Residual $\delta_\bb{t}$ & \textbf{0.0461}  & \textbf{0.0391}  & \textbf{0.0378} & \textbf{0.0587}  & \textbf{0.0351}  & \textbf{0.0340}  & \textbf{0.0424}  & \textbf{0.0341}  & \textbf{0.0373} \\ \hline
\multirow{2}{*}{MASt3R\cite{mast3r_arxiv24} + Bi-JCR} & Residual $\delta_R$ & 0.2613           & 0.2404           & 0.2020          & 0.2636           & 0.1541           & 0.1302           & 0.2350           & 0.1482           & 0.1312          \\
                                                                      & Residual $\delta_\bb{t}$ & 0.1639           & 0.1378           & 0.1219          & 0.1538           & 0.1012           & 0.0811           & 0.1514           & 0.1009           & 0.0896          \\ \hline
\multirow{2}{*}{VGGT\cite{wang2025vggt} + Bi-JCR}      & Residual $\delta_R$ & 0.3763           & 0.3728           & 0.5358          & 0.3793           & 0.3561           & 0.5186           & 0.5274           & 0.3492           & 0.3524          \\
                                                                      & Residual $\delta_\bb{t}$ & 0.1241           & 0.1255           & 0.1438          & 0.1290           & 0.1676           & 0.1818           & 0.2448           & 0.1692           & 0.1649          \\ 
                                                                \midrule[0.3pt]\bottomrule[1pt]
\end{tabular}
\end{adjustbox}
\caption{Quantitative result on selecting the best foundation model for Bi-JCR. The foundation model used by Bi-JCR, DUSt3R \cite{DUSt3R_cvpr24} is compared against MASt3R \cite{mast3r_arxiv24} and VGGT \cite{wang2025vggt} in term of rotational and translational residual loss from \Cref{eq:axxb_0}. Under all three scenarios, DUSt3R outperforms both MASt3R and VGGT in all number of views per manipulator.}
\label{table:foundation_model_ablation}
\vspace{-2em}
\end{table*}

\begin{figure}[t]
  \centering
  \begin{subfigure}[b]{0.99\linewidth}
    \centering
    \fbox{\includegraphics[width=.49\linewidth]{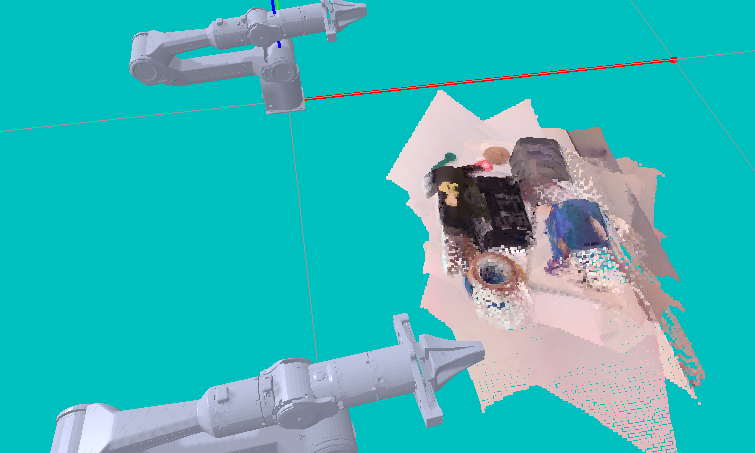}}%
    \fbox{\includegraphics[width=.49\linewidth]{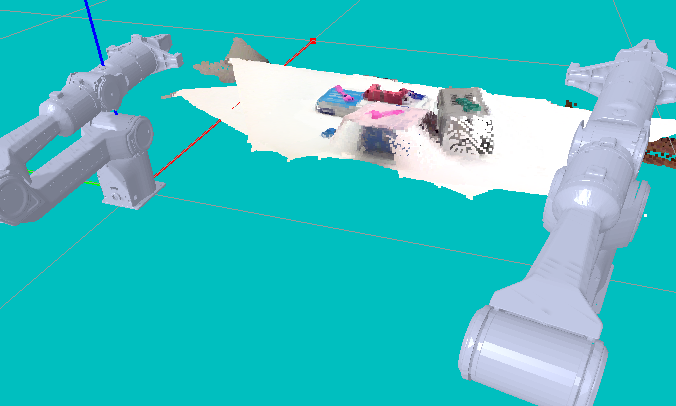}}
    
    \fbox{\includegraphics[width=.49\linewidth]{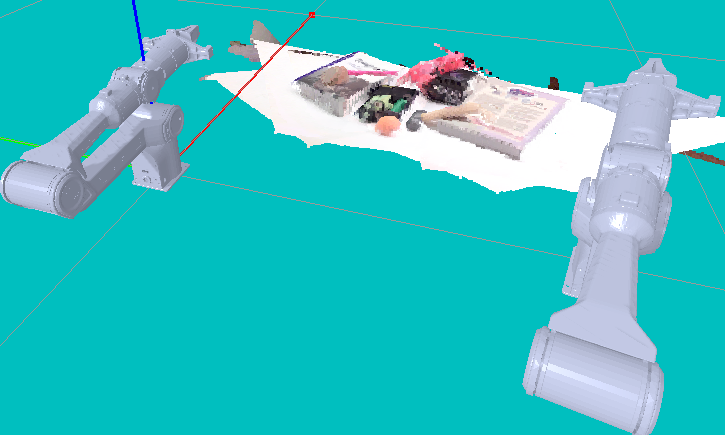}}%
    \fbox{\includegraphics[width=.49\linewidth]{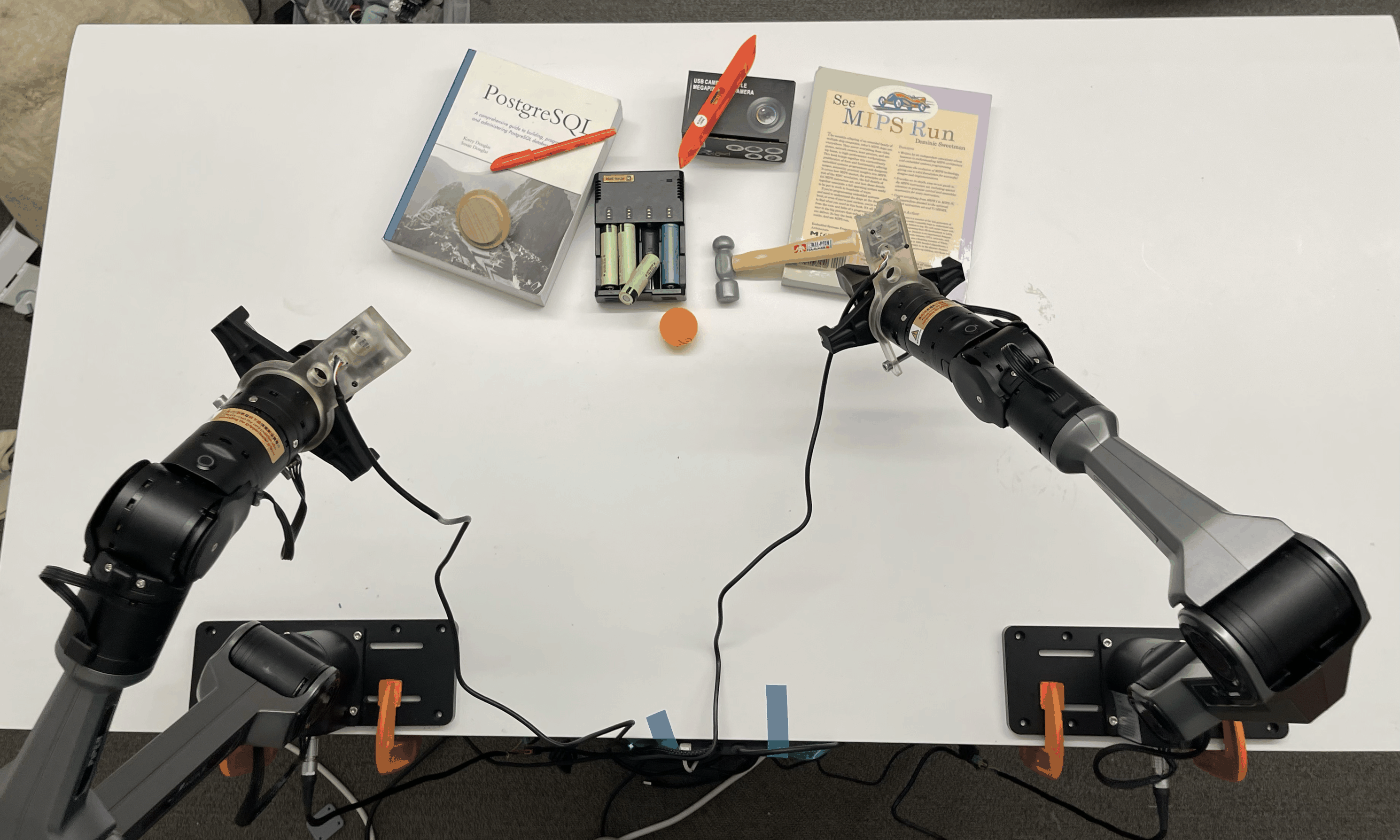}}
  \end{subfigure}

  \caption{Visualization of the base to base transformation recovery and transformation recovery of foundation model output frame to primary manipulator's base frame, including scene A (top left), scene B (top right), scene C (bottom left), and the real world bi-manual bases setup (bottom right).}
  \label{fig:base2base_viz}
  \vspace{-1.5em}
\end{figure}

\subsection{3D Representation in Primary Manipulator's Base Frame}
Besides scale, the quality of the 3D representation built by Bi-JCR is critical to downstream tasks. Here, we qualitatively assess the reconstructed 3D point cloud by visualizing it against images taken on the real world environment in \Cref{fig:reconstruction_quality}. We observe that Bi-JCR reconstructs the relative position and orientation of objects in the environment correctly, and the shape of each object is highly preserved. We further investigated whether representations can be accurately transformed into the robot's coordinate frame and the placement of the secondary manipulator base in the primary manipulator base frame. We inject the reconstruction, along with manipulator poses, into the PyBullet Simulator \cite{coumans2019}. As shown in \Cref{fig:base2base_viz}, the relative pose of the two manipulators highly resembles the relative pose of the two manipulators in the real world, indicating the correct estimation of the pose of the secondary manipulator in the base frame of the primary manipulator. The table 3D reconstruction in simulation remains parallel to the bases of manipulators, and object orientations and positions are visually correct relative to the bases of manipulators, indicating both a high-quality 3D reconstruction produced along with its accurate transformation into the primary manipulator's frame.

\subsection{Ablation Study on Different Foundation Models}
With recent development of 3D reconstruction using structure from motion (SfM), there have been many new foundation models that outperform the DUSt3R foundation model \cite{DUSt3R_cvpr24} chosen by us in various task benchmarks, such as MASt3R \cite{mast3r_arxiv24} and VGGT \cite{wang2025vggt}. Therefore, we evaluate the performance of Bi-JCR with different foundation models in 4, 7, 9 numbers of views per manipulator, and we choose the complete mode to find correspondences between all pairs of views for MASt3R, and we report the residual loss in \Cref{table:foundation_model_ablation}. Unlike VGGT, both MASt3R and DUSt3R receive a lower residual loss compared to VGGT in a higher number of views per manipulator, which is likely because MASt3R and DUSt3R utilize the ICP process to retain consistency in camera pose estimation. Furthermore, DUSt3R produces camera poses with better overall residual calibration loss, so DUSt3R remains the primary choice for Bi-JCR.

\subsection{Ablation Study on Bi-JCR's Gradient Descent}
Bi-JCR's leverages gradient descent to further refine initial solutions. We experimentally evaluate the benefit of the gradient descent refinement by comparing the residual losses of Bi-JCR, with and without refinement via Gradient Descent \cite{Bishop:2006}. As shown in \Cref{table:sgd_ablation}, the gradient descent refinement shows a marked improvement in rotational loss with fewer views. When the number of views per manipulator is doubled from 4 to 8, Bi-JCR with gradient descent refinement still outperforms Bi-JCR without refinement. We observe that gradient descent refinement plays a critical component in Bi-JCR under a sparse view setup, but is less impactful when the number of images per manipulator increases.

\begin{table}[t]
\begin{adjustbox}{width=\linewidth,center}
\begin{tabular}{ll|rr|rr}
\toprule[1pt]\midrule[0.3pt]
\multicolumn{2}{l|}{}                 & \multicolumn{2}{l|}{Darker Condition} & \multicolumn{2}{l}{Lighter Condition} \\ \hline
\multicolumn{2}{l}{Images per Manipulator} & 4                    & 8                  & 4                     & 8                   \\ \hline
\multirow{2}{*}{Bi-JCR w/ GD}     & Residual $\delta_R$   & \textbf{0.0785}      & \textbf{0.0698}      & \textbf{0.0743}       & \textbf{0.0619}      \\
                                 & Residual $\delta_\bb{t}$  & 0.0478               & \textbf{0.0390}      & \textbf{0.0424}       & \textbf{0.0372}      \\ \hline
\multirow{2}{*}{Bi-JCR w/o GD}    & Residual $\delta_R$   & 0.0920               & 0.0704               & 0.0927                & 0.0628               \\
                                 & Residual $\delta_\bb{t}$   & \textbf{0.0477}      & 0.0391               & 0.0427                & 0.0374               \\ \midrule[0.3pt]\bottomrule[1pt]
\end{tabular}
\end{adjustbox}
\caption{Quantitative result on the effect of Gradient Descent (GD) \cite{bishop2006pattern} in Bi-JCR, observe that under sparse images condition, GD is able to significantly improve rotational residual error, and Bi-JCR with GD also slightly outperform Bi-JCR without GD in large number of images setup.}
\label{table:sgd_ablation}
\vspace{-1em}
\end{table}

\begin{figure}[t]
  \centering
  \begin{subfigure}[b]{0.57\linewidth}   
      \centering
      \includegraphics[width=0.485\linewidth]{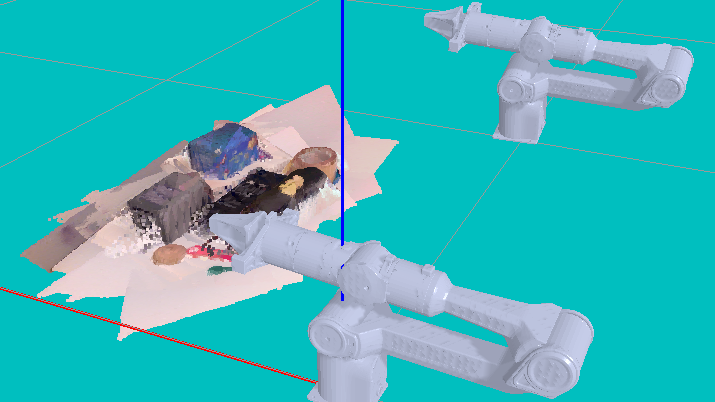}
      \includegraphics[width=0.485\linewidth]{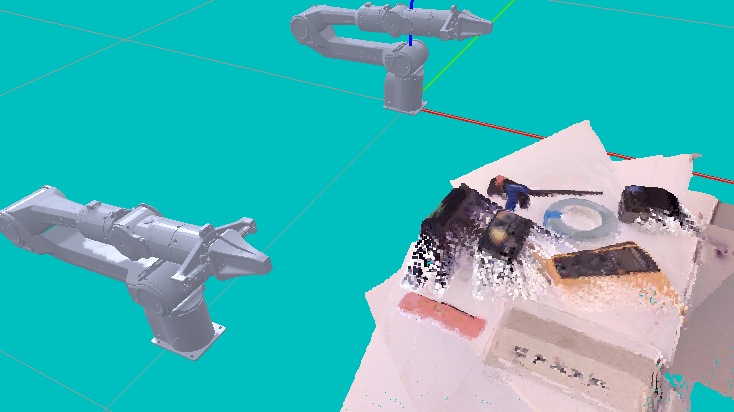}
    \caption{Representation in the robot's frame.}\label{fig:Bi-JCR_segmentation_a}
  \end{subfigure}
  \begin{subfigure}[b]{0.41\linewidth}
    \centering
      \fbox{\includegraphics[width=0.48\linewidth]{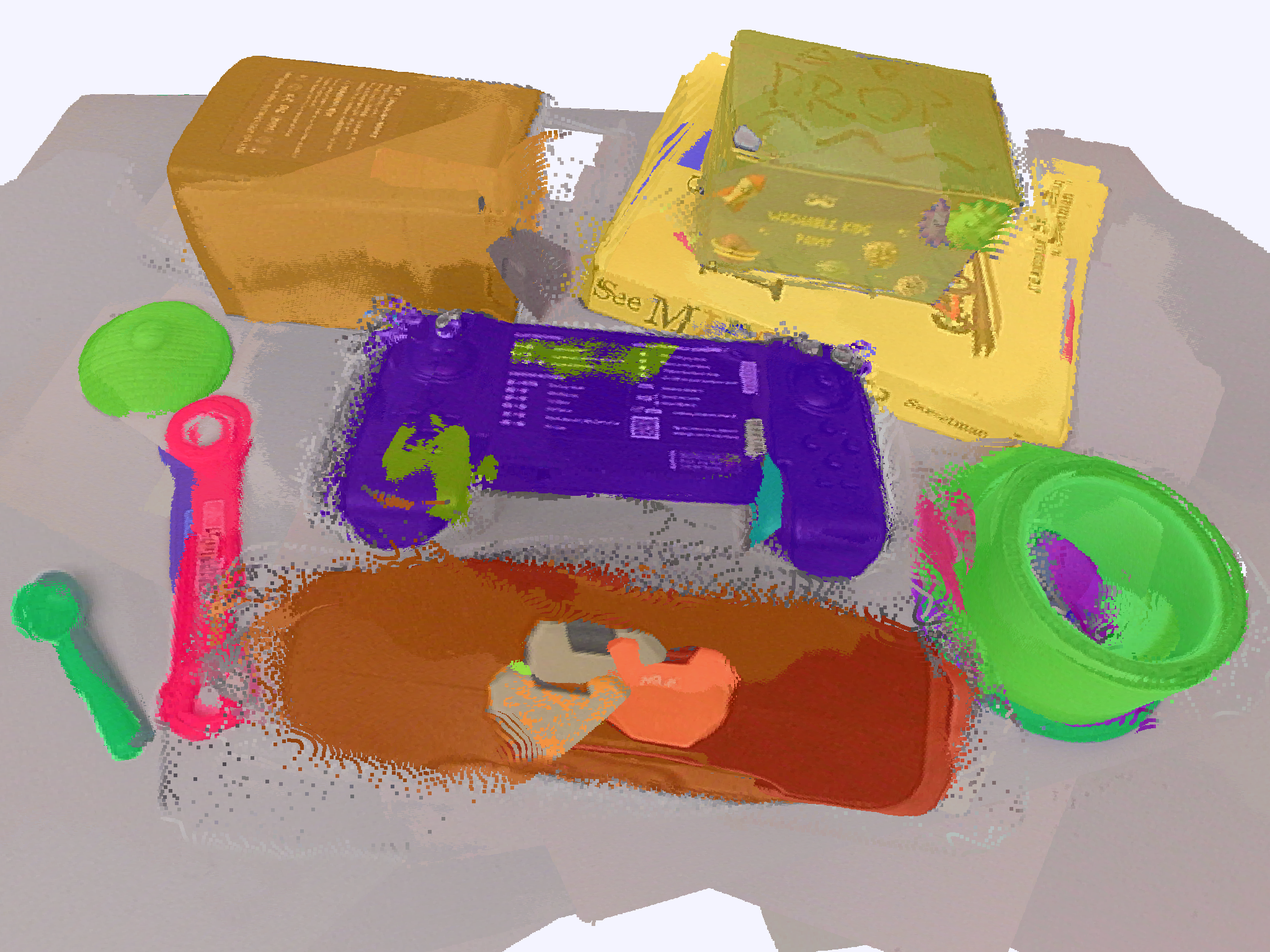}}%
      \fbox{\includegraphics[width=0.48\linewidth]{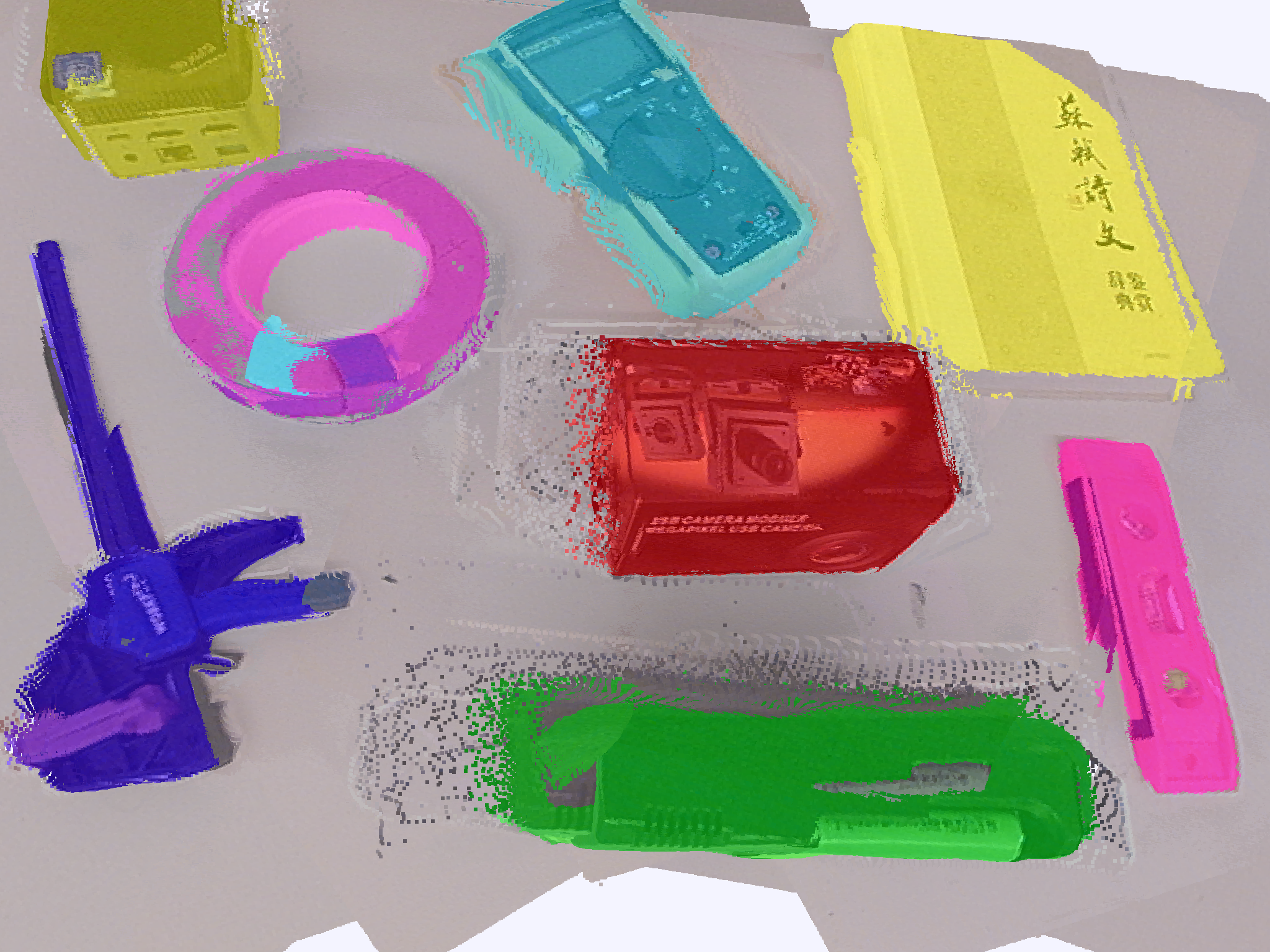}}
    \caption{Segmentated tabletop.}\label{fig:Bi-JCR_segmentation_b}
  \end{subfigure}
  \caption{Visualization of running segmentation algorithm in the real world metric scale reconstructed 3D representation from Bi-JCR, which allows various bi-manual downstream tasks such as joint grasping and passing.}
  \label{fig:Bi-JCR_segmentation}
  \vspace{-2em}
\end{figure}

\begin{figure}[t]
    \centering
    \begin{subfigure}[b]{\linewidth}   
        \centering
            \fbox{\includegraphics[width=0.245\linewidth]{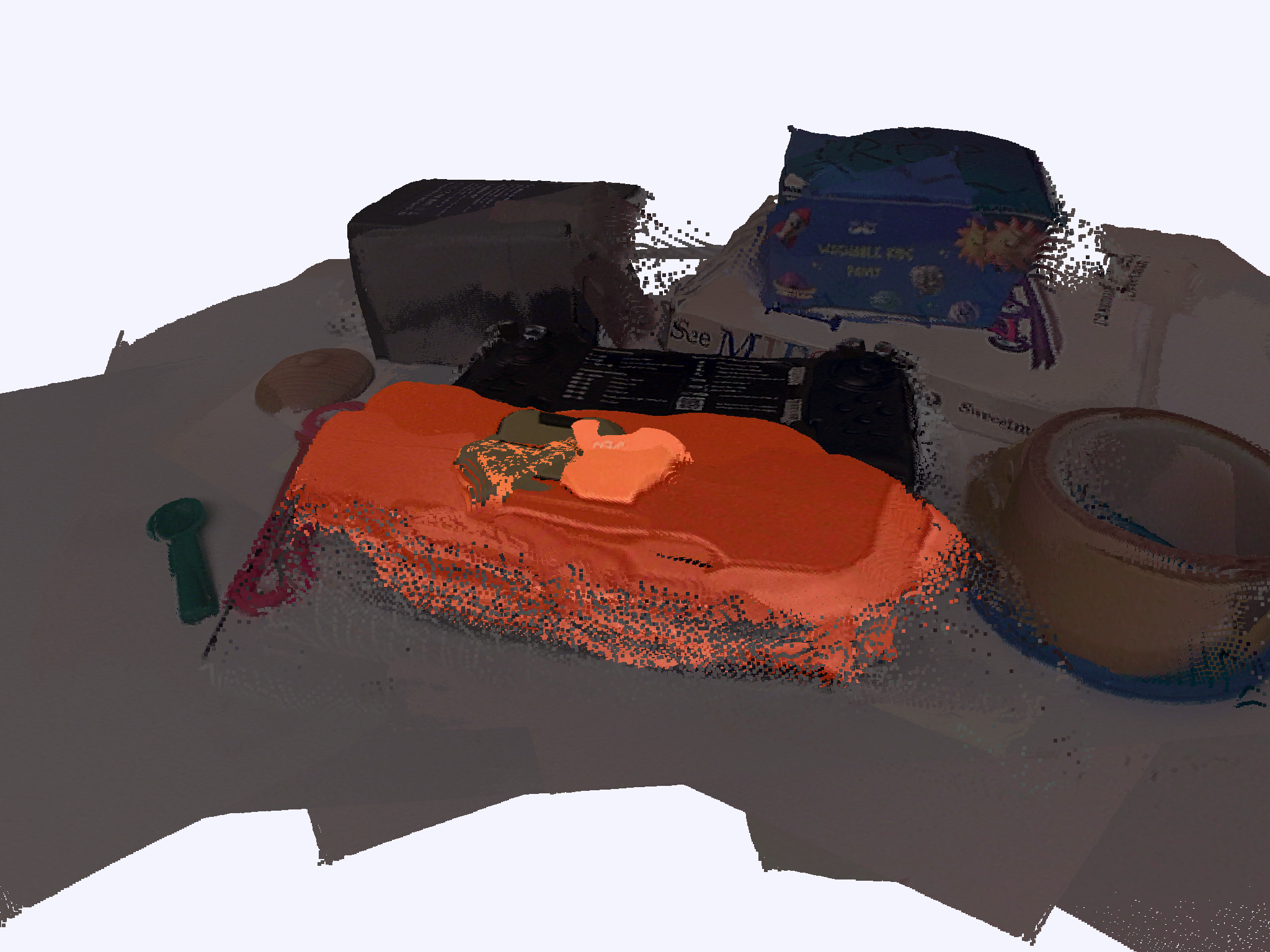}}%
            \fbox{\includegraphics[width=0.245\linewidth]{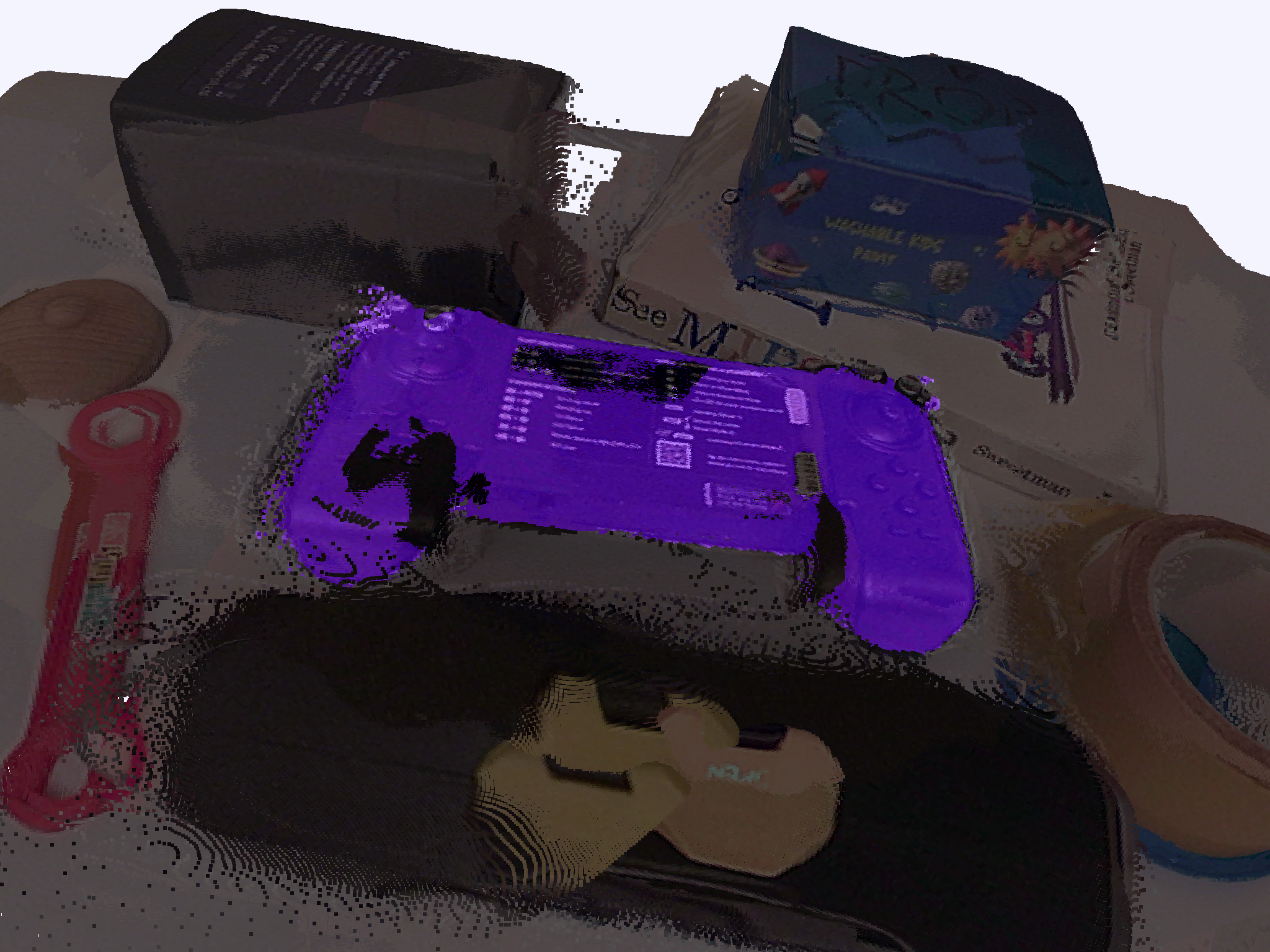}}%
            \fbox{\includegraphics[width=0.245\linewidth]{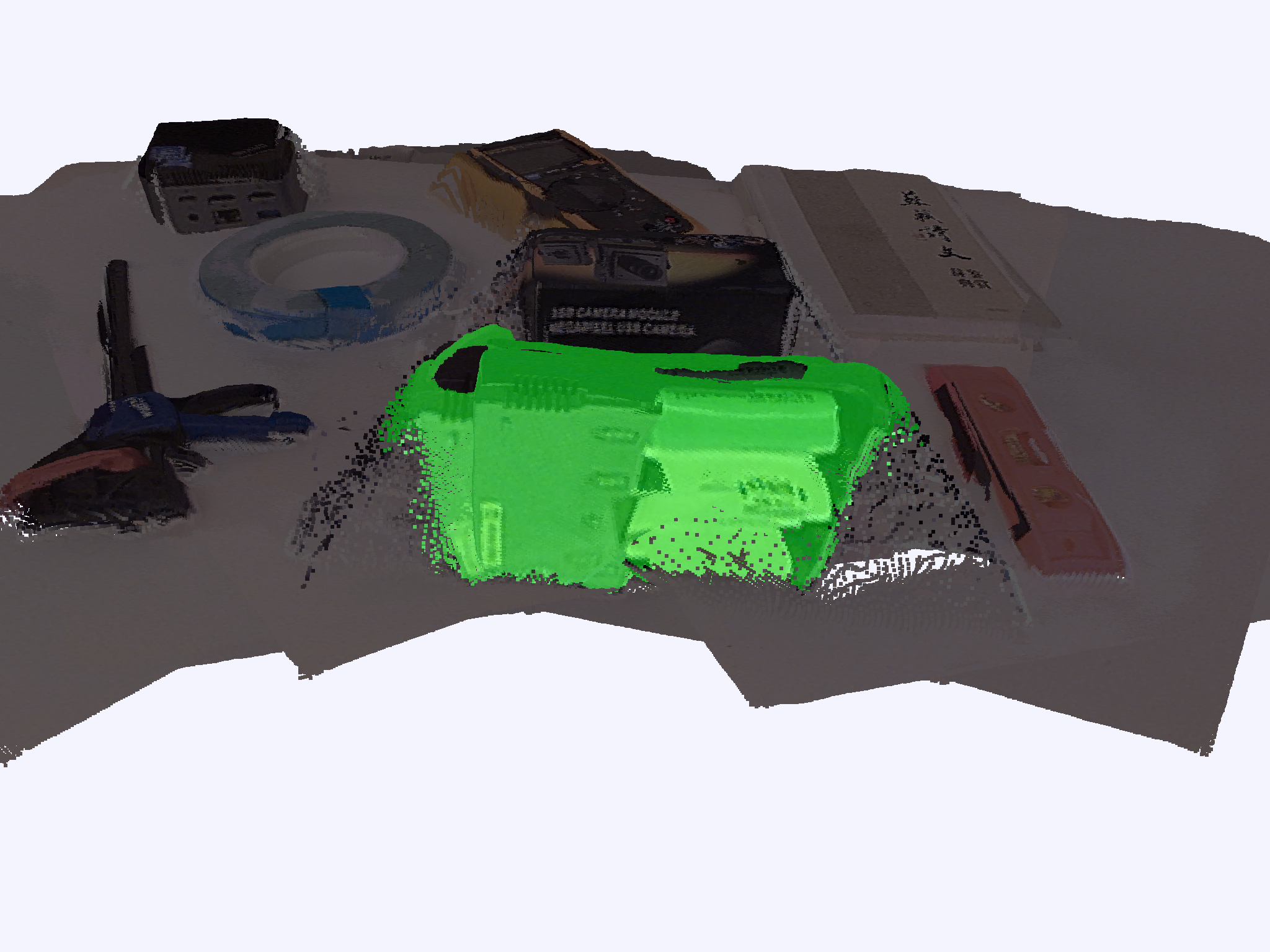}}%
            \fbox{\includegraphics[width=0.245\linewidth]{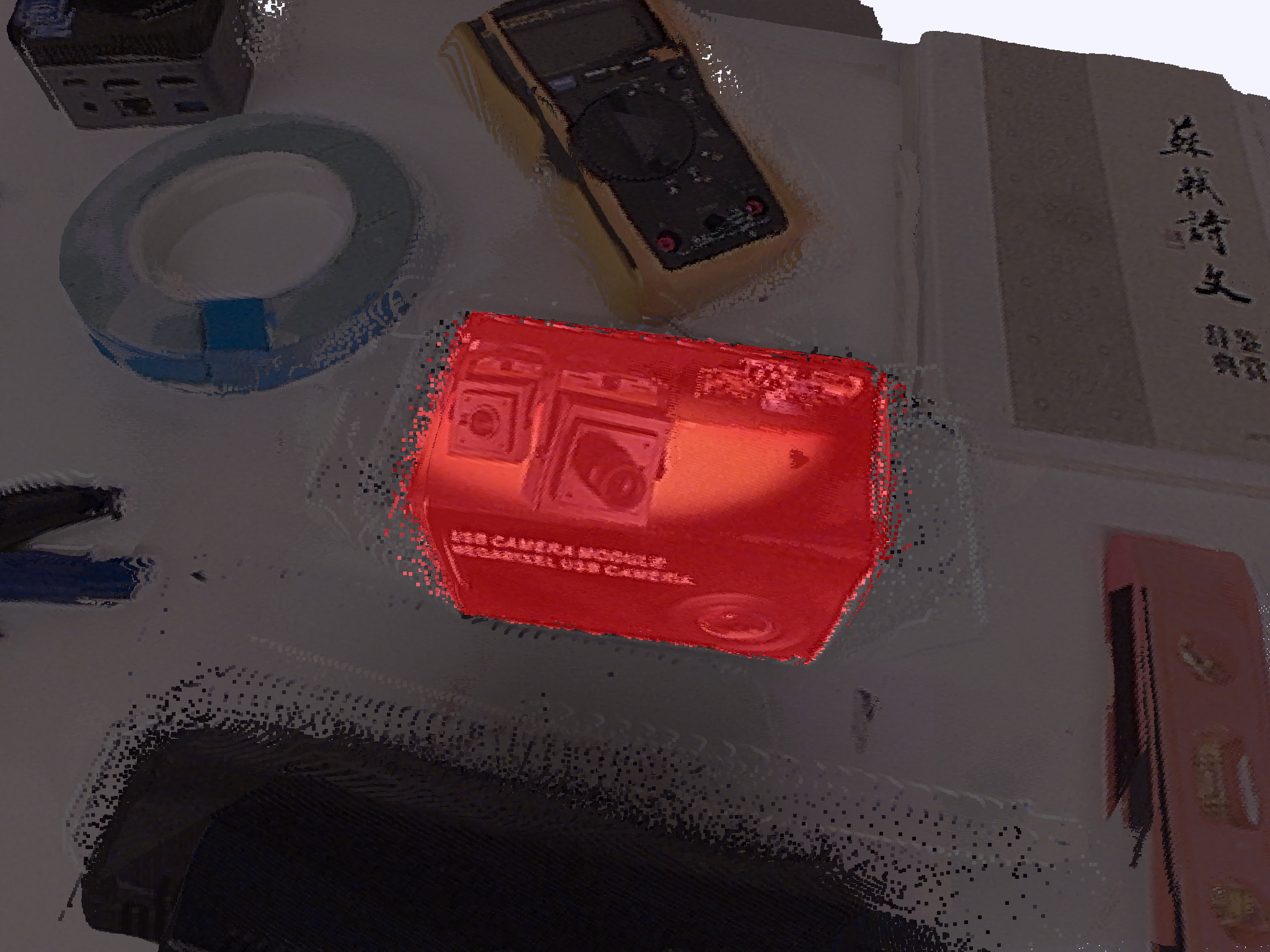}}%
        \caption{Heavier objects are selected from the scene, and can be segmented out.}\label{fig:joint_grasp_a}
    \end{subfigure}

    \begin{subfigure}[b]{\linewidth}   
        \centering
            \fbox{\includegraphics[width=0.245\linewidth]{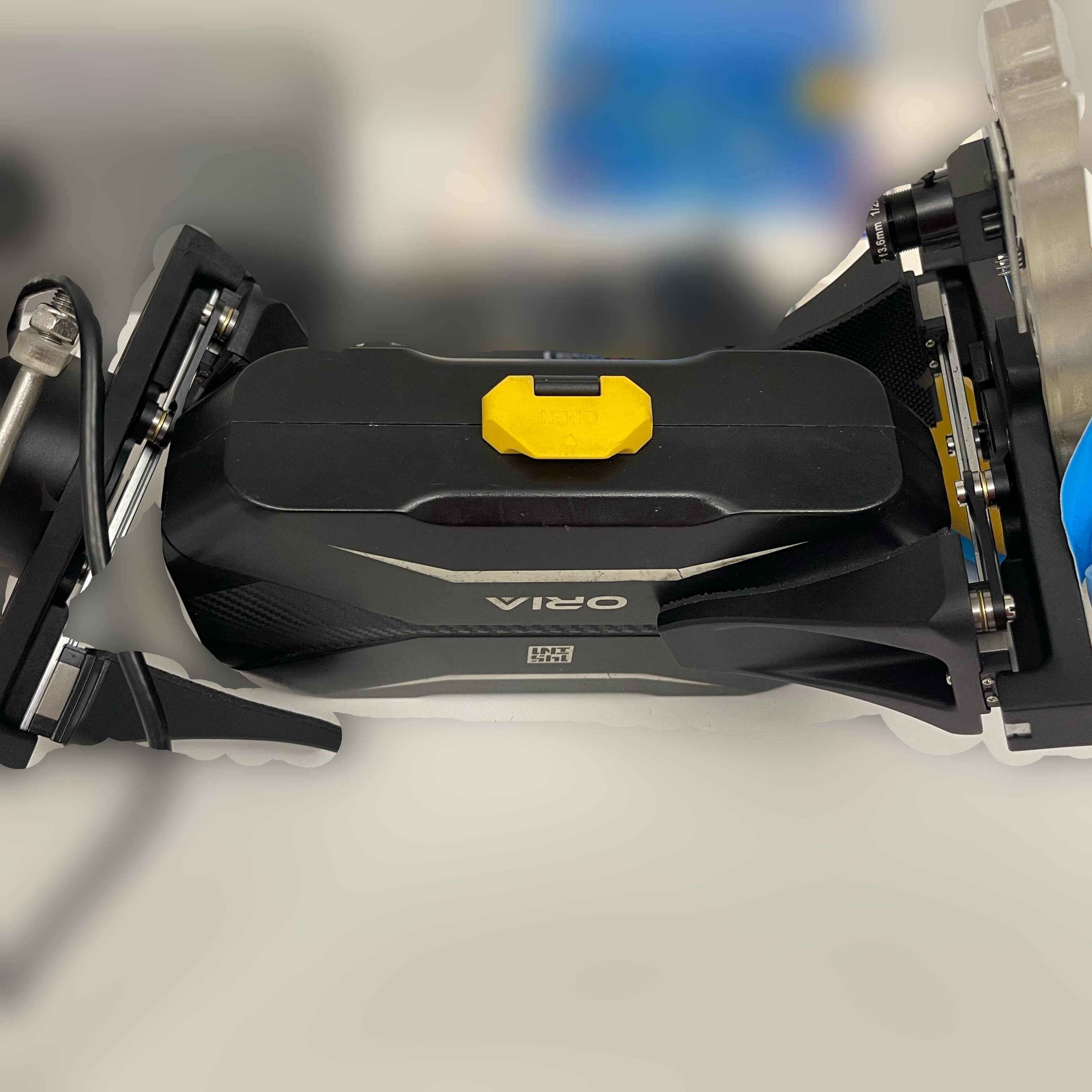}}%
                \fbox{\includegraphics[width=0.245\linewidth]{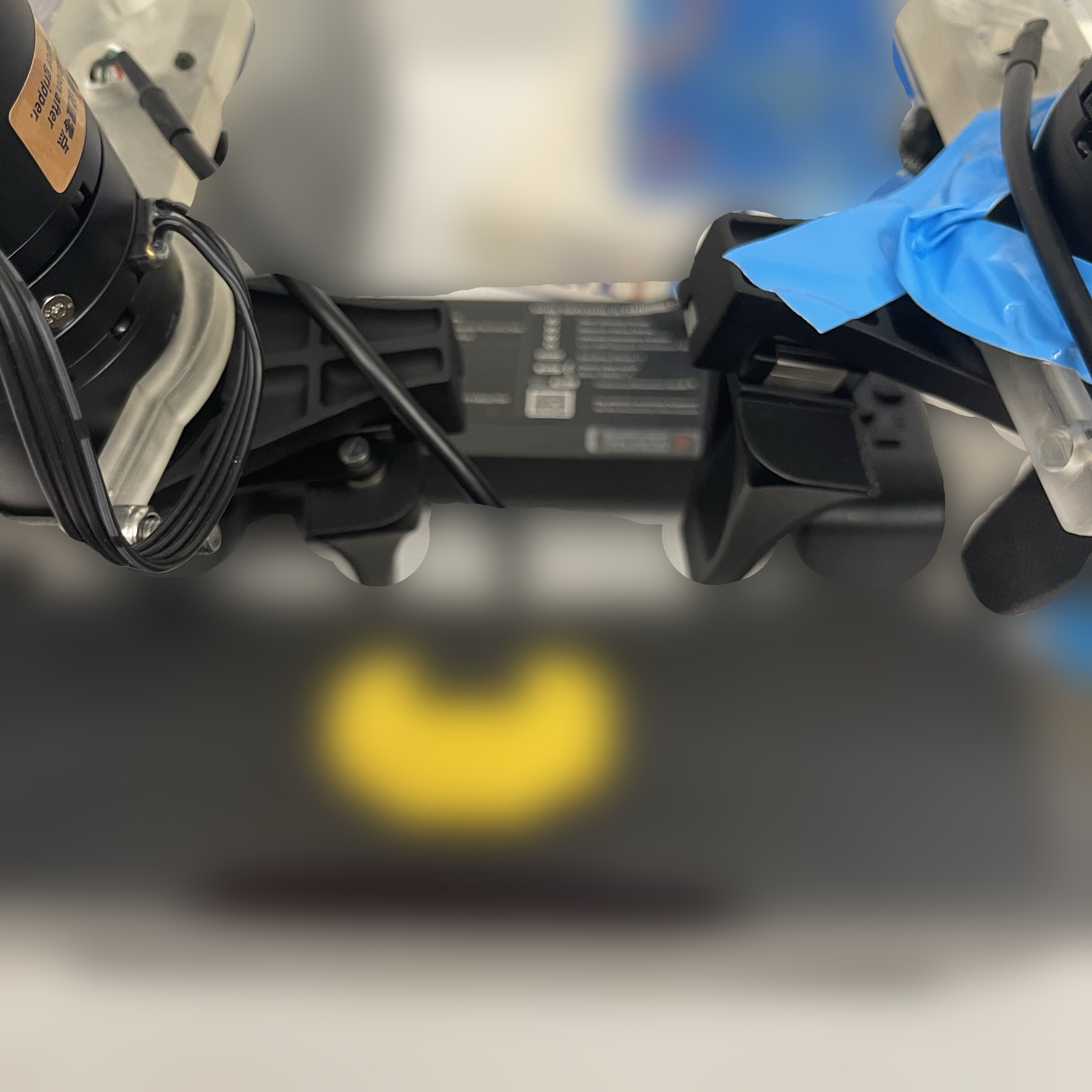}}%
                \fbox{\includegraphics[width=0.245\linewidth]{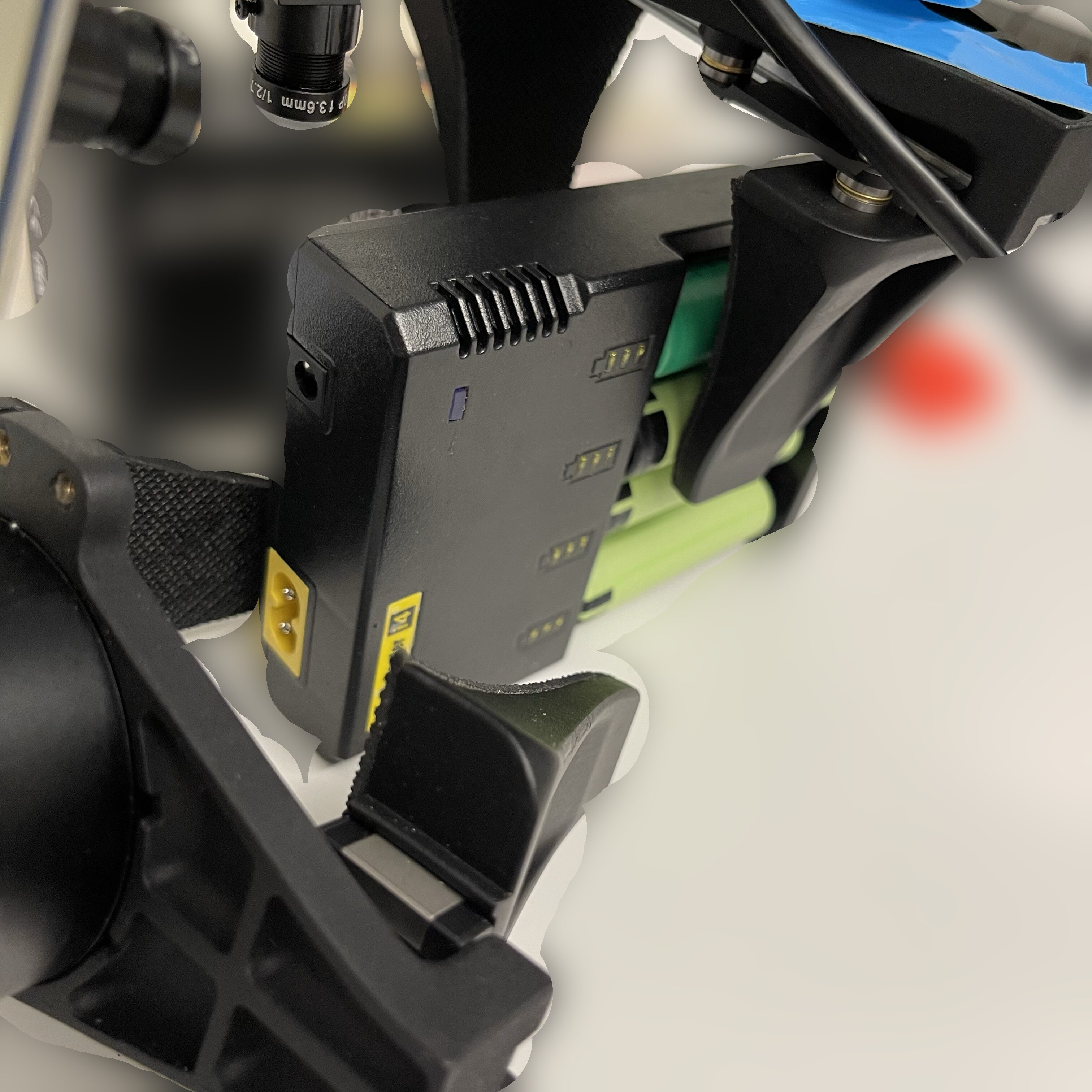}}%
                \fbox{\includegraphics[width=0.245\linewidth]{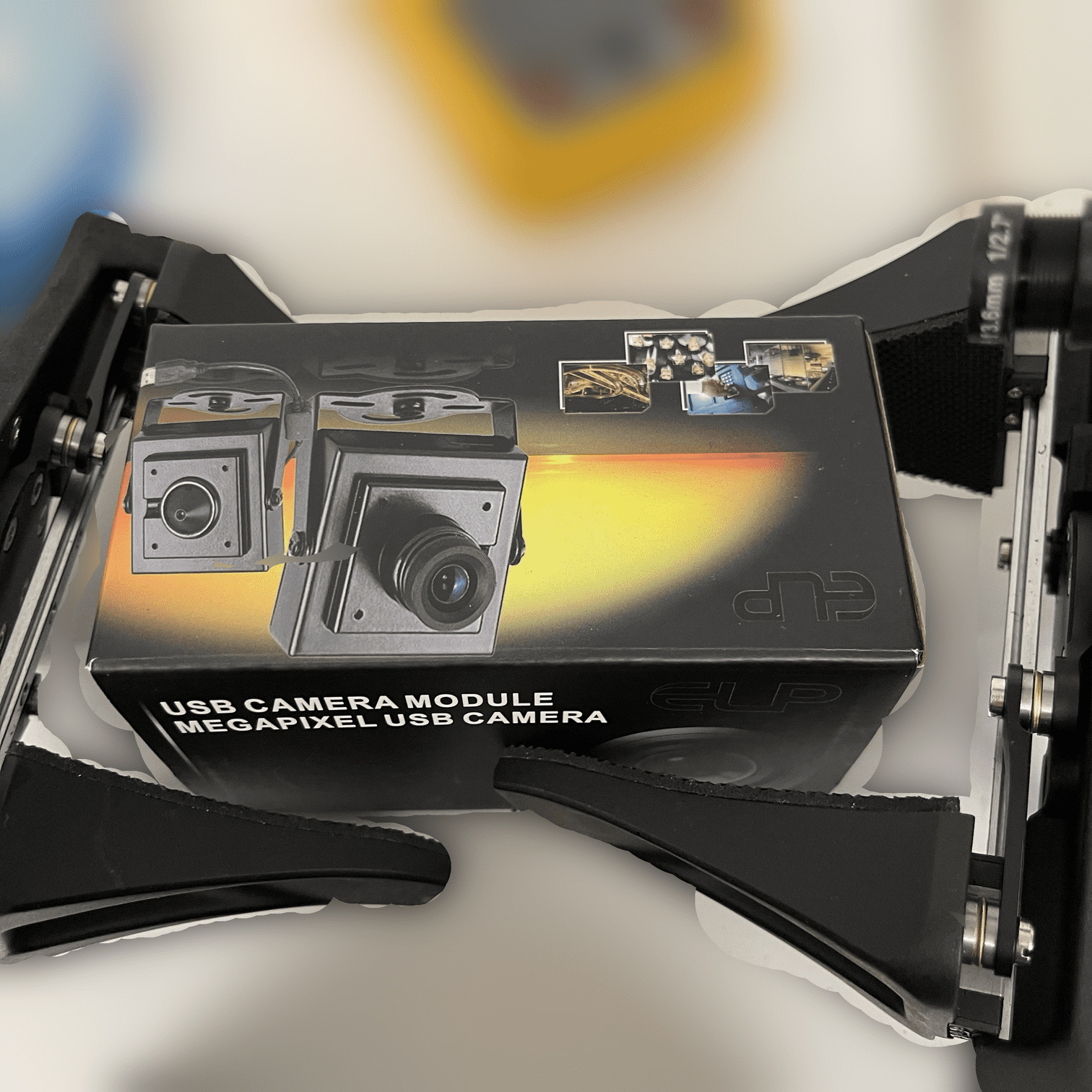}}

                \fbox{\includegraphics[width=0.245\linewidth]{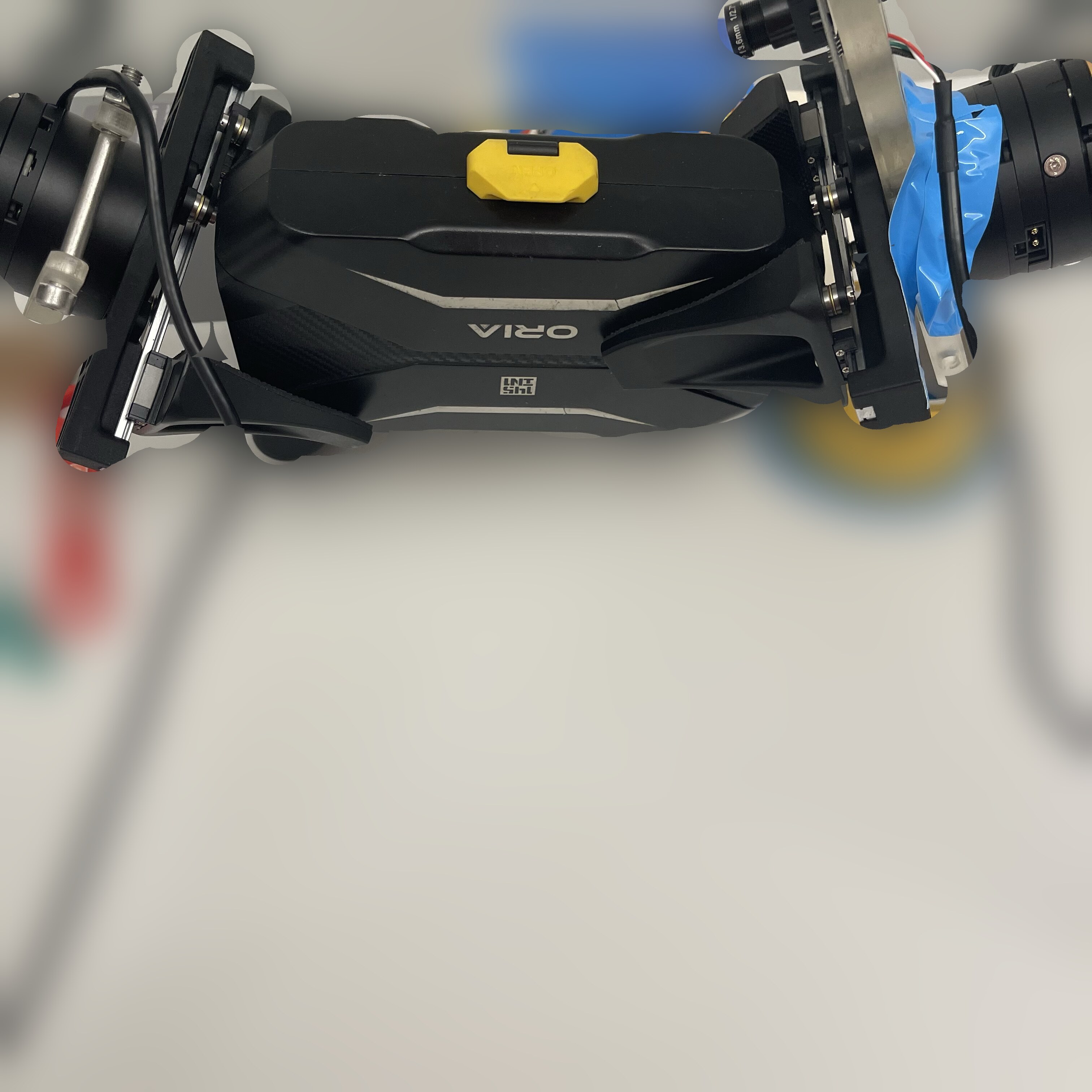}}%
                \fbox{\includegraphics[width=0.245\linewidth]{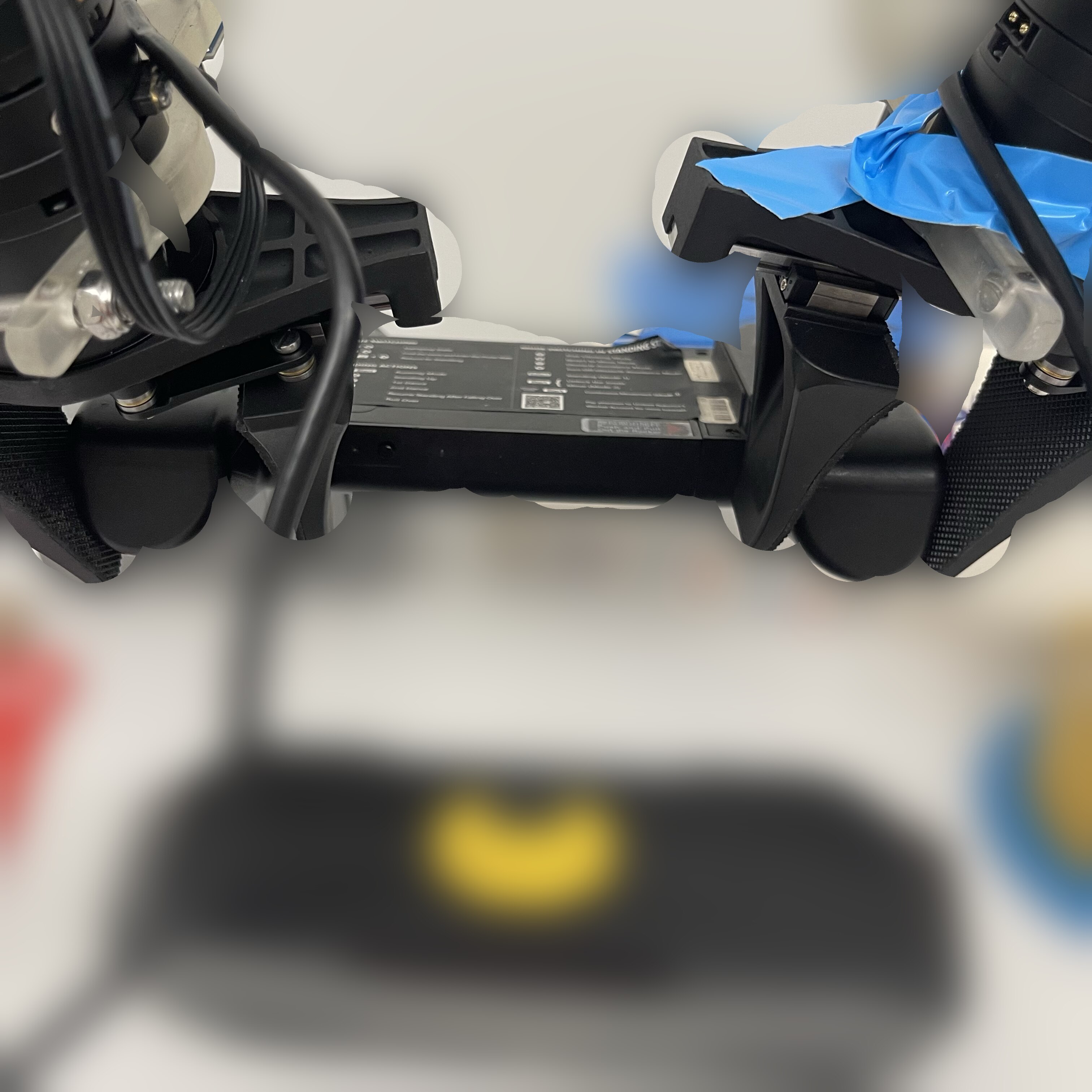}}%
                \fbox{\includegraphics[width=0.245\linewidth]{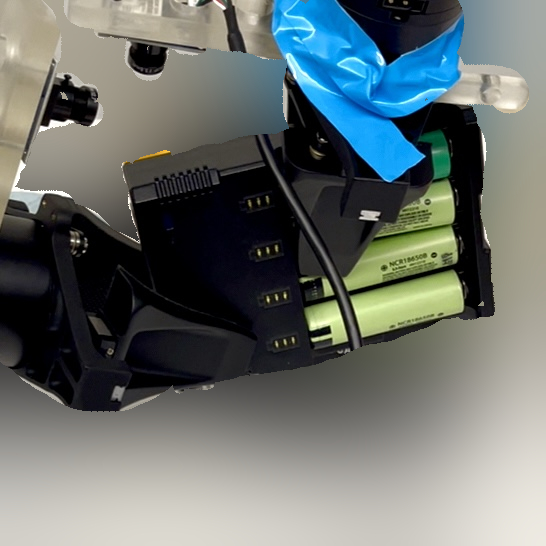}}%
                \fbox{\includegraphics[width=0.245\linewidth]{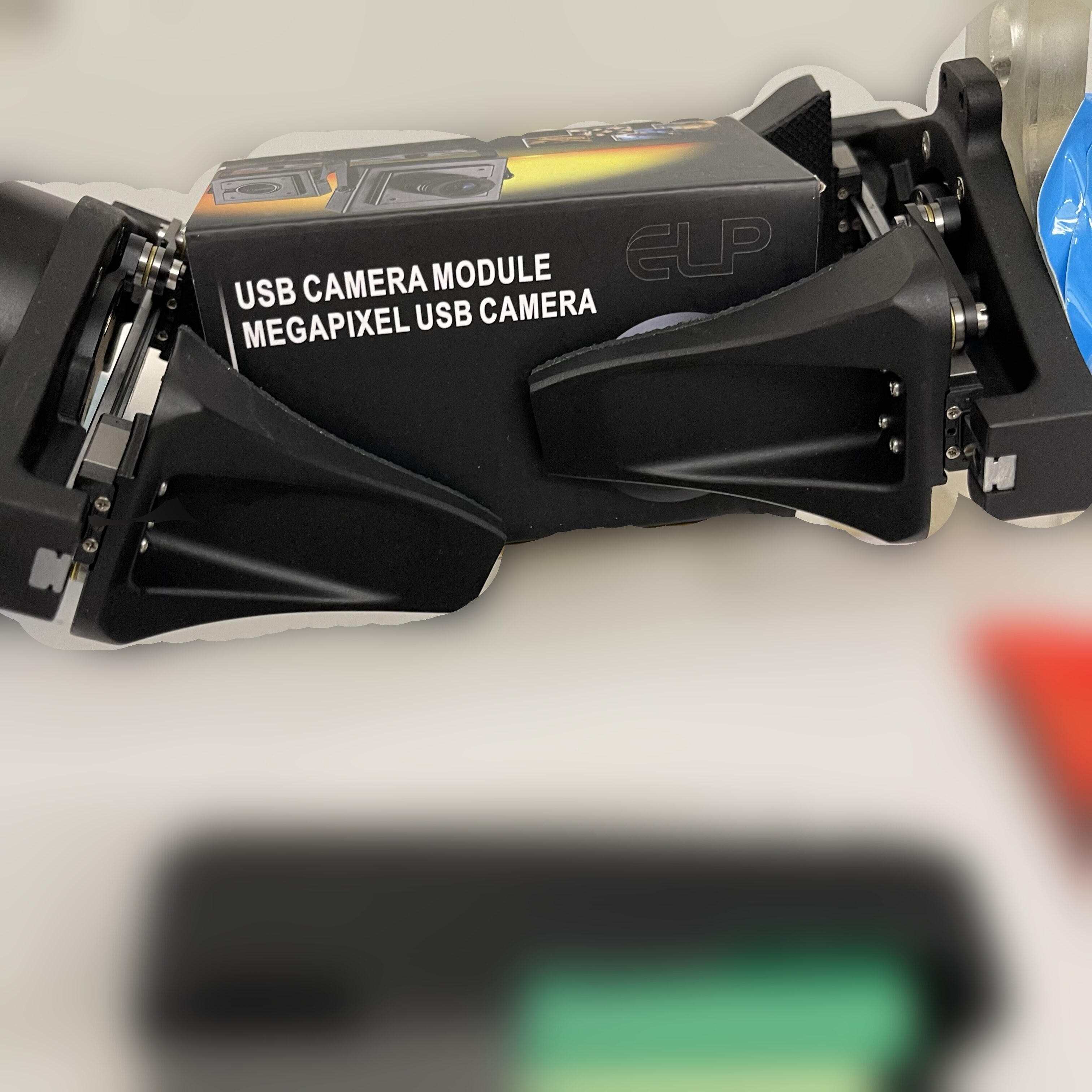}}
        \caption{We can execute Bi-manual grasps computed from the representation produced by Bi-JCR, in the real world. }\label{fig:joint_grasp_b}
    \end{subfigure}
    \caption{Execution of the bi-manual joint grasping on heavier objects in the scene. Background blurred for greater clarity.}
    \label{fig:joint_grasp}
    \vspace{-1em}
\end{figure}

\begin{figure}[t]
    \centering
    \fbox{\begin{subfigure}[b]{\linewidth}   
        \centering
            \includegraphics[width=0.2499\linewidth]{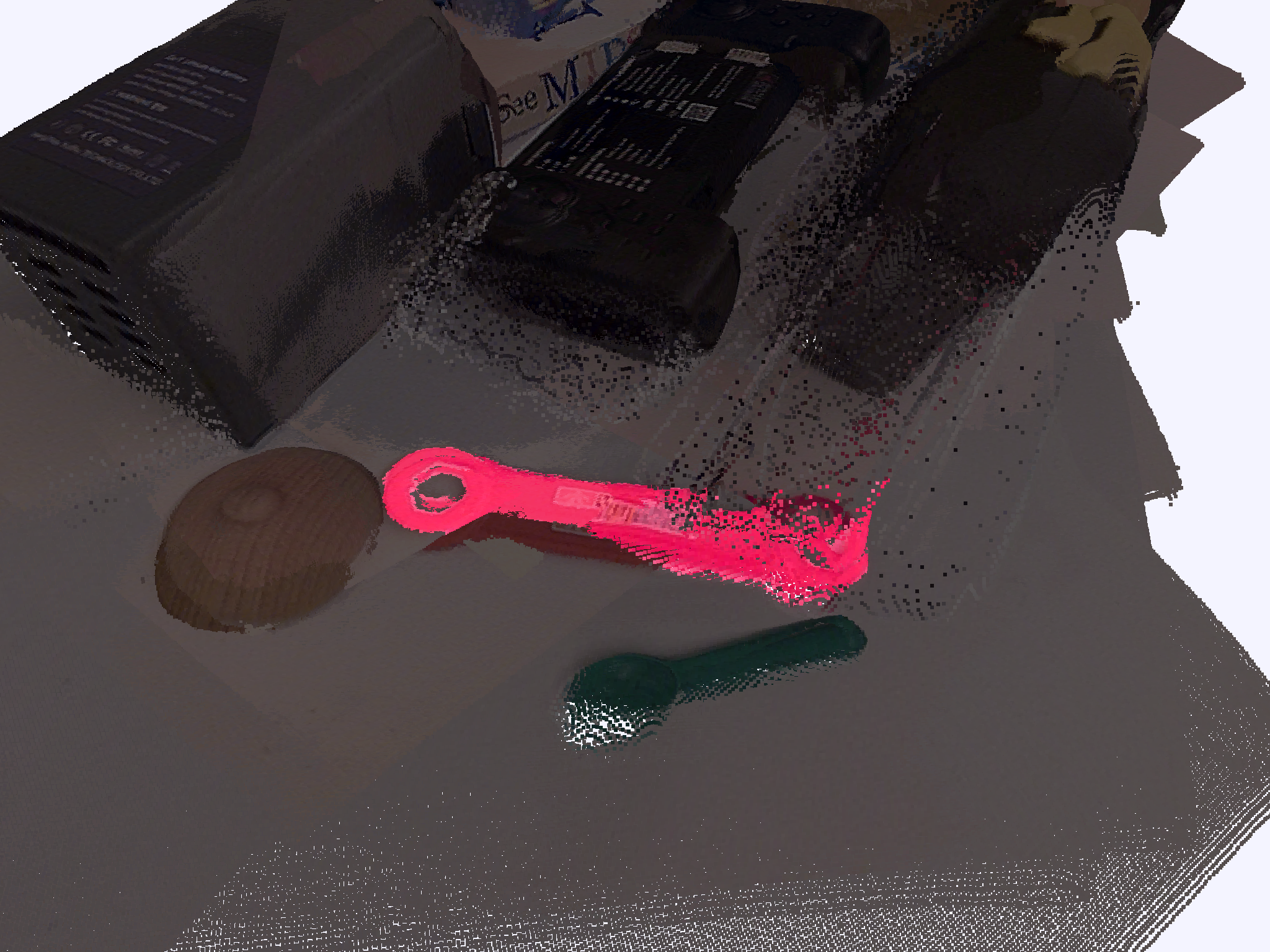}%
            \includegraphics[width=0.2499\linewidth]{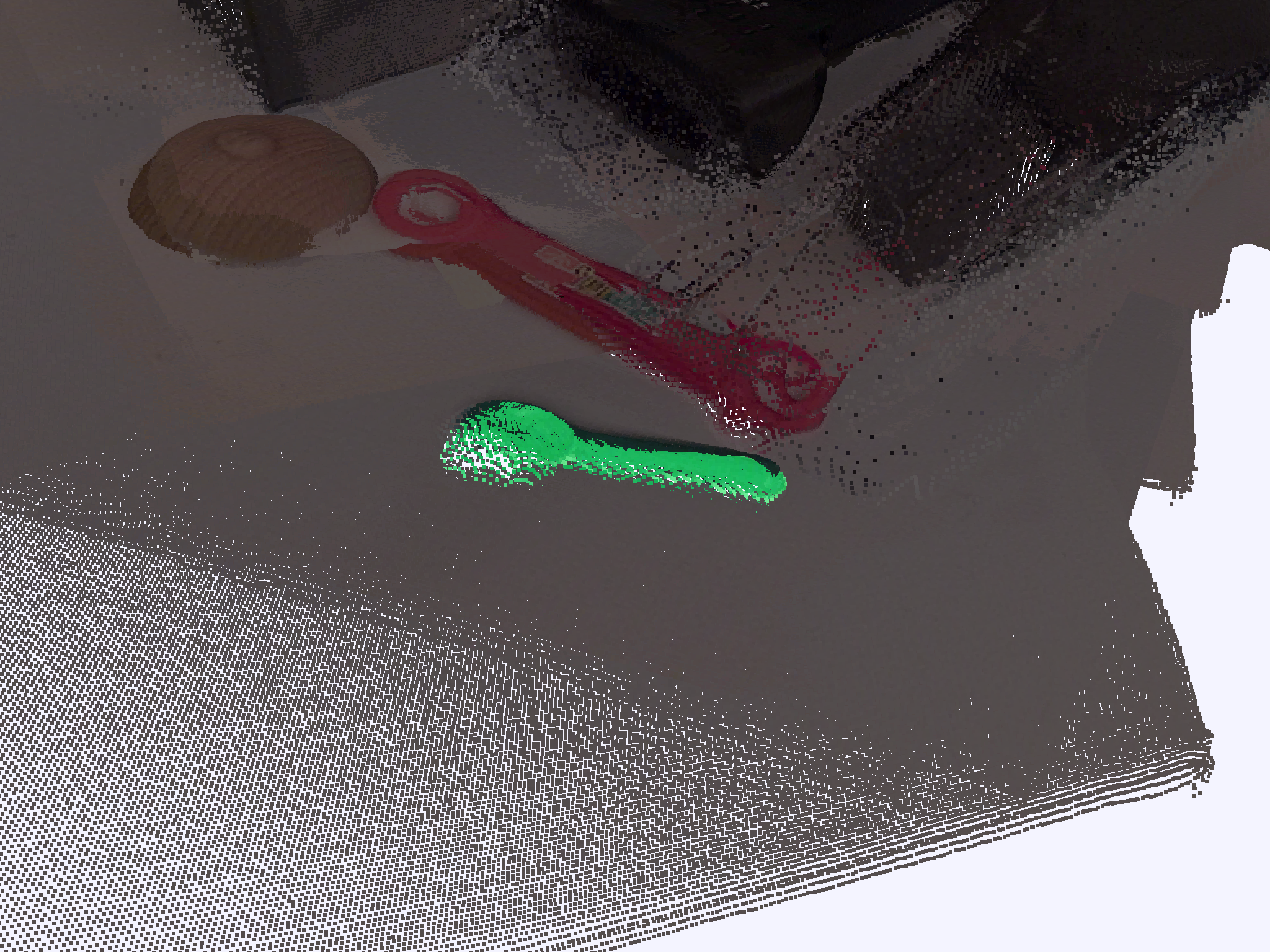}%
            \includegraphics[width=0.2499\linewidth]{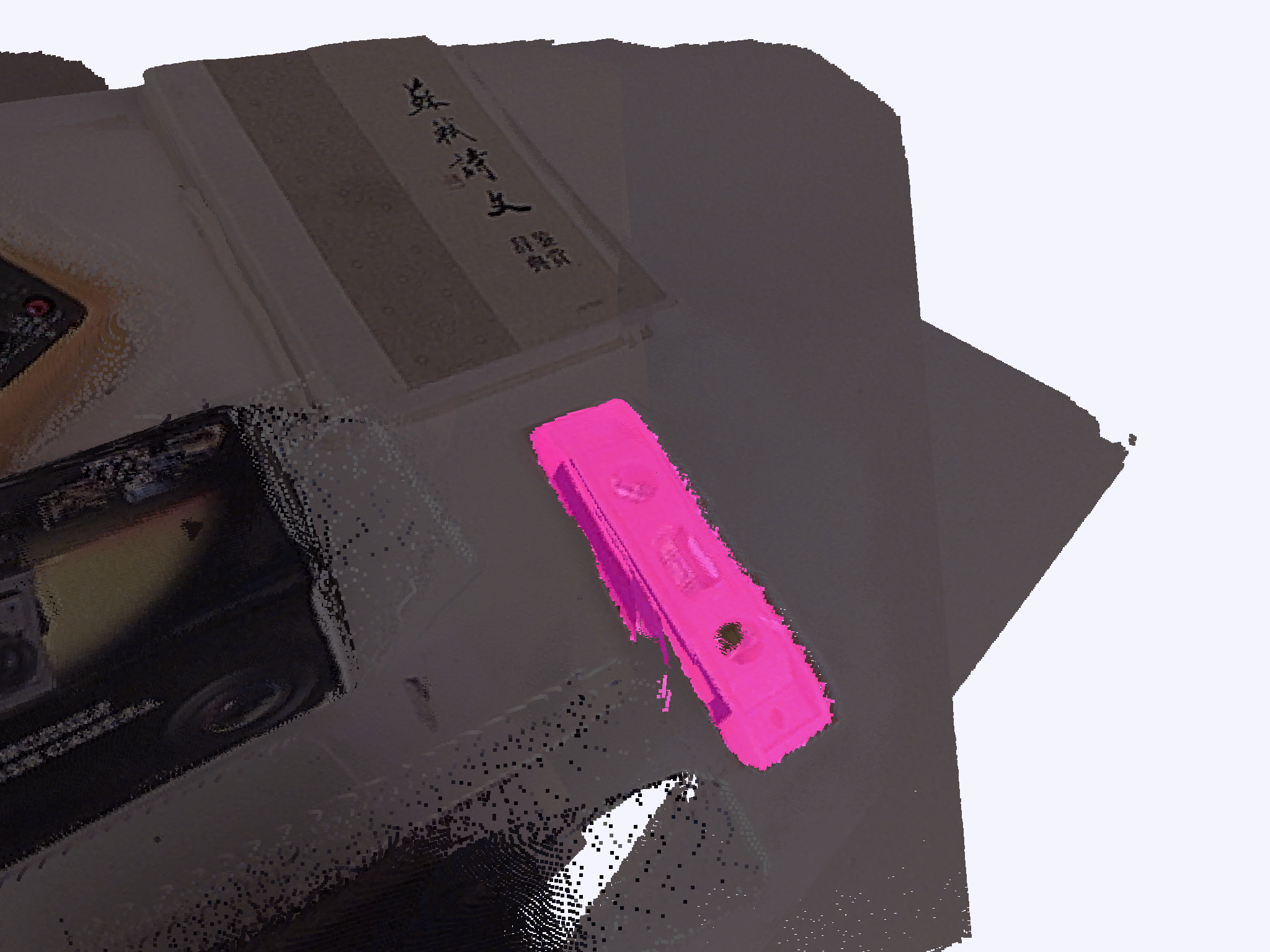}%
            \includegraphics[width=0.2499\linewidth]{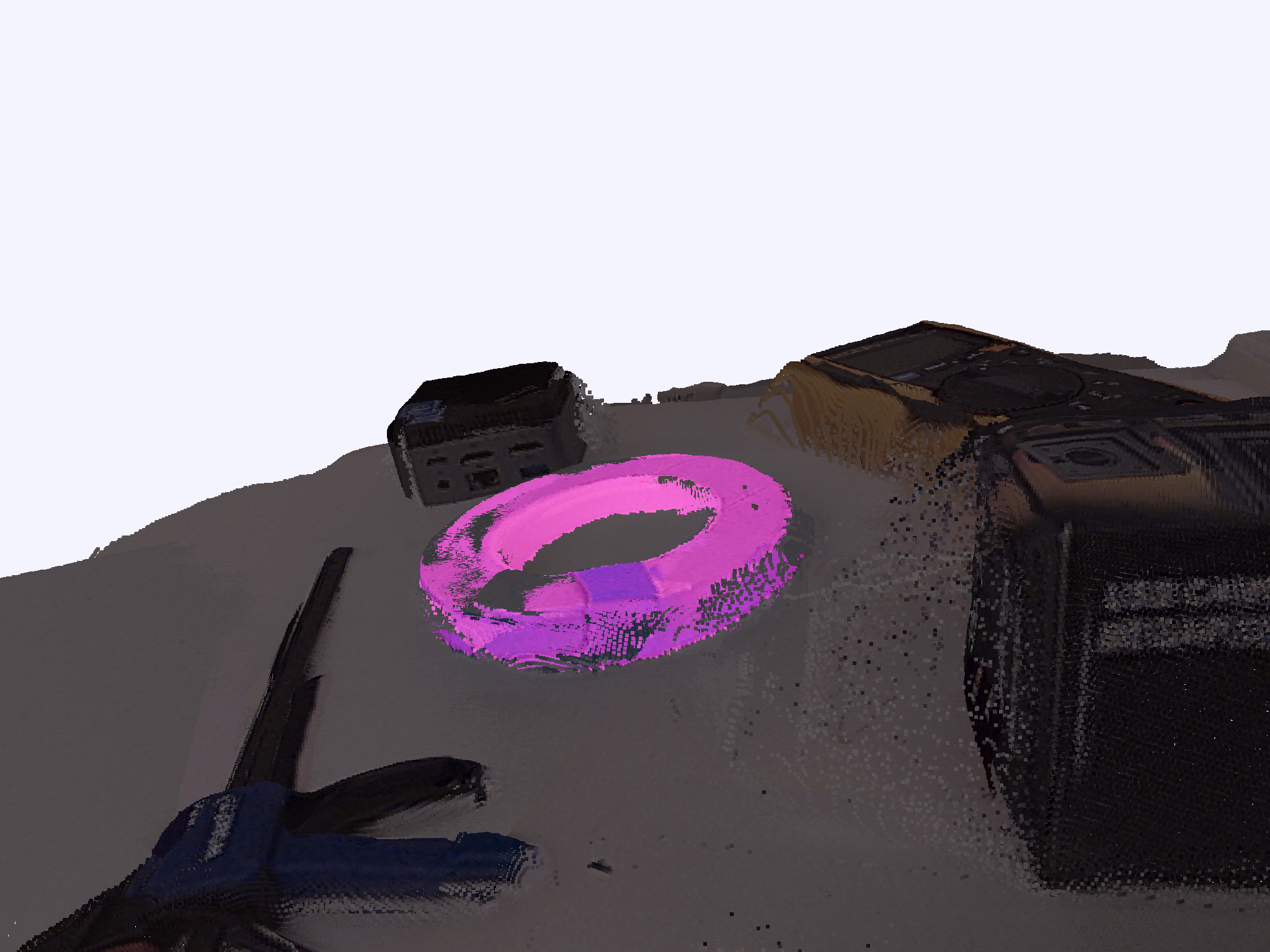}%
    \end{subfigure}}
    \fbox{\begin{subfigure}[b]{\linewidth}   
        \centering
            \includegraphics[width=0.2499\linewidth]{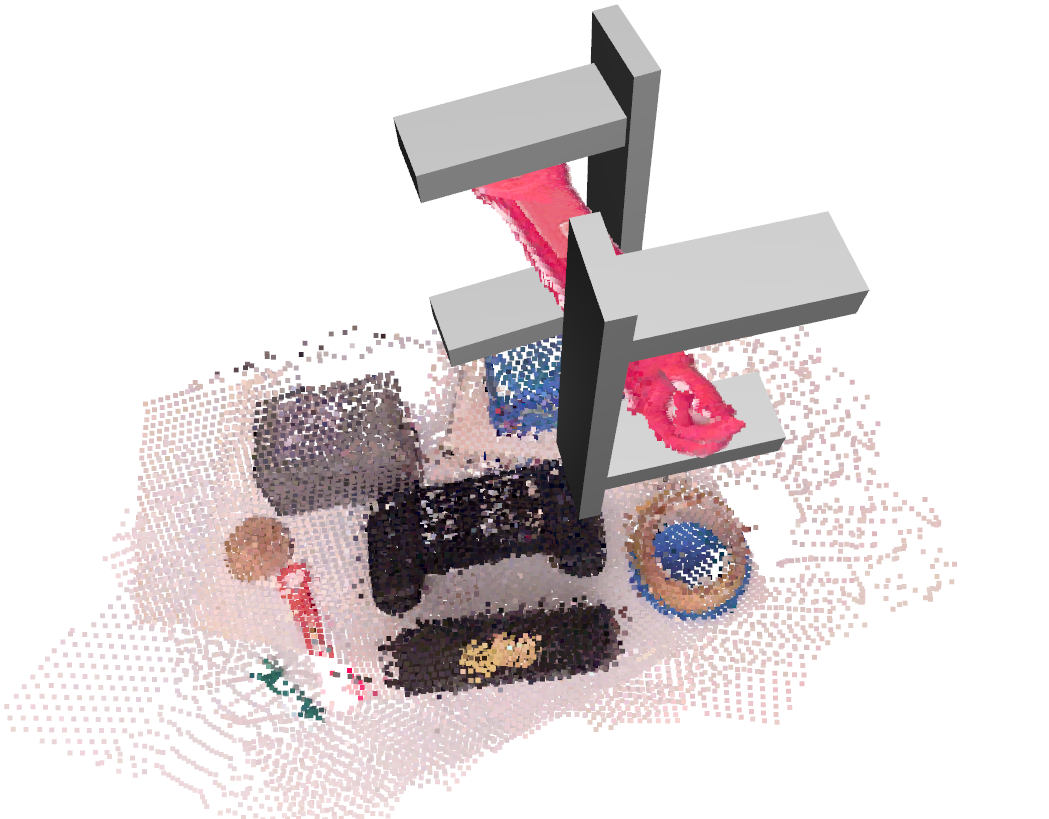}%
            \includegraphics[width=0.2499\linewidth]{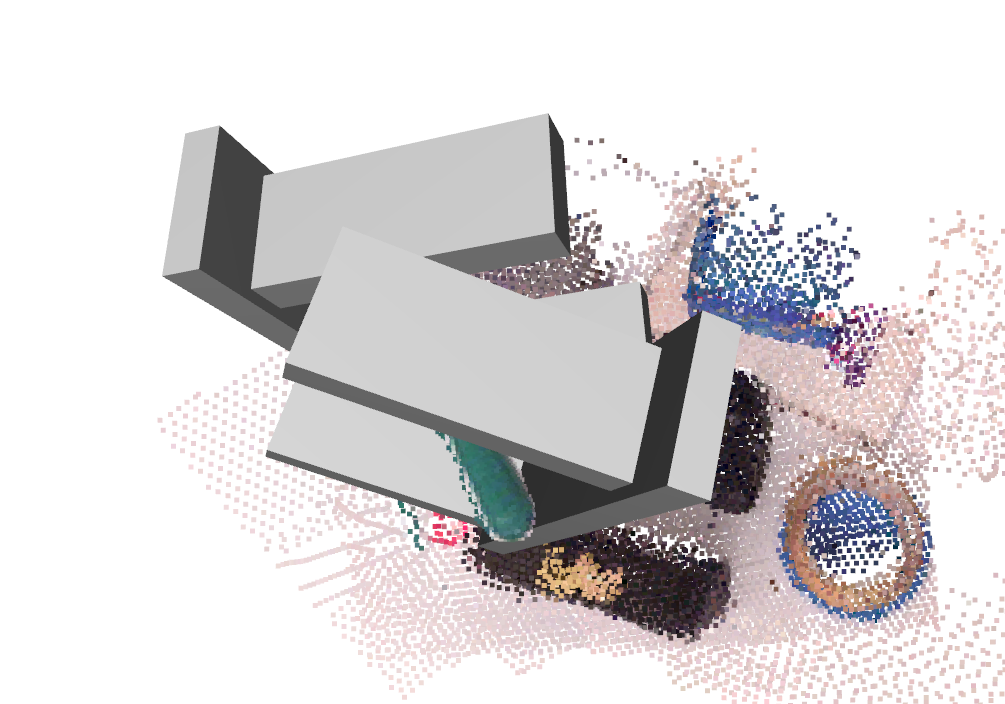}%
            \includegraphics[width=0.2499\linewidth]{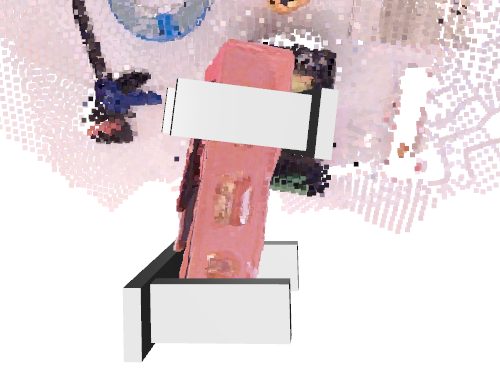}%
            \includegraphics[width=0.2499\linewidth]{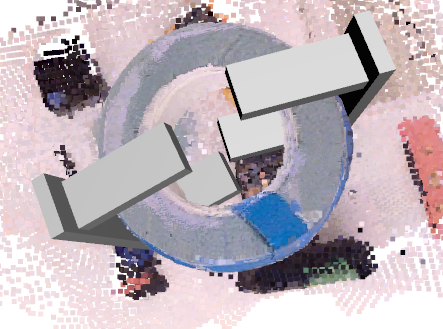}%
    \end{subfigure}}
    \caption{Top: Selected objects (wrench, spoon, balance meter and tape) segmented; Bottom: Generated robot end-effector grasping poses for manipulator hand-overs.}
    \label{fig:bi-passing}
    \vspace{-2em}
\end{figure}

\begin{figure}[t]  
\centering
\fbox{%
\begin{subfigure}[t]{\linewidth}
\centering
\includegraphics[width=0.333\linewidth]{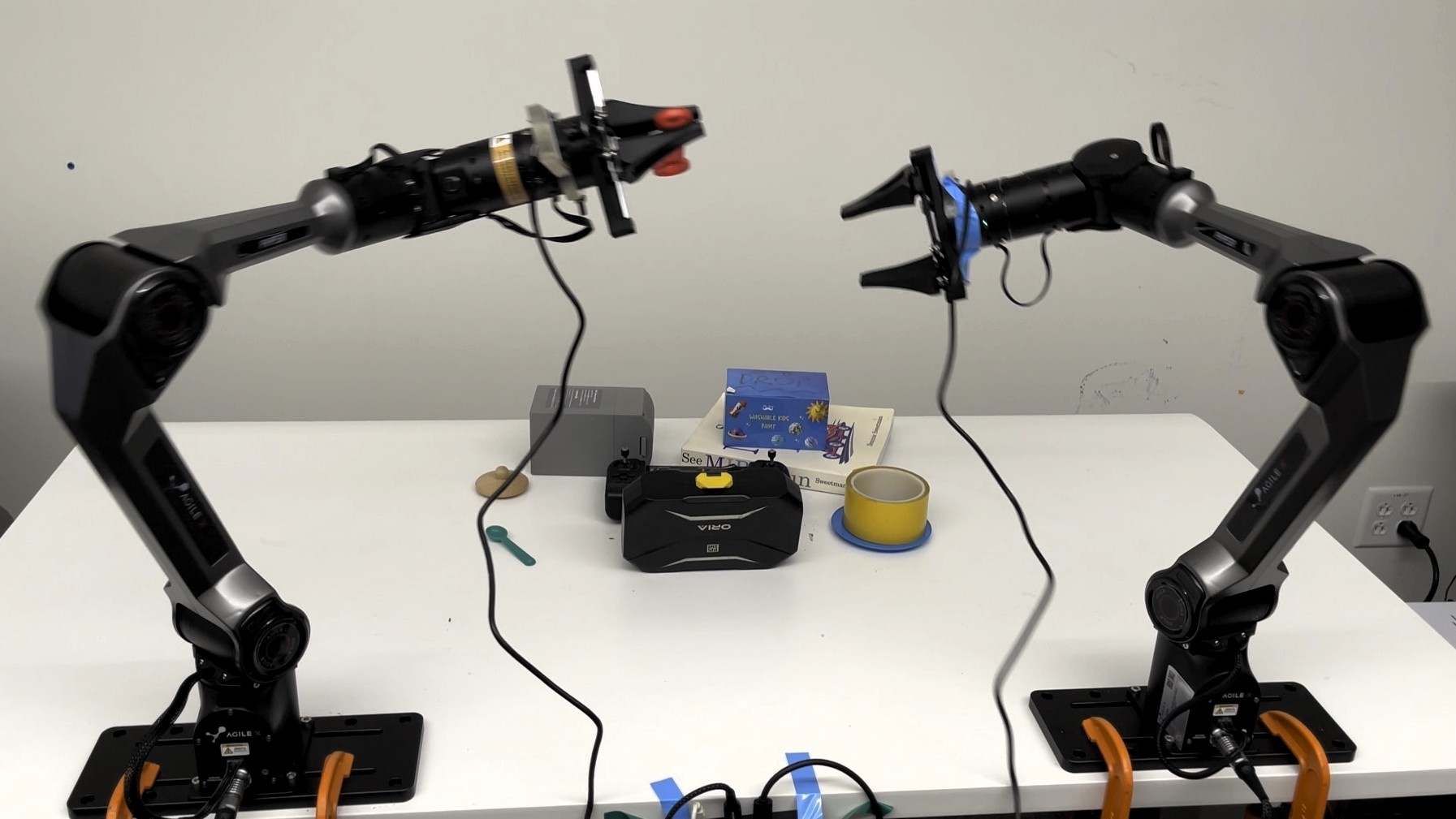}%
\includegraphics[width=0.333\linewidth]{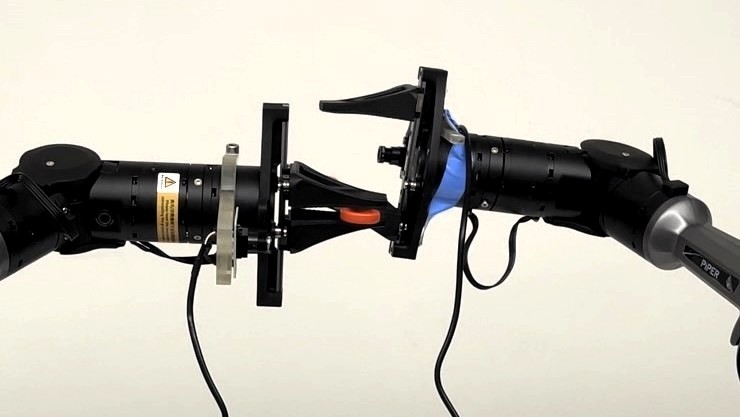}%
\includegraphics[width=0.333\linewidth]{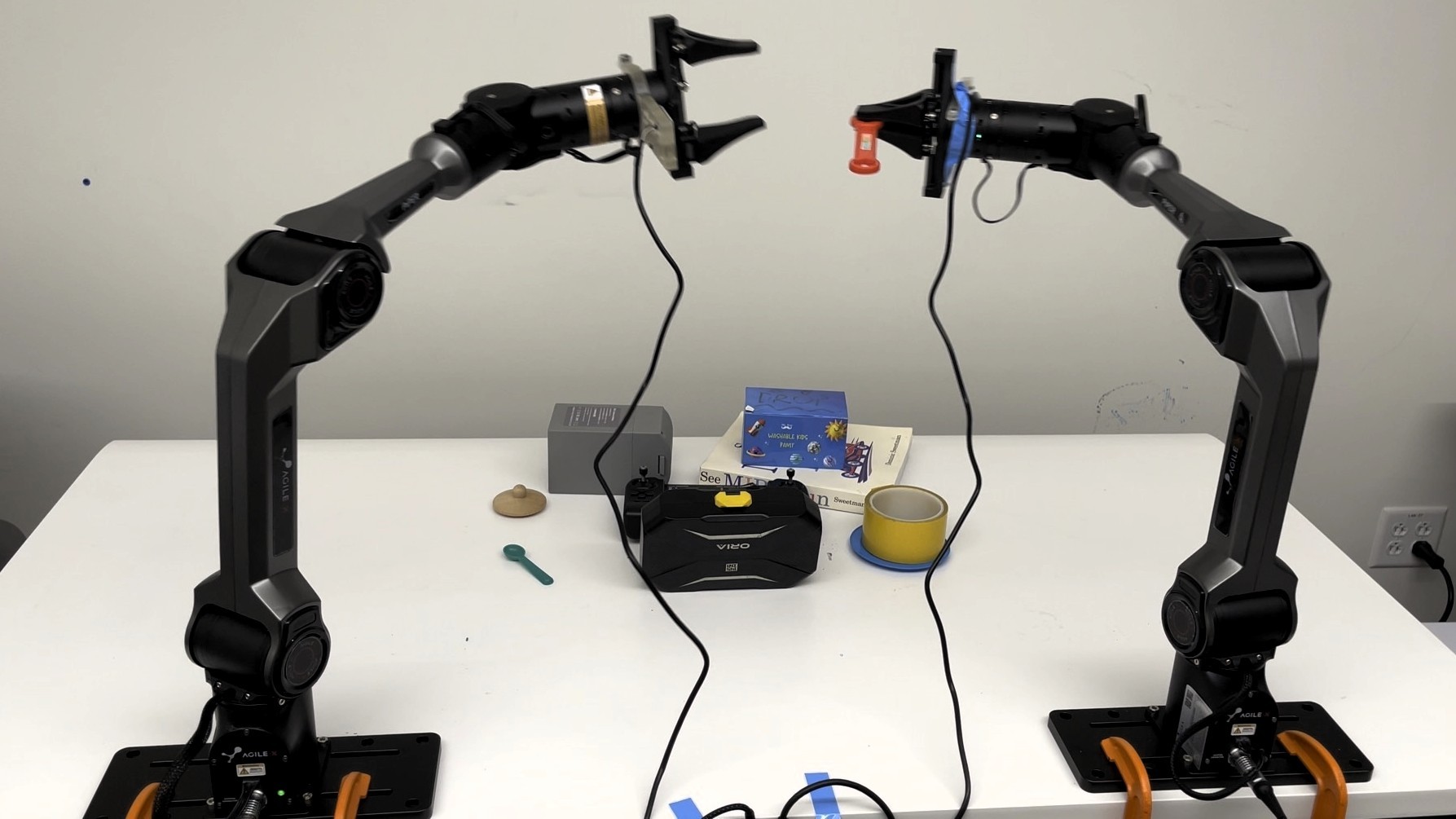}%
\end{subfigure}%
}
\fbox{%
\begin{subfigure}[t]{\linewidth}
\centering
\includegraphics[width=0.333\linewidth]{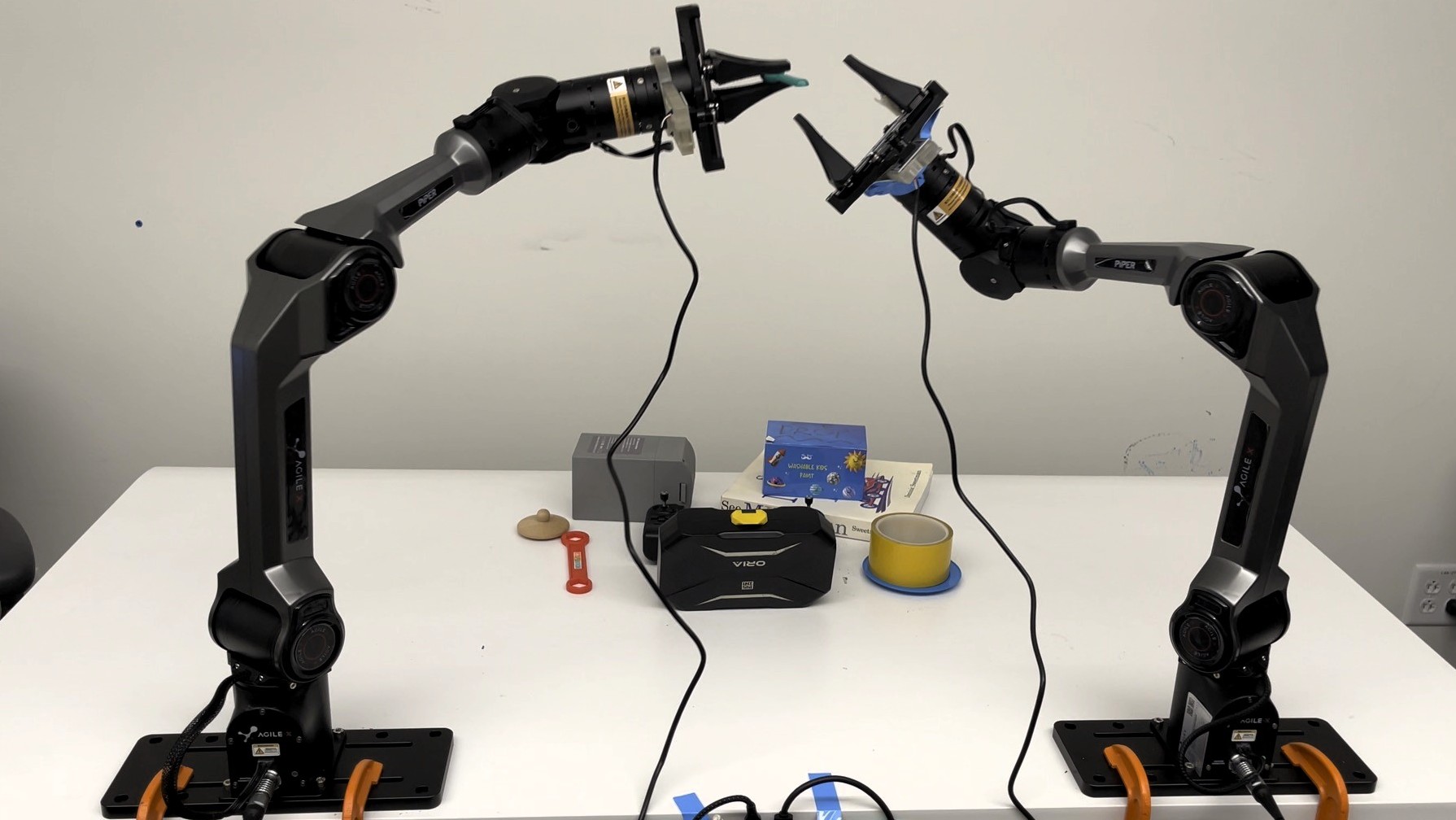}%
\includegraphics[width=0.333\linewidth]{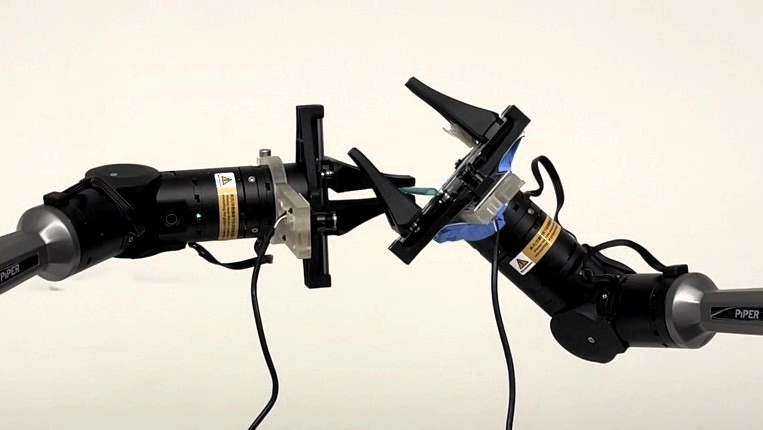}%
\includegraphics[width=0.333\linewidth]{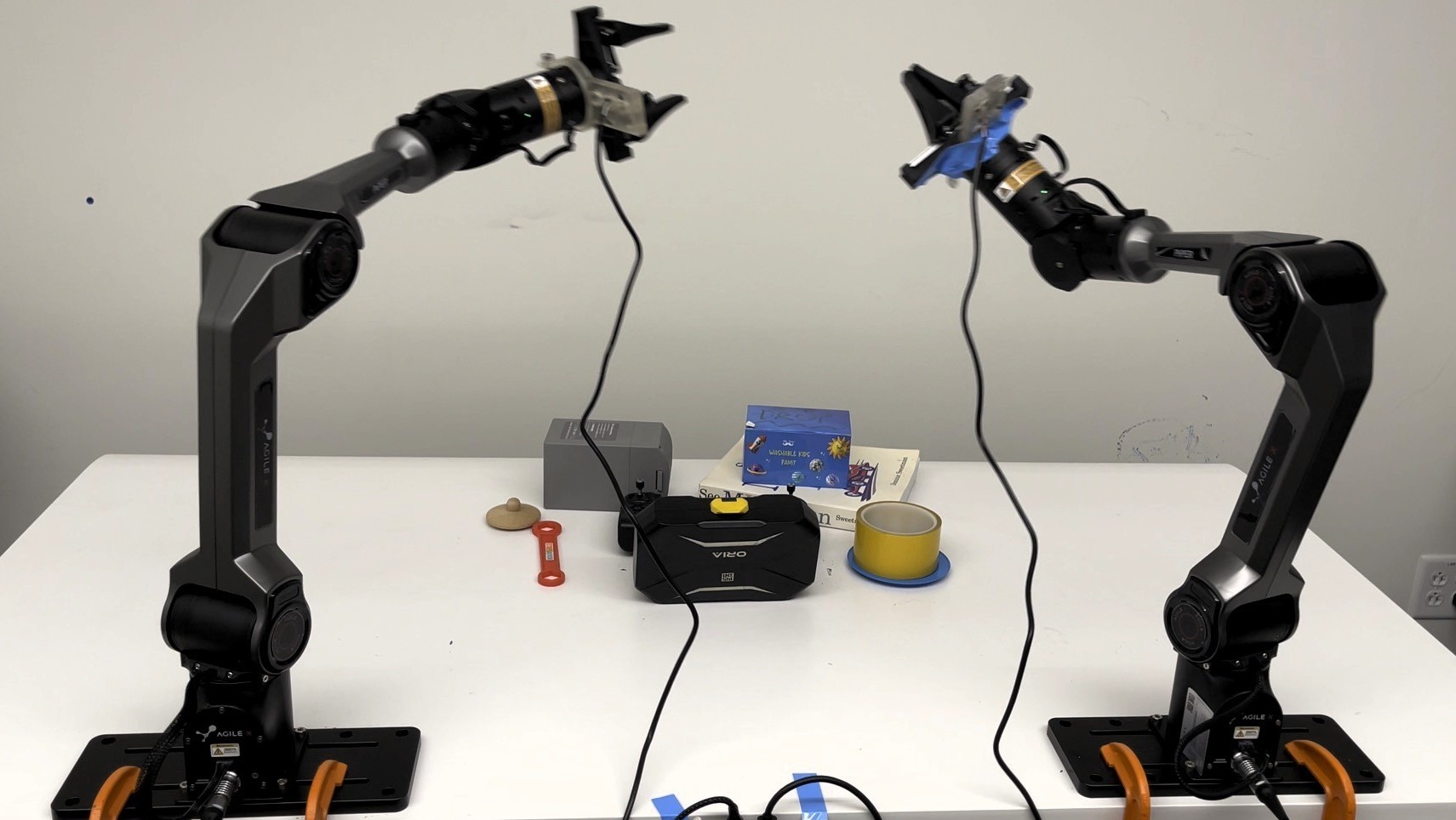}%
\end{subfigure}%
}
\fbox{%
\begin{subfigure}[t]{\linewidth}
\centering
\includegraphics[width=0.333\linewidth]{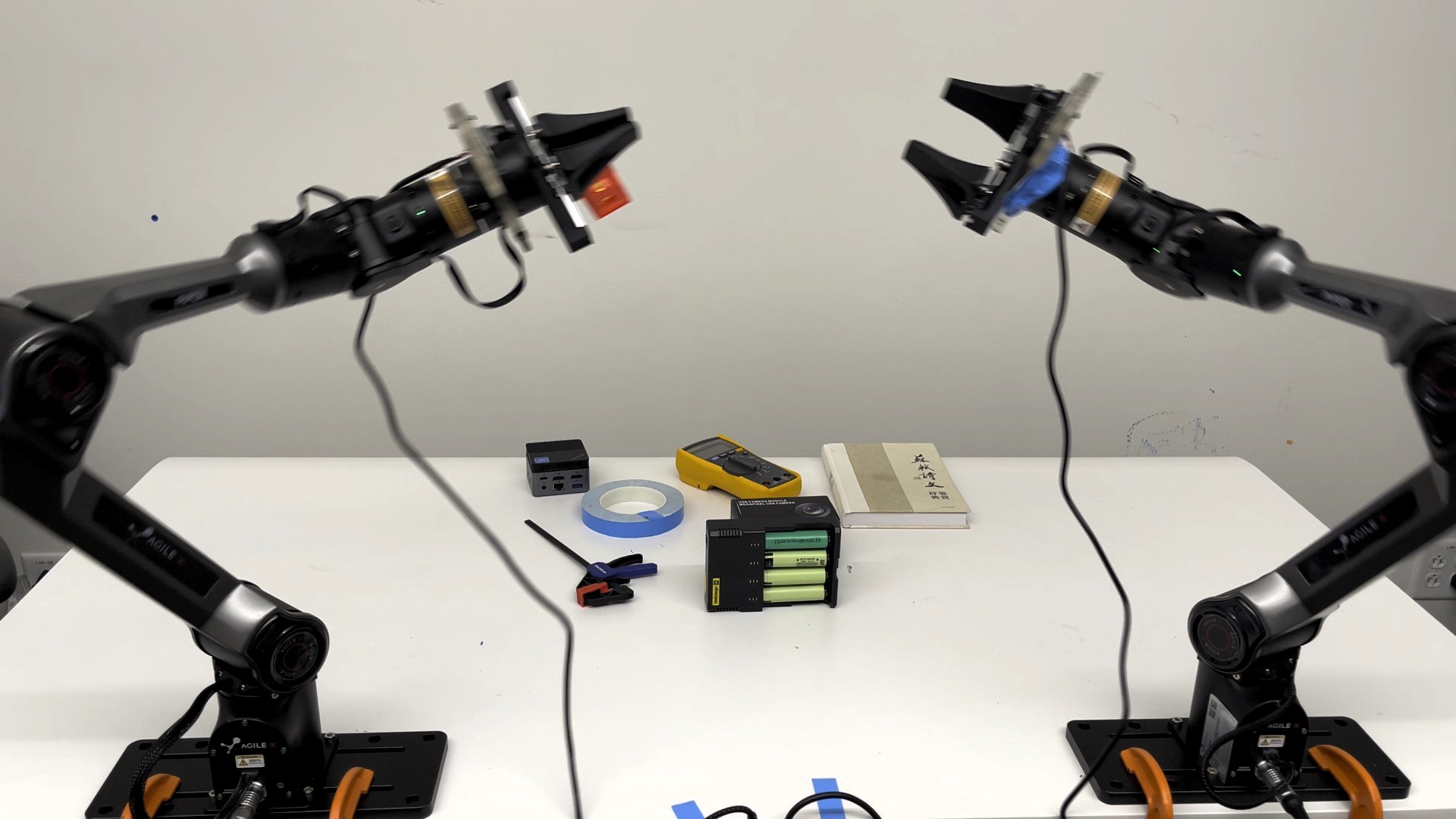}%
\includegraphics[width=0.333\linewidth]{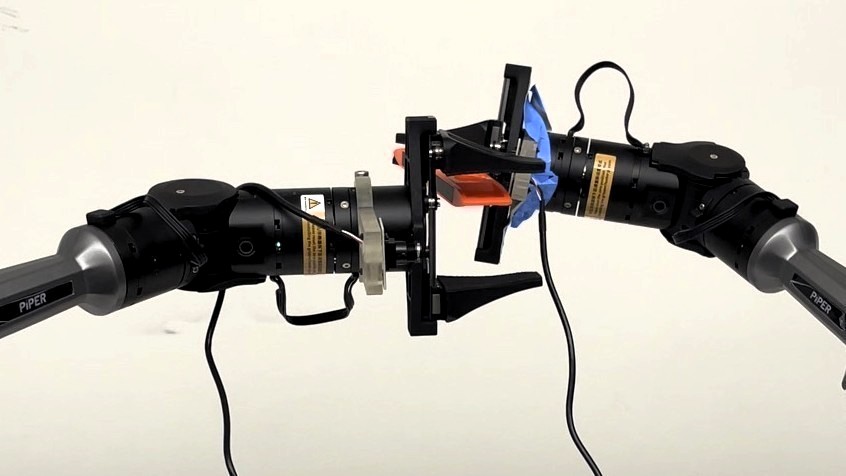}%
\includegraphics[width=0.333\linewidth]{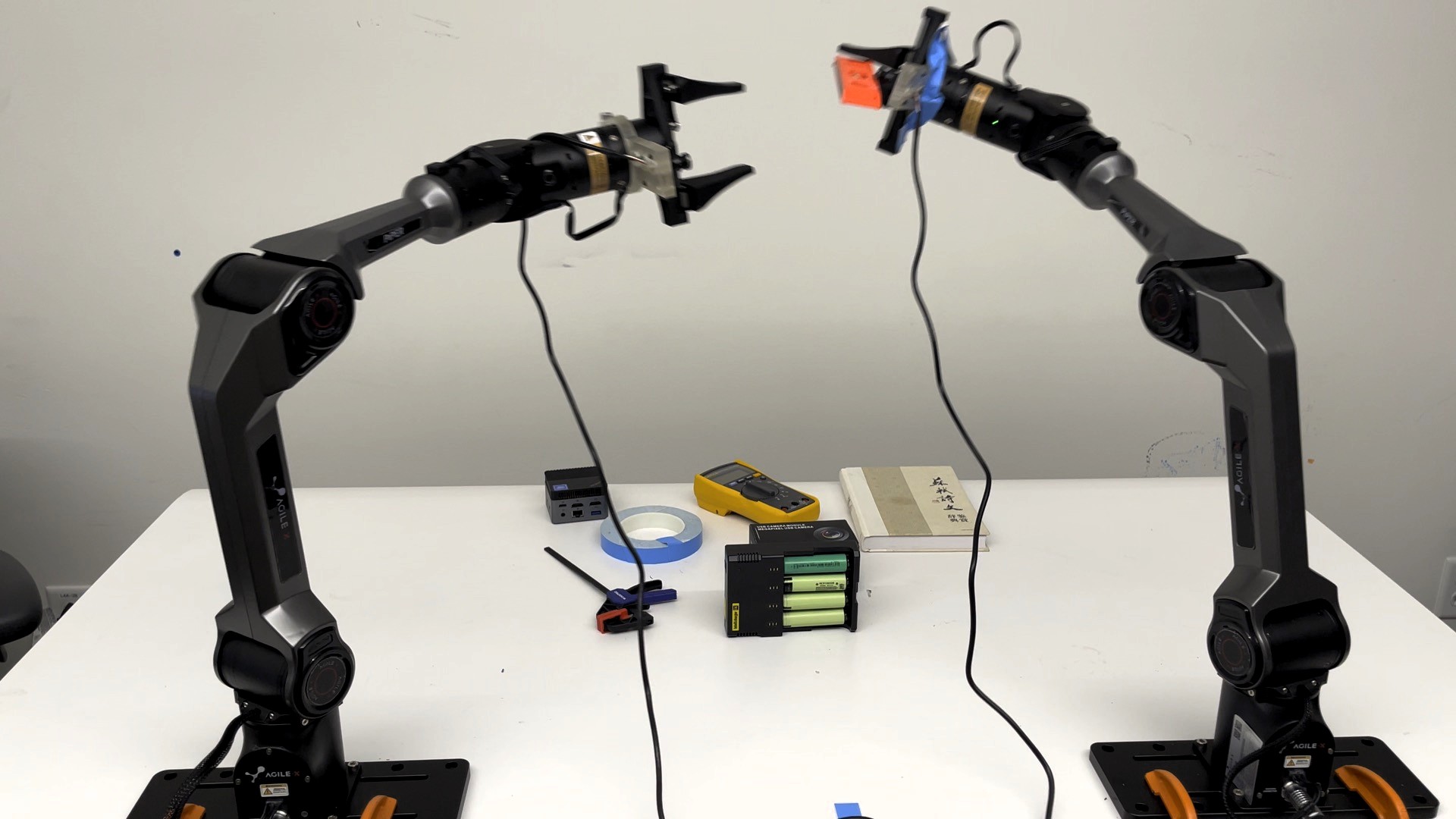}%
\end{subfigure}%
}
\fbox{%
\begin{subfigure}[t]{\linewidth}
\centering
\includegraphics[width=0.333\linewidth]{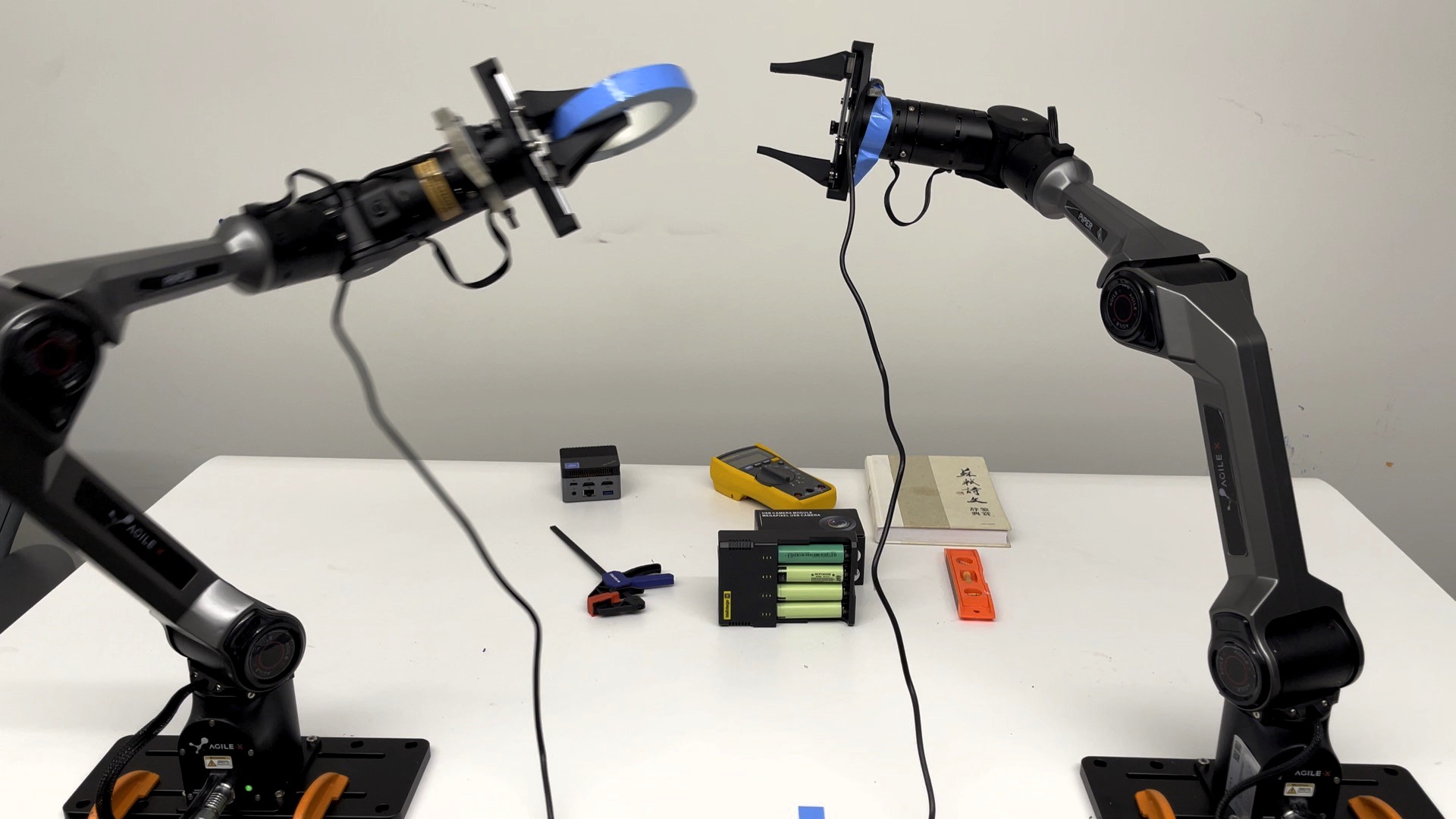}%
\includegraphics[width=0.333\linewidth]{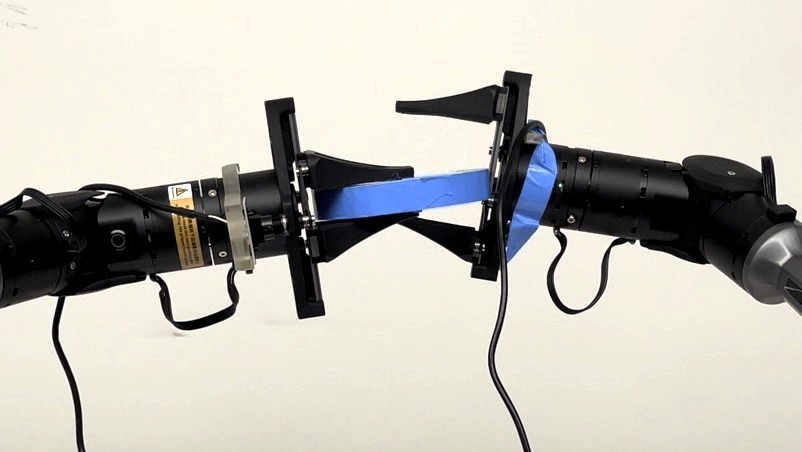}%
\includegraphics[width=0.333\linewidth]{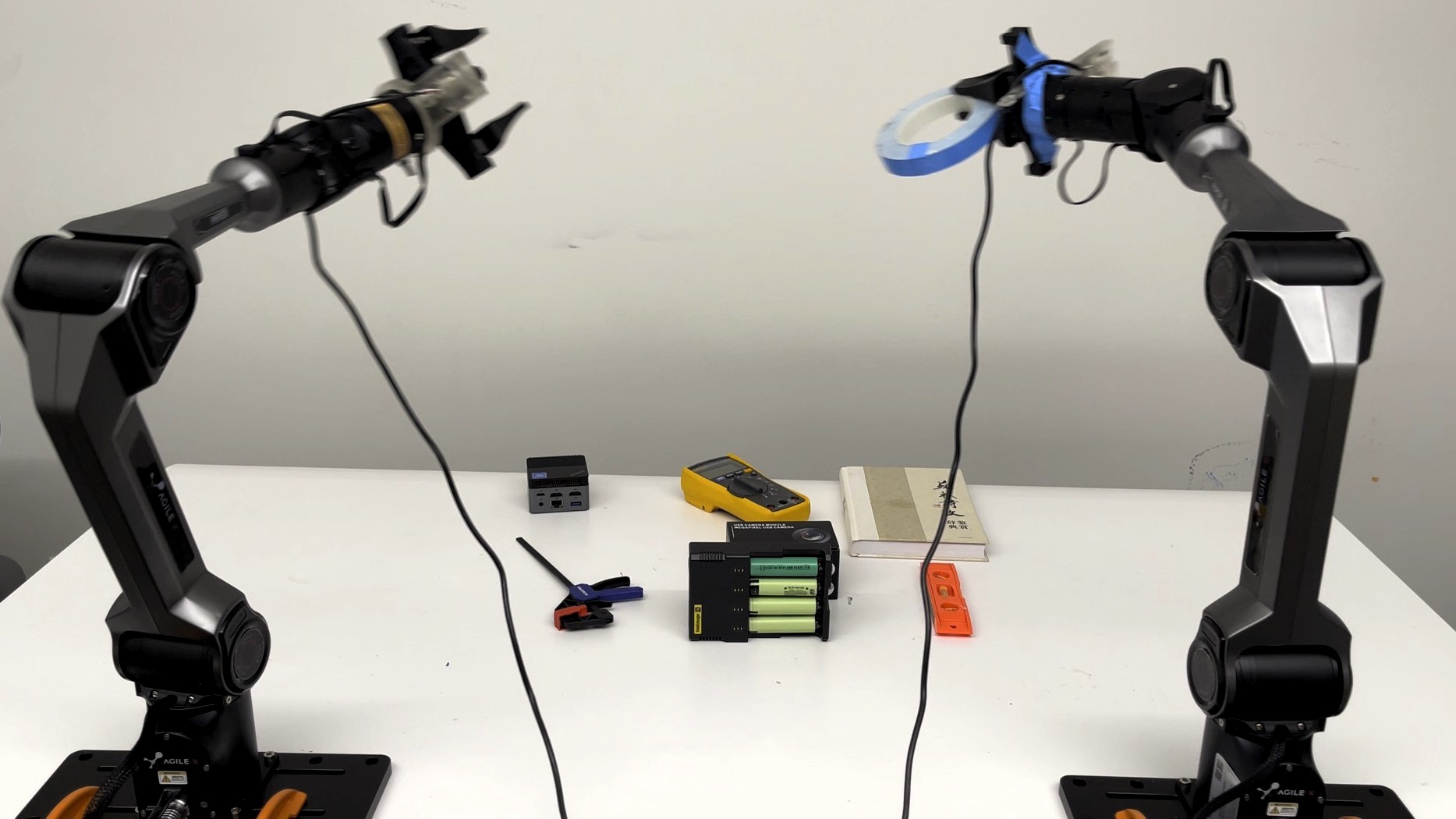}%
\end{subfigure}%
}
        \caption{We demonstrate passing of the objects: wrench, spoon, balance meter, tape from top to bottom.}\label{fig:passing_real_world}
        \vspace{-2em}
\end{figure}

\subsection{Bi-JCR Enables Bi-manual Downstream Tasks}
To assess both the relative‐pose estimation between the two manipulators’ bases and the fidelity of the reconstructed 3D scene, we conduct two real‐world bimanual tasks: (1) joint grasping of large objects and (2) passing of small objects. We begin by running Bi‐JCR to recover the base‐to‐base transformation and reconstruct the 3D environment in simulation (\cref{fig:Bi-JCR_segmentation_a}). Next, we apply the 3D point‐cloud segmentation algorithm from \cite{tang2025graphsegsegmented3drepresentations} to isolate each object on the tabletop (\cref{fig:Bi-JCR_segmentation_b}).

\textbf{Joint grasping:} We first select the cluster corresponding to the large objects: a toolbox, a controller, a battery pack and a box, as illustrated \cref{fig:joint_grasp}. A grasp pose for the primary manipulator is computed in its own base frame using \cite{ten2017grasp}, and likewise for the secondary manipulator. The secondary grasp must then be transformed into its base frame. Finally, both end‐effector poses are executed via inverse kinematics and joint control to perform the coordinated grasp.

\textbf{Bimanual passing:} We then focus on the small objects, a wrench, a spoon, a balance meter and a tape, shown in Fig. \ref{fig:bi-passing}. After choosing a target transfer location in the primary manipulator’s base frame, we translate the segmented point cloud to that location and generate precise poses to enable object passing for both arms. Again, the secondary end‐effector pose is reprojected into its base frame. The primary manipulator loads the object onto its gripper and then successfully hands it off to the secondary manipulator at the specified location (\cref{fig:bi-passing}). The successful passing of the objects highlights that the transformation between the local coordinate frames of each robot is accurately recovered. These experiments demonstrate that, Bi‐JCR reliably enables complex bimanual operations by accurately calibrating cameras on both manipulators and constructing a high‐quality dense 3D reconstruction.

\subsection{Summary of Empirical Results}
From our empirical experiments, we demonstrate that:
\begin{enumerate}
\item Bi-JCR’s calibration produces highly accurate eye-to-hand transformations.
\item The recovered scale factor successfully converts the foundation model’s output into real-world metric units.
\item Bi-JCR generates a high-quality dense 3D point cloud with correct object geometry and can be correctly transformed into the robot's frame.
\item Among 3D foundation models, DUSt3R \cite{DUSt3R_cvpr24} consistently delivers the best performance for Bi-JCR.
\item Gradient-descent refinement is effective under sparse-view conditions, although its marginal benefit diminishes as view density increases.
\item The accurate transformations and dense 3D representations produced by Bi-JCR enable various downstream bimanual tasks, including coordinated grasping and object transfer between manipulators.
\end{enumerate}

\section{Conclusion and Future work}
We introduced Bi-JCR, a unified framework for joint calibration and 3D representation in bimanual robotic systems with wrist-mounted cameras. Leveraging large 3D foundation models, Bi-JCR removes calibration markers and simultaneously estimates camera extrinsics, inter-arm relative poses, and a shared, metric-consistent scene representation. Our approach unifies the calibration and perception processes using only RGB images, and facilitates downstream bi-manual tasks such as grasping and object handover. Extensive real-world evaluations demonstrate Bi-JCR’s performance over diverse environments. Future work will leverage confidence masks from 3D foundation models to actively guide novel image collection, continuously complete the reconstructed scene, refine calibration, and then generate motion \cite{Diff_templates, diagrammaticlearning}.


\bibliographystyle{ieeetr} 
\bibliography{example}

\end{document}